\documentclass[journal]{IEEEtran}
\ifCLASSINFOpdf
\else
\fi
\usepackage{cite}
\usepackage{multirow}
\usepackage{multicol}
\usepackage{color}
\usepackage{amssymb,amsfonts,amsthm}
\usepackage[linkcolor=blue,colorlinks=true, citecolor = blue]{hyperref}
\usepackage{amsmath}       
\usepackage{amssymb}
\usepackage{mathtools}
\usepackage{amsthm}
\usepackage{mathrsfs}
  {
      \theoremstyle{plain}
      \newtheorem{assumption}{Assumption}
  }
\usepackage{algorithm}
\usepackage{algpseudocode}  
\usepackage{graphicx}
\usepackage[misc]{ifsym} 
\usepackage{subfigure} 
\usepackage{textcomp}
\usepackage{xcolor}
\usepackage{dsfont}
\usepackage{mathrsfs}
\usepackage{bbm}
\usepackage{booktabs} 
\usepackage{diagbox}
\usepackage{verbatim}
\usepackage{setspace}
\newtheorem{lemma}{Lemma}
\newtheorem{theorem}{Theorem}

\newtheorem{corollary}{Corollary}

\allowdisplaybreaks[4]
\def\BibTeX{{\rm B\kern-.05em{\sc i\kern-.025em b}\kern-.08em
    T\kern-.1667em\lower.7ex\hbox{E}\kern-.125emX}}
\begin{document}
%
\title{Decentralized Adversarial Training over Graphs}
%
%
%

\author{Ying~Cao*,  Elsa~Rizk,  Stefan~Vlaski, \IEEEmembership{Member,~IEEE,} Ali~H. Sayed, \IEEEmembership{Fellow,~IEEE}
\thanks{A short version of this work without proofs and focusing only on the strongly-convex case appears in the conference publication \cite{ying_icassp2023}.}
\thanks{*\ Corresponding author. Ying Cao, Elsa Rizk, Ali H. Sayed are with the School of Engineering, École Polytechnique Fédérale de Lausanne. Stefan Vlaski is with the Department of Electrical and Electronic Engineering, Imperial College London. Emails: \{ying.cao, ali.sayed\}@epfl.ch, elsa.rizk@alumni.epfl.ch, s.vlaski@imperial.ac.uk.}
}

\maketitle

\begin{abstract}
The vulnerability of machine learning models to adversarial attacks has been attracting considerable attention in recent years. Most existing studies focus on the behavior of stand-alone single-agent learners. In comparison, this work studies adversarial training over graphs, where individual agents are subjected to perturbations of varied strength levels across space. It is expected that interactions by linked agents, and the heterogeneity of the attack models that are possible over the graph, can help enhance robustness in view of the coordination power of the group. Using a min-max formulation of distributed learning, we develop a decentralized adversarial training framework for multi-agent systems. Specifically, we devise two decentralized adversarial training algorithms by relying on two popular decentralized learning strategies--diffusion and consensus. We analyze the convergence properties of the proposed framework for strongly-convex, convex, and non-convex environments, and illustrate the enhanced robustness to adversarial attacks. 
\end{abstract}

\begin{IEEEkeywords}
Robustness, adversarial training, decentralized setting, diffusion learning, multi-agent system
\end{IEEEkeywords}

%
\IEEEpeerreviewmaketitle

\section{Introduction}
In many machine learning algorithms, small malicious perturbations that are imperceptible to the human eye can cause classifiers to reach erroneous conclusions  \cite{szegedy2013intriguing, BaiL0WW21,sayed_2023}. This sensitivity is problematic for important applications, such as computer vision \cite{song2017pixeldefend, HendrycksZMTLSS22}, natural language processing \cite{miyato2016adversarial,jia2017adversarial}, and reinforcement learning \cite{pinto2017robust}.  Several defense mechanisms have been proposed in the literature \cite{GuR14, kannan2018adversarial, LuoBRPZ15, PapernotM0JS16, madry2017towards}  to mitigate the negative effect of adversarial examples, including the popular scheme based on adversarial training \cite{madry2017towards}. In this approach, clean training samples are augmented by adding purposefully crafted perturbations. Due to the lack of an explicit definition for the imperceptibility of perturbations, additive attacks are usually restricted within a small bounded region. 

Most earlier studies, such as \cite{miyato2016adversarial,goodfellow2014explaining,madry2017towards,maini2020adversarial,zhang2019theoretically}, focus on studying adversarial training in the context of single agent learning. However, distributed settings, which consist of a group of agents, are becoming more prevalent in a highly connected world \cite{chang2020distributed, sayed2014adaptive, KarMR12, Gul21, SridharP23,BandeGV19}.  Examples abound in transportation networks, communication networks and biological networks. {In this work, we propose a general decentralized framework to improve the robustness of networked models to adversarial attacks. Specifically, we devise two robust training algorithms for multi-agent systems by relying on two popular decentralized strategies, i.e., diffusion and consensus \cite{sayed_2023,sayed2014adaptation,sayed2014adaptive}, and analyze their convergence properties in strongly-convex, convex, and non-convex environments.}

Although the convergence properties of decentralized learning algorithms in standard training (i.e., with clean data) have been well-studied in the literature \cite{sayed2014adaptation, YuanLY16, ChenS15, vlaski2019diffusion} for convex and non-convex environments, their approaches cannot be directly extended to the adversarial learning scenario. This limitation arises because traditional analyses rely heavily on the Lipshcitz continuity of the gradients of risk functions (or equivalently, the assumption of bounded Hessian matrices). However, the adversarial perturbations introduce changes to the properties of risk functions. As we will demonstrate, in adversarial learning, formulating the ``real" loss function as a maximization problem over perturbations results in risk functions that do not necessarily satisfy the Lipschitz condition. Moreover, in the general case where the optimal perturbation of the maximization problem is not unique, the risk functions are non-differentiable. These factors make the analysis of the decentralized learning algorithms in the adversarial setting more challenging. To address these issues, we use analytical frameworks that are different from the literature on clean training data to establish the convergence of the proposed decentralized adversarial training algorithms.

 Our analytical framework in the strongly-convex and convex cases is inspired by \cite{ying2018performance2}, according to which we first verify the convergence of the network disagreement, which we subsequently use to analyze the convergence of the proposed algorithms. However, our proofs differ from \cite{ying2018performance2}, which focuses on the strongly-convex case and requires homogeneous networks, as well as the initial distance of all local models from the global minimizer to be bounded. In contrast, we establish convergence results for both the strongly-convex and convex cases, remove the bounded initial distance assumption, and consider more general heterogeneous environments. 
In the analysis for the non-convex case, existing works of single-agent adversarial training usually assume the loss function in the inner maximization step to be strongly-concave \cite{sinha2017certifying, WangM0YZG19, zhang2020distributed}, which ensures some useful properties for the risk functions. In contrast, we consider the general nonconcave case for adversarial learning in the multi-agent setting. To carry out the analysis, we extend the analytical framework of \cite{jin2020local, DavisD19}, which resorts to the Moreau envelope as a surrogate for non-differentiable and non-smooth risk functions when examining the stationarity of gradient algorithms. 

There of course exist other works in the literature that have applied adversarial training to a multiplicity of agents, albeit using a different architecture {or solving different problems.} The works \cite{qin2019adversarial,zhang2020distributed,liu2021concurrent} employ multiple GPUs and a fusion center to accelerate the training process. {The work \cite{zhang2020distributed} extends adversarial training from the single-agent setting to a centralized distributed framework. In their approach, local agents transmit their gradient information to a central server, which aggregates the gradients. Subsequently, all agents update their parameters using the aggregated gradient. Another line of work examines the adversarial robustness of neural networks in the context of federated learning \cite{Li0Y23, KumarMC24}, while works \cite{feng2019graph,xu2019topology,wang2019graphdefense} investigate the adversarial robustness of graph neural networks, and \cite{BukharinLYZCZZZ23,LiuL23} focus on the adversarial robustness of multi-agent reinforcement learning. In this paper, we focus on a fully decentralized architecture \cite{MageshV22,MerhavC20,LoizouR21,FereydounianMPH23} where each agent corresponds to a learning unit in its own right, and interactions occur locally over neighborhoods determined by a graph topology. There are also works focusing on improving robustness of decentralized frameworks to Byzantine attacks, in which some malicious agents send incorrect information to their neighbors to destabilize or mislead the network\cite{WuCL23,YangGB20}. In contrast, we focus on improving the adversarial robustness of models to imperceptible perturbations on samples.}

The contributions of this work are listed as follows:

(1) We formulate a sequential minimax optimization problem involving adversarial samples, where the perturbations are within some general $\ell_p$ norm-bounded region. To solve the problem, we propose a {general} decentralized framework, where all agents are subjected to adversarial examples and work together through local interactions to defend the network globally. {Specifically, to improve the robustness of multi-agent systems against adversarial examples, we develop two decentralized adversarial training algorithms based on two widely used decentralized strategies--diffusion and consensus. In our algorithms, if the closed-form solution of the inner maximization problem is accessible--which holds for some popular convex loss functions--it is directly added to the clean data. However, in the general case where the exact solution is not obtainable, an approximate one is used in the algorithms.} 

(2) In the performance analysis, we examine the convergence of the proposed framework for strongly-convex, convex, and non-convex optimization environments. Basically, we show that for strongly-convex loss functions, {in the general case where the optimal perturbation of the inner maximization problem is not unique and analytically accessible, then the risk functions are not differentiable. In this case, the proposed algorithms are able to approach the global minimizer 
within $O(\mu)+ O(\epsilon)$ after sufficient iterations at a linear rate, where $\mu$ is the step-size parameter, and the error term $O(\epsilon)$ arises due to the approximation error when solving the maximization problem over perturbations. Also, $\epsilon$ is the maximum perturbation bound over the entire graph, which is small in the context of adversarial training. In particular, if the closed-form solution of the optimal perturbation is known, then the error term can be removed. In comparison, in the convex case, the proposed algorithms are verified to converge to a $O(\mu)$ neighborhood of a global minimum after sufficient iterations at a sub-linear rate of $O(\frac{1}{\mu N})$, where $N$ is the number of iterations. In addition,} for non-convex losses, the algorithms are guaranteed to converge to a point that is $O(\mu) + O(\epsilon^2)$ close to an approximate stationary point of the global objective function {at a sub-linear rate $O(\frac{1}{\mu N})$. Still, the error term $O(\epsilon^2)$ reflects the approximation error associated the inner maximization.}  

(3) In the simulations, we illustrate how the robustness of the multi-agent systems can be improved by the proposed algorithms. We simulate both the convex and non-convex environments. Furthermore, we simulate both homogeneous and heterogeneous networks. We illustrate how decentralized adversarial training algorithms enable more robust behavior than non-cooperative and even centralized methods in  non-convex environments. This fact illustrates the role of the graph topology in resisting adversarial attacks.

\section{{Foundations and Approach}}\label{sec2}
\subsection{Problem formulation}\label{s2A}
We consider a collection of $K$ agents over a graph, where each agent $k$ observes independent realizations of some random data $(\boldsymbol{x}_k,\boldsymbol{y}_k)$, in which $\boldsymbol{x}_k$ plays the role of the feature vector and $\boldsymbol{y}_k$ plays the role of the label variable. Adversarial learning in the heterogeneous decentralized setting deals with the following stochastic minimax problem:
\begin{equation}\label{formu}
w^\star  = \mathop{\mathrm{argmin}}\limits_{{w}\in \mathbbm{R}^M} \left\{J( w) \overset{\Delta}{=} \sum\limits_{k=1}^K \pi_k J_k( w)\right\}
\end{equation}
where $M$ is the dimension of the feature vectors,  $\{\pi_k\}_{k=1}^{K}$ are positive scaling weights adding up to one, and each individual risk function is defined by
\begin{equation}\label{jk}
\begin{split}
    J_k(w) &= \mathds{E}_{\{\boldsymbol{x}_k,\boldsymbol{y}_k\}}\left\{ \max\limits_{\left\Vert\delta\right\Vert_{p_k}\le \epsilon_k}  Q_k(w;\boldsymbol{x}_k +\delta, \boldsymbol{y}_k)\right\} \\
\end{split}
\end{equation}
in terms of a loss function $Q_k(\cdot)$. In this formulation, $\boldsymbol{y}_k$ is the true label of sample $\boldsymbol{x}_k$, and the variable $\delta$ represents a norm-bounded perturbation {added to the clean sample}.  In this paper, we assume all agents observe independently and identically distributed (i.e., iid) data over time, but allow {for} non-iid samples over space. The type of the norm $p_k$ and the upper bound $\epsilon_k$ {are allowed to} vary across agents, thus leading to the possibility of heterogeneous models over the graph. In particular, $\ell_2$ and $\ell_\infty$ norms are two popular choices under which the results of this paper hold. We refer to $w^{\star}$ as the \textit{robust model} since it relies on the worst-case perturbation found from \eqref{jk}.
\subsection{Algorithm development}\label{s2B}
One methodology for solving (\ref{formu}) is to first determine the inner maximizer in (\ref{jk}), thus reducing the minimax problem to a standard stochastic minimization formulation. Then, the traditional stochastic (sub)gradient method could be used to seek the {robust} minimizer. 

We denote the true maximizer of the perturbed loss function in (\ref{jk}) by
\begin{align}\label{maximizer_t}
    \boldsymbol{\delta}_k^{\star} (w) \in \mathop{\mathrm{argmax}}\limits_{\left\Vert\delta\right\Vert_{p_k}  \le \epsilon_k}   Q_k(w;\boldsymbol{x}_k +\delta,\boldsymbol{y}_k)
\end{align}
where the dependence of $\boldsymbol{\delta}_k^\star$ on $w$ is shown explicitly. We use the boldface notation for $\boldsymbol{\delta}_{k}^{\star}(\cdot)$ because it is now a function of the random data $(\boldsymbol{x}_k,\boldsymbol{y}_k)$. To apply the stochastic gradient method, we would need to evaluate the gradient of $Q(w;\boldsymbol{x}_k+\boldsymbol{\delta}_k^{\star}(w),\boldsymbol{y}_k)$ relative to $w$, which can be challenging since $\boldsymbol{\delta}_k^{\star}(w)$ is also dependent on $w$. Fortunately, this difficulty can be resolved by appealing to Danskin's theorem -- see section 6.11 in \cite{bertsekas2009convex} and section B.3 in \cite{Thekumparampil019}. {Let 
\begin{align}\label{f_g}
    f_k(w;\boldsymbol{x}_k,\boldsymbol{y}_k) \overset{\Delta}{=} \max_{\left\Vert\delta\right\Vert_{p_k}\le \epsilon_k} Q_k(w;\boldsymbol{x}_k+\delta,\boldsymbol{y}_k)
\end{align}
which can be viewed as the ``real" loss function of (\ref{jk}), so that the risk function of each agent in (\ref{jk}) is given by
\begin{align}\label{J_k_f}
 J_k(w) &= \mathds{E}_{\{\boldsymbol{x}_k,\boldsymbol{y}_k\}} \left\{f_k(w;\boldsymbol{x}_k,\boldsymbol{y}_k)\right\} 
\end{align}}Then, the theorem asserts that $f(w;\boldsymbol{x}_k,\boldsymbol{y}_k)$ remains (strongly) convex over $w$ if $Q_k(w;\cdot,\cdot)$ is {(strongly) convex} over $w$. However, $f_k(w;\boldsymbol{x}_k,\boldsymbol{y}_k)$ need not be differentiable over $w$ even when $Q_k(w;\cdot,\cdot)$ is differentiable. Nevertheless, and importantly for our purposes, we can determine subgradients for $f_k(w;\boldsymbol{x}_k,\boldsymbol{y}_k)$ by using the actual gradient of the loss evaluated at the worst perturbation, namely, it holds that
\begin{align}\label{danskin_ch}
    \partial_w f_k(w;\boldsymbol{x}_k,\boldsymbol{y}_k) = \mathrm{conv}\bigg\{ \nabla_w Q_k(w;\boldsymbol{x}_k+\boldsymbol{\delta}_k^\star,\boldsymbol{y}_k)\notag\\
     \Big\vert\boldsymbol{\delta}_k^\star \in \mathop{\mathrm{argmax}}\limits_{\left\Vert\delta\right\Vert_{p_k}\le \epsilon_k} Q_k(w;\boldsymbol{x}_k+\delta,\boldsymbol{y}_k)\bigg\}
\end{align}
which means the set of subgradients $\partial_w f_k(w;\boldsymbol{x}_k,\boldsymbol{y}_k)$ is the convex hull of the gradients $\nabla_w Q_k(w;\boldsymbol{x}_k+\boldsymbol{\delta}_k^\star,\boldsymbol{y}_k)$ evaluated over all possible maximizers $\boldsymbol{\delta}_k^\star$. This also implies that
\begin{align}
\label{danskin}
    \nabla_w Q_k(w;\boldsymbol{x}_k+\boldsymbol{\delta}_k^\star,\boldsymbol{y}_k)\in \partial_w f_k(w;\boldsymbol{x}_k,\boldsymbol{y}_k)
\end{align}
In {(\ref{danskin_ch}) and (\ref{danskin})}, the gradient of $Q_k(\cdot)$ relative to $w$ at the perturbed data is computed by treating $\boldsymbol{\delta}_k^{\star}$ as a stand-alone vector and ignoring its dependence on $w$. When $\boldsymbol{\delta}_k^{\star}$ in (\ref{maximizer_t}) happens to be \textit{unique}, then the gradient on the left in (\ref{danskin}) will be equal to the right side, so that in that case the function $ f_k(w;\boldsymbol{x}_k,\boldsymbol{y}_k)$ will be \textit{differentiable}. In this scenario, the stochastic gradient methods can be employed to solve \eqref{formu}. However, in the more general case where  $\boldsymbol{\delta}_k^{\star}$ is not unique, only the subdifferential property in (\ref{danskin_ch}) can be guaranteed. In this case, stochastic subgradient methods are applicable.

Moreover, while closed-from solutions to the maximization problem in (\ref{f_g}) exist for some convex loss functions (as will be discussed later), obtaining such solutions is generally not possible, especially for non-convex loss functions. Instead, one is satisfied with some common approximations \cite{maini2020adversarial,miyato2016adversarial,madry2017towards}. One popular strategy is to resort to a first-order Taylor expansion for each agent around $\boldsymbol{x}_k$ as follows:
\begin{align}\label{eqtaylor}
\begin{small}
  Q_k(\boldsymbol{w}_k;\boldsymbol{x}_k +\delta,\boldsymbol{y}_k)\approx  Q_k(\boldsymbol{w}_k;\boldsymbol{x}_k,\boldsymbol{y}_k)+ \delta^{\sf T}\nabla_x Q_k(\boldsymbol{w}_k;\boldsymbol{x}_k,\boldsymbol{y}_k)
  \end{small}
\end{align}
from which the following approximate maximizer can be derived:
\begin{align}\label{maximizer}
    \boldsymbol{\widehat{\delta}}_k = \mathop{\text{argmax}}\limits_{\left\Vert\delta\right\Vert_{p_k} \le \epsilon_k} \delta^{\sf T} \nabla_x Q_k(\boldsymbol{w}_k;\boldsymbol{x}_k,\boldsymbol{y}_k)
\end{align}
Note that different choices for the $p_k$ norm lead to different expressions for the approximate maximizer. For instance,  when $p_k=2$, we get:
\begin{equation}\label{p2}
\widehat{\boldsymbol{\delta}}_k = \epsilon_k \frac{\nabla_x Q_k(\boldsymbol{w}_k;\boldsymbol{x}_k,\boldsymbol{y}_k)}{\left\Vert\nabla_x Q_k(\boldsymbol{w}_k;\boldsymbol{x}_k,\boldsymbol{y}_k)\right\Vert}    
\end{equation}
 which motivates the fast gradient method (FGM) \cite{miyato2016adversarial} in single-agent learning, while for $p= \infty$ we have
 \begin{equation}\label{pin}
    \widehat{\boldsymbol{\delta}}_k = \epsilon_k 
\mathop{\mathrm{sign}}\Big(\nabla_x Q_k(\boldsymbol{w}_k;\boldsymbol{x}_k,\boldsymbol{y}_k)\Big) 
 \end{equation}
which is the expression of the fast gradient sign method (FGSM) \cite{goodfellow2014explaining}. Moreover, depending on the number of iterations used when generating perturbations, the first-order gradient based methods include single-step attacks, such as FGM and FGSM, and multi-step ones, such as projected gradient descent (PGD) \cite{madry2017towards}, which can also be called multi-step FGM for $\ell_2-$bounded attacks and multi-step FGSM for $\ell_\infty-$bounded attacks. Although the latter multi-step perturbations have been empirically shown to be stronger than the former ones \cite{madry2017towards} and lead to  more robust models, its computational complexity is also much higher\cite{WongRK20}.

Motivated by the above discussion, and using (\ref{danskin}), we can now devise algorithms to enhance the robustness of multi-agent systems to adversarial perturbations. To do so, we consider two popular decentralized strategies.

The first one is the adapt-then-combine (ATC) version of the \textit{diffusion} strategy \cite{sayed_2023, sayed2014adaptive},  for which we can write down the mini-batch adversarial extension to solve (\ref{formu})--(\ref{jk}). Basically, it consists of the following main three steps,
\begin{subequations}
\begin{align}
\label{d_d}
&\widehat{\boldsymbol{x}}_{k,n}^{b}= \boldsymbol{x}_{k,n}^b + \widehat{\boldsymbol{\delta}}^{b}_{k,n} \\
\label{d_x_e}
    &\boldsymbol{\phi}_{k,n} =\boldsymbol{w}_{k,n-1} - \frac{\mu}{B}\sum\limits_{b=1}^{B}\nabla_w Q_k(\boldsymbol{w}_{k,n-1};\widehat{\boldsymbol{x}}_{k,n}^b,\boldsymbol{y}_{k,n}^b)\\
\label{d_combine_e}
&\boldsymbol{w}_{k,n}  = \sum\limits_{\ell \mathcal{\in N}_k} a_{\ell k} \boldsymbol{\phi}_{\ell,n}  
\end{align}
\end{subequations}
{
where 
\begin{equation}\label{max_k}
    \widehat{\boldsymbol{\delta}}^{b}_{k,n} \approx \mathop{\mathrm{argmax}}\limits_{\left\Vert\delta\right\Vert_{p_k}  \le \epsilon_k}   Q_k(\boldsymbol{w}_{k,n-1};\boldsymbol{x}_{k,n}^b +\delta,\boldsymbol{y}_{k,n}^b)
\end{equation}
computes the approximate maximizer for each sample from the mini batch at iteration $n$,} and $\boldsymbol{\phi}_{k,n}$ is the intermediate state obtained by locally performing a stochastic gradient step at each agent. The intermediate states $\boldsymbol{\phi}_{\ell,n}$ from the neighbors of agent $k$ are then combined together using the scalars $a_{\ell k}$, which are non-negative and add to one over the neighborhood of agent $k$, denoted by $\ell \in {\cal N}_k$. These scalars are collected into a left stochastic matrix $A$.
Note that when the optimal perturbation (i.e., the exact maximizer) of (\ref{max_k}), denoted by $\boldsymbol{\delta}^{b,\star}_{k,n}$, is available, it should be used in place of the approximate one in the proposed algorithm. A detailed list of closed-form solutions of the maximization problem for some commonly used convex loss functions will be provided later. We list the details of adversarial diffusion training in Algorithm \ref{al_ro_1}.

\begin{algorithm}[htb]
\caption{: Adversarial diffusion training (ADT).}
\label{al_ro_1}
\begin{algorithmic}
\Require {Step size $\mu$, batch size $B$, combination matrix $A$.}
\Ensure{A group of robust models $\{w_{k,N}\}_{k=0}^{K}$.}
 \State{Initialize models for all agents at $\{w_{k,-1}\}_{k=0}^{K}$.}
 \For{Iteration $n = 0 \to N$}
    \For{ Agent $k = 1 \to K$}
        \State Select mini-batch training samples $\{\boldsymbol{x}_{k,n}^b,\boldsymbol{y}_{k,n}^b\}_{b=1}^B$.
        \For{  $b = 1 \to B$}
        \State
    {{$\widehat{\boldsymbol{x}}_{k,n}^{b}= \boldsymbol{x}_{k,n}^b + \widehat{\boldsymbol{\delta}}^{b}_{k,n} $}}
 \EndFor
 \State{$\boldsymbol{\phi}_{k,n} =\boldsymbol{w}_{k,n-1} - \dfrac{\mu}{B }\sum\limits_{b=1}^{B}\nabla_w Q_k(\boldsymbol{w}_{k,n-1};{\widehat{\boldsymbol{x}}_{k,n}^{b}},\boldsymbol{y}_{k,n}^b)$}
 \EndFor
 \For{ Agent $k = 1 \to K$}
 \State{$\boldsymbol{w}_{k,n}  = \sum\limits_{\ell \mathcal{\in N}_k} a_{\ell k} \boldsymbol{\phi}_{\ell,n}$}
 \EndFor
 \EndFor
\end{algorithmic}
\end{algorithm}

The second algorithm is built upon the consensus strategy \cite{sayed_2023, sayed2014adaptive}, and we list its adversarial extension in Algorithm \ref{al_ro_2}, which mainly consists of the following three steps,
\begin{subequations}
\begin{align}
\label{d_consen}
&\widehat{\boldsymbol{x}}_{k,n}^{b}= \boldsymbol{x}_{k,n}^b + \widehat{\boldsymbol{\delta}}^{b}_{k,n} \\
\label{x_e_consen}
&\boldsymbol{\psi}_{k,n} = \sum\limits_{\ell \mathcal{\in N}_k} a_{\ell k} \boldsymbol{w}_{\ell,n-1}\\
\label{combine_e_consen}
&\boldsymbol{w}_{k,n}  = \boldsymbol{\psi}_{k,n}  - \frac{\mu}{B}\sum\limits_{b=1}^{B}\nabla_w Q_k(\boldsymbol{w}_{k,n-1};\widehat{\boldsymbol{x}}_{k,n}^{b},\boldsymbol{y}_{k,n}^b)
\end{align}
\end{subequations}
Similar to the diffusion algorithm, the approximated maximizer of \eqref{max_k} is computed with the local model $\boldsymbol{w}_{k,n-1}$ for each sample from the mini batch. If the optimal perturbation is available, it should be used in the proposed algorithm directly. However, in this case, the existing iterates $\boldsymbol{w}_{\ell,n-1}$ are combined to generate the intermediate value $\boldsymbol{\psi}_{k,n}$, after which (\ref{combine_e_consen}) is applied, which leads to an asymmetry on the right-hand side of (\ref{combine_e_consen}). This asymmetry has been shown to shrink the stability range of consensus implementations over its diffusion counterparts in the convex and clean case, namely, diffusion is mean-square stable for a wider range of step-sizes $\mu$ \cite{sayed2014adaptive, sayed_2023}.

\begin{algorithm}[htb]
{
\caption{{: {Adversarial consensus training (ACT).}}}
\label{al_ro_2}
\begin{algorithmic}
\Require {Step size $\mu$, batch size $B$, combination matrix $A$.}
\Ensure{A group of robust models $\{w_{k,N}\}_{k=0}^{K}$.}
 \State{Initialize models for all agents at $\{w_{k,-1}\}_{k=0}^{K}$.}
 \For{Iteration $n = 0 \to N$}
    \For{ Agent $k = 1 \to K$}
        \State Select mini-batch training samples $\{\boldsymbol{x}_{k,n}^b,\boldsymbol{y}_{k,n}^b\}_{b=1}^B$.
        \For{  $b = 1 \to B$}
        \State
        {$\widehat{\boldsymbol{x}}_{k,n}^{b}= \boldsymbol{x}_{k,n}^b + \widehat{\boldsymbol{\delta}}^{b}_{k,n} $}
 \EndFor
 \For{ Agent $k = 1 \to K$}
 \State{$\boldsymbol{\phi}_{k,n} = \sum\limits_{\ell \mathcal{\in N}_k} a_{\ell k} \boldsymbol{w}_{\ell,n-1}$}
 \EndFor
 \State{$ \boldsymbol{w}_{k,n}  = \boldsymbol{\phi}_{k,n}  - \dfrac{\mu}{B}\sum\limits_{b=1}^{B}\nabla_w Q_k(\boldsymbol{w}_{k,n-1};\widehat{\boldsymbol{x}}_{k,n}^{b},\boldsymbol{y}_{k,n}^b)$}
 \EndFor
 \EndFor
\end{algorithmic}}
\end{algorithm}

 {We next} list the following assumptions for both convex and non-convex cases, which are commonly used in the literature of decentralized multi-agent learning, as well as in works on single-agent adversarial training \cite{sayed_2023, sayed2014adaptation, sinha2017certifying,  vlaski2021distributed, vlaski2021distributed2}.

First, it is necessary to define the combination matrix $A$ according to which the agents interact over the graph topology.
\begin{assumption}\label{as1}
    \textbf{(Strongly-connected graph)} The entries of the combination matrix $A = [a_{\ell k}]$ satisfy $a_{\ell k } \ge 0$ and the entries on each column add up to one, which means that $A$ is left-stochastic. Moreover, the graph is  strongly-connected, meaning that there exists a path with nonzero weights $\{a_{\ell k}\}$ linking any pair of agents and, in addition, at least one node $k$ in the network has a self-loop with $a_{kk}>0$.
    
     $\hfill\square$
\end{assumption}

It follows from the Perron-Frobenius theorem \cite{sayed2014adaptation,sayed_2023} that $A$ has a single eigenvalue at 1. Let $\pi = \{\pi_k\}_{k=1}^{K}$ be the corresponding right eigenvector of $A$. It then follows from the same theorem that we can normalize the entries of $\pi$ to satisfy:
\begin{align}\label{comb_matrix}
    A\pi = \pi, \quad \mathbbm{1}^{\sf T}\pi=1, \quad \pi_k>0
\end{align}

Next, we require the loss function $Q_k$ of each agent to be differentiable and smooth, which is a common assumption in the field of min-max optimization and adversarial learning \cite{sinha2017certifying, lin2020gradient, Thekumparampil019}. 
\begin{assumption}\label{as2}
 (\textbf{Smooth loss functions}) For each agent $k$, the gradients of the loss function relative to $w$ and $x$ are Lipschitz in relation to the  variables $\{w, x, y\}$. More specifically, it holds that
\begin{subequations}
\begin{small}
\begin{align}
\label{smooth_q}
&\left\Vert\nabla_w  Q_k(w_2;x+\delta,y) -  \nabla_w  Q_k(w_1;x+\delta,y) \right\Vert\le   L_{1}\left\Vert w_2-w_1\right\Vert\\
\label{as2_4}
&\left\Vert\nabla_w  Q_k(w;x_2+\delta,y) - \nabla_w Q_k(w;x_1+\delta,y) \right\Vert\le L_{2}\left\Vert x_2-x_1\right\Vert\\
\label{as2_5}
&\left\Vert\nabla_w  Q_k(w;x+\delta,y_2) - \nabla_w Q_k(w; x+\delta, y_1) \right\Vert\le L_{3}\left\Vert y_2-y_1\right\Vert
\end{align}
\end{small}
\end{subequations}
and
\begin{subequations}
\begin{small}
\begin{align}
\label{as2_3}
&\left\Vert\nabla_x  Q_k(w_2;x+\delta,y) - \nabla_x  Q_k(w_1;x+\delta,y) \right\Vert\le L_{4}\left\Vert w_2-w_1\right\Vert\\
\label{as2_2}
&\left\Vert\nabla_x  Q_k(w;x_2+\delta,y) - \nabla_x  Q_k(w;x_1+\delta,y) \right\Vert\le L_{5}\left\Vert x_2-x_1\right\Vert
\end{align}
\end{small}
\end{subequations}with $\Vert\delta\Vert_{p_k} \le \epsilon_k$. To simplify the notation, we {let}
\begin{align}
    L = \max\{L_{1}, L_{2}, L_{3},L_4, L_5\}.
\end{align}

$\hfill\square$
\end{assumption}

\section{{Analysis in (strongly-)convex environments}}

This section analyzes the convergence of decentralized adversarial training algorithms, as listed in Algorithms \ref{al_ro_1} and \ref{al_ro_2}, in the context of (strongly) convex loss functions. To do so, it is critical to distinguish between the cases where the optimal solution of \eqref{maximizer_t} is accessible and unique, and the case where it is not. As discussed in Section \ref{s2B}, when the closed-form solution to (\ref{maximizer_t}) is accessible and unique, it is used in the algorithms directly. In this case, we are actually using stochastic gradient methods to solve (\ref{formu}). Analytical solutions for \eqref{maximizer_t}, corresponding to several popular (strongly) convex loss functions, are provided in Appendix~\ref{ap_deltas}. Conversely, if the optimal perturbation of  (\ref{maximizer_t}) is not obtainable, we approximate it using the Taylor expansion, as shown in (\ref{eqtaylor}). In this case, since the ``real" loss function $f_k(w;\boldsymbol{x}_k, \boldsymbol{y}_k)$, as defined in (\ref{f_g}), may not be differentiable, we consider the general non-differentiable scenario. Consequently, Algorithms \ref{al_ro_1} and \ref{al_ro_2} operate as stochastic subgradient methods.

Moving on, considering $\epsilon = \max\limits_k \left\{\epsilon_k\right\}$, we first establish the following affine Lipschitz results. The proof can be found in Appendix \ref{ap1}.
{
\begin{lemma}\label{affine_l}
(\textbf{Affine Lipschitz})  For each agent $k$, and any $w_1, w_2, \delta_1, \delta_2$, it holds that
\begin{align}\label{affine_l_o}
 &\Vert \nabla_w  Q_k(w_2; \boldsymbol{x}_k+\delta_2,\boldsymbol{y}_k) - \nabla_w Q_k(w_1; \boldsymbol{x}_k+\delta_1,\boldsymbol{y}_k)\Vert  \notag\\
 &\le L\Vert w_2 - w_1 \Vert + O(\epsilon)   
\end{align}
where the $\ell_{p_k}-$norms of $\delta_1$ and $\delta_2$ are bounded by $\epsilon_k$. Specifically, when the solution of (\ref{maximizer_t}) is unique, $f_k(w;\boldsymbol{x}_k, \boldsymbol{y}_k)$ is differentiable. Then, the gradient of $f_k(w;\boldsymbol{x}_k, \boldsymbol{y}_k)$ is affine Lipschitz, namely, 
\begin{equation}\label{affine_l_2}
    \Vert \nabla_w f_k(w_2;\boldsymbol{x}_k, \boldsymbol{y}_k) - \nabla_w f_k(w_1;\boldsymbol{x}_k, \boldsymbol{y}_k) \Vert \le L\Vert w_2 - w_1 \Vert + O(\epsilon)
\end{equation}
which also implies
\begin{equation}\label{affine_l_3}
    \Vert \nabla_w J_k(w_2) - \nabla_w J_k(w_1) \Vert \le L\Vert w_2 - w_1 \Vert + O(\epsilon)
\end{equation}
$\hfill\square$
\end{lemma}}

{Contrary to the traditional analysis of decentralized learning algorithms \cite{sayed2014adaptation, ChenS15, YuanLY16}, which relies heavily on the gradient-Lipschitz conditions of the risk functions $J_k(w)$, adversarial perturbations introduce a challenge. Specifically, the gradients of the risks in \eqref{jk} are now affine Lipschitz, as demonstrated in \eqref{affine_l_2}, due to the additional term $O(\epsilon)$. Consequently, the traditional analysis methods are not directly applicable in this setting.} This fact motivates adjustments to the convergence arguments. A similar situation arises, for example, when we study the convergence of decentralized learning under non-smooth losses and clean environments --- see \cite{ying2018performance2,sayed_2023}.

To enable the convergence analysis of decentralized learning under the affine Lipschitz condition in the strongly-convex environment, the authors of \cite{ying2018performance2} assume that the initial distance of all agents’ models, i.e., $\boldsymbol{w}_{k,0}$, from the global minimizer are bounded. This assumption allows them to prove that the distance between the models of all agents and the global minimizer remains bounded throughout the training process. 
However, their argument assumes a homogeneous setting, where the minimizers of all local risk functions are identical. In this work, we extend the analysis to heterogeneous settings, where the minimizers of local risk functions may differ. To do so, we introduce the following assumption, which requires the network heterogeneity is not extreme.

\begin{assumption}\label{as4}
(\textbf{Bounded gradient disagreement}) For any pair of agents $k$ and $\ell$, the squared gradient disagreements are uniformly bounded on average, namely, for any $w$, it holds that
\begin{equation}\label{as4_1}
    \mathds{E}_{\{\boldsymbol{x},\boldsymbol{y}\}} \Vert \nabla_w Q_k(w;\boldsymbol{x} + \delta,\boldsymbol{y}) - \nabla_w Q_{\ell}(w;\boldsymbol{x} + \delta,\boldsymbol{y})\Vert^2 \le C^2
\end{equation}
for any $\Vert\delta\Vert_{p_k} \le \epsilon_k$, $\Vert\delta\Vert_{p_\ell}  \le \epsilon_\ell$.

 $\hfill\square$
\end{assumption}

This type of assumption is common in the literature, and is used for example in \cite{vlaski2021distributed, vlaski2021distributed2}. This condition is automatically satisfied when all agents use the same loss function, or in the centralized case where only a single agent is considered. More generally, when the agents are heterogeneous, the condition ensures that the level of the heterogeneity is not extreme.  {Moreover, as demonstrated later in the proof, this assumption facilitates the convergence of the distance between local agents and their centroid, which is crucial for verifying the convergence of decentralized adversarial training strategies in heterogeneous environments.}

%




We now proceed with the network analysis. To begin with, we introduce the gradient noise process.
Recall first the maximization problem 
\begin{align}\label{delta_kns}
 \boldsymbol{\delta}_{k,n}^{\star} \in \mathop{\mathrm{argmax}}\limits_{\left\Vert\delta\right\Vert_{p_k}  \le \epsilon_k}   Q_k(\boldsymbol{w}_{k,n-1};\boldsymbol{x}_{k,n} +\delta,\boldsymbol{y}_{k,n})   
\end{align}
In the general non-differentiable case, where the optimal perturbation $\boldsymbol{\delta}_{k,n}^{\star}$ is not necessarily unique, Danskin's theorem guarantees
\begin{align}
    \mathds{E}\nabla_w Q_k(\boldsymbol{w}_{k,n-1};\boldsymbol{x}_{k,n} + \boldsymbol{\delta}_{k,n}^{\star},\boldsymbol{y}_{k,n}) \in \partial_w J_k(\boldsymbol{w}_{k,n-1})
\end{align}
In this case, if the optimal perturbation of (\ref{delta_kns}) is not accessible, we can only use an approximate one, as shown in (\ref{max_k}). The stochastic gradient noise is defined by
\begin{align}\label{d_gn}
     \boldsymbol{s}_{k,n}\overset{\Delta}{=}& \nabla_w Q_k(\boldsymbol{w}_{k,n-1};\widehat{\boldsymbol{x}}_{k,n},\boldsymbol{y}_{k,n})\notag\\
     &- \mathds{E}\nabla_w Q_k(\boldsymbol{w}_{k,n-1};\widehat{\boldsymbol{x}}_{k,n},\boldsymbol{y}_{k,n})
\end{align}
This quantity measures the difference between the stochastic gradient at the approximate maximizer and the expected one over all samples at iteration $n$. Similarly, we define the gradient noise for the more general mini-batch version as
\begin{align}\label{d_gn_B}
  \boldsymbol{s}_{k,n}^{B} \overset{\Delta}{=}&  \frac{1}{B}\sum\limits_{b=1}^{B}\nabla_w Q_k(\boldsymbol{w}_{k,n-1};\widehat{\boldsymbol{x}}_{k,n}^{b},\boldsymbol{y}_{k,n}^{b})\notag\\
  &- \mathds{E}\nabla_w Q_k(\boldsymbol{w}_{k,n-1};\widehat{\boldsymbol{x}}_{k,n},\boldsymbol{y}_{k,n})
\end{align} 
However, if the optimal perturbation $\boldsymbol{\delta}_{k,n}^{\star}$ is accessible and unique, as is verified for some popular (strongly-)convex loss functions in Appendix \ref{ap_deltas}, then it will be used in the proposed algorithms in place of the approximate one. In this case, the functions $f_k(w;\boldsymbol{x}_k, \boldsymbol{y}_k)$ (as defined in (\ref{f_g})) and $J_k(w)$ (as defined in (\ref{J_k_f})) are differentiable. Then, it follows from Danskin's theorem that 
\begin{align}
    & \mathds{E}\nabla_w Q_k(\boldsymbol{w}_{k,n-1};\boldsymbol{x}_{k,n} + \boldsymbol{\delta}_{k,n}^{\star},\boldsymbol{y}_{k,n}) \notag\\
    & =\mathds{E} \nabla_w f_k(\boldsymbol{w}_{k,n-1};\boldsymbol{x}_{k,n},\boldsymbol{y}_{k,n})\notag\\
    &= \nabla_w J_k(\boldsymbol{w}_{k,n-1})
\end{align}
In this case, the stochastic mini-batch gradient noise is defined by the difference between the approximate gradient (represented by the average gradient of the loss over a mini batch of data) and the true
gradient (represented by the gradient of the risk), namely,
\begin{align}\label{dgnb_2}
   \boldsymbol{s}_{k,n}^{B} &\overset{\Delta}{=}  \frac{1}{B}\sum\limits_{b=1}^{B}\nabla_w f_k(\boldsymbol{w}_{k,n-1};{\boldsymbol{x}}_{k,n}^{b},\boldsymbol{y}_{k,n}^{b}) -  \nabla_w J_k(\boldsymbol{w}_{k,n-1})
\end{align} 

Next, we exploit the eigen-structure of $A$ by noting that the $K\times K$ matrix $A$ admits a Jordan canonical decomposition of the form \cite{sayed2014adaptation}:
\begin{equation}\label{decompose}
    A = VJ V^{-1}
\end{equation}
{where the matrices $\{V, J\}$ are}
\begin{equation}\label{vjv}
    V = \left[{\pi}\quad V_{R}\right],\quad
    J = \left[
        \begin{array}{cc}
        1&0\\
        0&J_{\alpha}
        \end{array}
    \right],\quad
    V^{-1} = \left[
    \begin{array}{c}
    \mathbbm{1}^{\sf T}\\
    V_{L}^{\sf T}
    \end{array}
    \right]
\end{equation}
Here, $J_{\alpha}$ is a block Jordan matrix with eigenvalues from the second largest eigenvalue $\lambda_2$ to the smallest  eigenvalue  $\lambda_K$ of $A$ appearing on the diagonal and the positive scalar $\alpha$, which can be chosen arbitrarily small, appearing on the first lower sub-diagonal. Also, for later use, we consider the extended graph matrix:
\begin{equation}
    \mathcal{A} \overset{\Delta}{=} A \otimes I_{M}
\end{equation}
where $I_{M}$ is the identity matrix of size $M$, while $\otimes$ denotes the Kronecker product operation. Then, the extended matrix $\mathcal{A}$ satisfies
\begin{equation}\label{AC_9}
    \mathcal{A} = \mathcal{V}\mathcal{J}\mathcal{V}^{-1}
\end{equation}
where
\begin{align}\label{evjv}
    \mathcal{V} = V\otimes I_M,  \quad \mathcal{J} = J\otimes I_M,\quad \mathcal{ V}^{-1} = V^{-1}\otimes I_M
\end{align}
Since
\begin{align}
    V V^{-1} =  V^{-1}V = I_K
\end{align}
we have
\begin{align}\label{AC_12a}
    &{\pi}\mathbbm{1}^{\sf T} + V_RV_L^{\sf T} = \mathbbm{1}\pi^{\sf T} + V_LV_R^{\sf T} = I_K \notag\\ &V_L^{\sf T}V_R = I_{K-1}\notag\\
    &V_L^{\sf T}\pi = V_R^{\sf T}\mathbbm{1} = 0
\end{align}

Next, we collect the parameters from all agents into the $K\times 1$ block column vector:
\begin{align}\label{AC_o}
    {{\boldsymbol{\scriptstyle\mathcal{W}}}}_{n} \overset{\Delta}{=} \mbox{\rm col}_k \{{\boldsymbol{w}}_{k,n}\}
\end{align}
for $k = 1, \ldots, K$. Similarly, we gather the stochastic gradients from all agents into:
\begin{align}\label{qnhat}
    \boldsymbol{q}_n  \overset{\Delta}{=} \mbox{\rm col}_k\left\{\frac{1}{B}\sum\limits_{b=1}^B\nabla_w Q_k(\boldsymbol{w}_{k,n-1};\widehat{\boldsymbol{x}}_{k,n}^{b},\boldsymbol{y}_{k,n}^b)\right\}
\end{align}
Note that when the optimal perturbation of (\ref{delta_kns}) is accessible and unique, $\boldsymbol{q}_n$ becomes:
\begin{align}\label{qnstar}
    \boldsymbol{q}_n  \overset{\Delta}{=} \mbox{\rm col}_k\left\{\frac{1}{B}\sum\limits_{b=1}^B\nabla_w f_k(\boldsymbol{w}_{k,n-1};\boldsymbol{x}_{k,n},\boldsymbol{y}_{k,n}^b)\right\}
\end{align}

{Using these variables, the recursions of the decentralized adversarial training algorithms listed in Algorithm \ref{al_ro_1} and \ref{al_ro_2} can be rewritten using a unified description as follows:
\begin{align}\label{AC_2}{\boldsymbol{\scriptstyle\mathcal{W}}}_{n} &= \mathcal{A}_2^{\sf T}\left(\mathcal{A}_1^{\sf T}\boldsymbol{\scriptstyle\mathcal{W}}_{n-1} - \mu\boldsymbol{q}_n \right)  
\end{align}
where 
\begin{align}
    \mathcal{A}_2 = A_2\otimes I_M, \quad\mathcal{A}_1 = A_1\otimes I_M,\quad A_1A_2 = A
\end{align}
and the choices for the matrices $A_1$ and $A_2$ depend on the nature of the algorithm. Basically, for the consensus strategy, we set
\begin{align}
    A_1 = A,\quad A_2 = I_K
\end{align}
while for diffusion, it holds that 
\begin{align}
    A_1 = I_K,\quad A_2 = A
\end{align}
Recalling (\ref{comb_matrix}), and since $A_1$ and $A_2$ are either $A$ or $I_K$, they are left-stochastic matrices. We defined $\pi$ to be the right eigenvector corresponding to the eigenvalue 1, namely,
\begin{align}\label{a1a2p}
   A_1 \pi = A_2 \pi = \pi, \quad A_1^{\sf T}\mathbbm{1} = A_2^{\sf T}\mathbbm{1} = 1
    \end{align}}

Multiplying (\ref{AC_2}) from the left by $\mathcal{V}^{\sf T}$, and substituting (\ref{AC_9}) into it, we obtain the following relation:
\begin{align}\label{AC_13}
     \mathcal{V}^{\sf T}{\boldsymbol{\scriptstyle\mathcal{W}}}_{n}  &= \left[\begin{array}{c}
         (\pi^{\sf T} \otimes I_M){\boldsymbol{\scriptstyle\mathcal{W}}}_{n} \\
         (V_R^{\sf T}\otimes I_M){\boldsymbol{\scriptstyle\mathcal{W}}}_{n}
    \end{array}\right] \overset{\Delta}{=} \left[\begin{array}{c}
         \boldsymbol{w}_{c,n}  \\
          \ddot{\boldsymbol{w}}_n
    \end{array}\right] \notag\\
    &= \mathcal{V}^{\sf T}\mathcal{A}^{\sf T}{\boldsymbol{\scriptstyle\mathcal{W}}}_{n-1}  - \mu \mathcal{V}^{\sf T}\mathcal{A}_2^{\sf T }\boldsymbol{q}_n\notag\\
    &\overset{(a)}{=}\mathcal{J}^{\sf T}\mathcal{V}^{\sf T}\boldsymbol{\scriptstyle\mathcal{W}}_{n-1} - \mu\mathcal{V}^{\sf T}\mathcal{A}_2^{\sf T}\boldsymbol{q}_n\notag\\
     &=  \left[\begin{array}{cc}
        I_M & \boldsymbol{0} \\
        \boldsymbol{0} & J_{\alpha}^{\sf T}\otimes I_M
    \end{array}\right] \left[\begin{array}{c}
         \boldsymbol{w}_{c, n-1}  \\
          \ddot{\boldsymbol{w}}_{n-1}
    \end{array}\right] \notag\\
    &\quad\ - \mu \left[\begin{array}{c}
         \pi^{\sf T} \otimes I_M  \\
          V_R^{\sf T}\otimes I_M
    \end{array}\right]\mathcal{A}_2^{\sf T}\boldsymbol{q}_n
\end{align}
where $(a)$ follows from (\ref{AC_9}). Then, it follows from (\ref{a1a2p}) and (\ref{AC_13}) that we can split
(\ref{AC_2}) into the following two parts in terms of the transformed quantities $\{\boldsymbol{w}_{c,n}, \ddot{\boldsymbol{w}}_{n}\}$:
\begin{align}\label{AC_14}
    \boldsymbol{w}_{c,n} = \sum_{k=1}^K \pi_k\boldsymbol{w}_{k,n} = \boldsymbol{w}_{c,n-1} - \mu(\pi^{\sf T} \otimes I_M)\boldsymbol{q}_n
\end{align}
and
\begin{align}\label{AC_15}
    \ddot{\boldsymbol{w}}_{n} = \mathcal{V}_{R}^{\sf T}{\boldsymbol{\scriptstyle\mathcal{W}}}_{n} =  \mathcal{J}_\alpha^{\sf T}\ddot{\boldsymbol{w}}_{n-1} - \mu\mathcal{V}_R^{\sf T}\mathcal{A}_{2}^{\sf T}\boldsymbol{q}_n
\end{align}
where
\begin{align}\label{AC_16}
    \mathcal{J}_\alpha = J_\alpha \otimes I_M,\quad \mathcal{V}_R = V_R\otimes I_M
\end{align}
Note that $\boldsymbol{w}_{c,n}$ represents the network average (also called centroid) as it is the weighted mean of the parameters of all agents. In contrast, $\ddot{\boldsymbol{w}}_n$ is associated with the network disagreement, which quantifies the deviation of local agents from their average. Basically, by extending the centroid $\boldsymbol{w}_{c,n}$ to a block vector ${\boldsymbol{\scriptstyle\mathcal{W}}}_{c,n}$
\begin{align}
  {\boldsymbol{\scriptstyle\mathcal{W}}}_{c,n} \overset{\Delta}{=} (\mathbbm{1}\otimes I_M)\boldsymbol{w}_{c,n}
\end{align}
the relationship of the network disagreement and $\ddot{\boldsymbol{w}}_n$ is given by
\begin{align}\label{net_dis_convex}
    {\boldsymbol{\scriptstyle\mathcal{W}}}_{n} -  {\boldsymbol{\scriptstyle\mathcal{W}}}_{c,n}&= (I_{KM} - (\mathbbm{1}\pi^{\sf T})\otimes I_M){\boldsymbol{\scriptstyle\mathcal{W}}}_{n} \notag\\
    &\overset{(a)}{=} \mathcal{V}_L\mathcal{V}_R^{\sf T}{\boldsymbol{\scriptstyle\mathcal{W}}}_{n} \overset{(b)}{=} \mathcal{V}_L\ddot{w}_n
\end{align}
where $(a)$ follows from (\ref{AC_12a}), and $(b)$ follows from \eqref{AC_15}. We further substitute the gradient noise term defined in (\ref{qnhat}) into (\ref{AC_14}), and obtain the following recursion in general:
\begin{align}\label{AC_14_2}
 \boldsymbol{w}_{c,n} =& \boldsymbol{w}_{c,n-1} - \mu\sum_k\pi_k\mathds{E}\nabla_w Q_k(\boldsymbol{w}_{k,n-1};\widehat{\boldsymbol{x}}_{k,n}^{b},\boldsymbol{y}_{k,n}^b)  \notag\\
 &- \mu\sum_k \pi_k \boldsymbol{s}_{k,n}^{B}
\end{align}
As for the case of the obtainable optimal perturbation, we substitute (\ref{qnstar}) into (\ref{AC_14}), and obtain
\begin{align}\label{AC_14_2_opt}
 \boldsymbol{w}_{c,n} = \boldsymbol{w}_{c,n-1} - \mu\sum_k\pi_k\nabla_w J_k(\boldsymbol{w}_{k,n-1})- \mu\sum_k \pi_k \boldsymbol{s}_{k,n}^{B}
\end{align}

As discussed before, traditional analyses of decentralized learning algorithms  rely heavily on the Lipschitz condition over the gradients of risk or loss functions. However, in our adversarial setting, where the gradient of risk functions is only guaranteed to be affine Lipschitz, the traditional analyses are no longer applicable. Motivated by \cite{ying2018performance2}, we adopt a different framework to verify the convergence of the proposed decentralized adversarial training algorithms. To this end, we first verify the convergence of the term $\ddot{\boldsymbol{w}}_{n}$ for later use, for which the proof can be found from Appendix \ref{ap_w_dd}.

\begin{theorem}(\textbf{Mean-square stability of network disagreement in (strongly)-convex environments.})\label{thddw}
    Consider a network of $K$ agents running the decentralized adversarial training algorithms covered by recursion (\ref{AC_2}). Under Assumptions \ref{as1}--\ref{as4}, for sufficiently small step-sizes $\mu$, and after enough iterations $n\ge \ddot{n}$, it holds that
    \begin{align}
    \mathds{E}\Vert\ddot{\boldsymbol{w}}_{n} \Vert ^2 \le O(\mu^2)
    \end{align}
    with 
    \begin{align}
      \ddot{n} = O\left(\frac{\log \mu}{\log (\lambda_2+O(\mu^2))}\right)  
    \end{align}
    Moreover, the average mean-square distance between the network centroid and local agents will be bounded by 
    \begin{align}\label{th_wkc}
        \mathds{E}\Vert {\boldsymbol{\scriptstyle\mathcal{W}}}_{n} - {\boldsymbol{\scriptstyle\mathcal{W}}}_{c,n} \Vert ^2 \le O(\mu^2)
    \end{align}
 $\hfill\square$
\end{theorem}
The above theorem implies that, with sufficiently small step sizes, recursion (\ref{AC_15}) will be upper bounded by a small value after enough iterations. Moreover, all agents will converge to the network centroid in the long run. Importantly, the proof of Theorem~\ref{thddw} does not rely on convexity or strong convexity, making it applicable to both convex and strongly convex scenarios. Furthermore, as shown in Appendix~\ref{ap_w_dd}, Theorem~\ref{thddw} holds regardless of whether the optimal perturbations or approximate ones are used in the algorithms.

{\subsection{Convergence analysis for the strongly-convex case}}

{In this section, we examine the convergence of the proposed decentralized adversarial training algorithms in the strongly-convex environment.} {\begin{assumption}\label{as3}
  (\textbf{Strong convexity}) For each agent k, the loss function $Q_k(w;\cdot)$ is strongly-convex over $w$, namely, for any $w_1$ and $w_2$, it holds that
 \begin{align}\label{asconvex}
 Q_k(w_2;x,y) \ge \; & Q_k(w_1;x,y) + \nabla_{w^{\sf T}} Q_k(w_1;x,y)(w_2 - w_1)  \notag\\
 &+ \frac{\nu}{2}\Vert w_2 - w_1\Vert^2
 \end{align}
 where $\nu > 0$.
 
 $\hfill\square$
\end{assumption}

We remark that it follows from Danskin's theorem \cite{bertsekas2009convex, lin2020gradient, rockafellar2015convex, Thekumparampil019} that the maximization over a strongly-convex function remains strongly-convex.  One proof for this statement can be found in section 6.11 in \cite{bertsekas2009convex} and section B.3 in \cite{Thekumparampil019}. Basically, when $Q_k(w;\cdot,\cdot)$ is $\nu-$strongly convex over $w$, then its maximization function over the perturbation variable, i.e., $f_k(w;\cdot,\cdot)$, is also $\nu-$strongly convex. As a result, the adversarial risk $J_k(w)$ defined by \eqref{jk} and the aggregate risk $J(w)$ in (\ref{formu}) will be $\nu-$strongly convex as well.

{Assume $w^{\star}$ is the global minimum of $J(w)$, and introduce the error vector
\begin{align}
   \widetilde{\boldsymbol{w}}_{k,n-1} \overset{\Delta}{=} w^\star - {\boldsymbol{w}}_{k,n-1} 
\end{align}
which measures the difference between the model of agent $k$ and the global minimizer at iteration $n$. The following result establishes some useful properties for the gradient noise process defined in (\ref{d_gn_B}) and (\ref{dgnb_2}) in strongly-convex case, namely, it has zero mean and bounded second-order moment (conditioned on past history), for which proofs can be found in Appendix \ref{ap2}.}
\begin{lemma}\label{p_gn}
(\textbf{Moments of gradient noise in the strongly-convex environment}) For each agent k, the gradient noise defined in  (\ref{d_gn_B}) and (\ref{dgnb_2}) have zero mean and their variance is bounded:
\begin{align}
   &\label{s0_convex} \mathds{E}\left\{\boldsymbol{s}_{k,n}^{B }|\boldsymbol{\mathcal{F}}_{n-1}\right\}=0 \\
    &\label{s2_convex} \mathds{E}\left\{\left\Vert \boldsymbol{s}_{k,n}^{B }\right\Vert^2|\boldsymbol{\mathcal{F}}_{n-1}\right\} \le \beta_{k}^2 \left\Vert \widetilde{\boldsymbol{w}}_{k,n-1} \right\Vert^2 + \sigma_{k}^2
\end{align}
for some non-negative scalars given by
\begin{align}
    &\beta_{k}^2 = \frac{16L^2}{B}  \notag\\
    &\sigma_{k}^2  = \frac{8}{B}\mathds{E}\Vert \nabla_w Q_k(w^\star;\boldsymbol{x}_{k,n}+\boldsymbol{\delta}_k^\star(w^\star),\boldsymbol{y}_{k,n})\Vert^2 + O(\epsilon^2)
\end{align} where 
\begin{align}
\boldsymbol{\delta}_k^\star(w^\star) = \mathop{\rm{argmax}}\limits_{\left\Vert{\delta}\right\Vert_{p_k}  \le \epsilon_k}   Q_k(w^\star;\boldsymbol{x}_{k,n} +{\delta},\boldsymbol{y}_{k,n})
\end{align} The quantity
 $\boldsymbol{\mathcal{F}}_{n-1}$ is the filtration generated by the past history of the random process $\mbox{{\rm{col}}} {\{\boldsymbol{w}_{k,j}\}_{k=1}^K}$ for $j\le n-1$.
 
$\hfill\square$
\end{lemma}

To proceed, we collect the error vectors from all agents into the $K \times 1$ block column vector:
\begin{align}\label{AC_1}
{\widetilde{\boldsymbol{\scriptstyle\mathcal{W}}}}_{n} \overset{\Delta}{=} {\mbox{\rm col}}\{\widetilde{\boldsymbol{w}}_{k,n}\}_{k=1}^{K}
\end{align}
{with which we obtain a unified recursion for the decentralized adversarial training algorithms listed in Algorithm \ref{al_ro_1} and \ref{al_ro_2}:
\begin{align}\label{re_w_t}
\widetilde{\boldsymbol{\scriptstyle\mathcal{W}}}_{n} &= \mathcal{A}_2^{\sf T}\left(\mathcal{A}_1^{\sf T}\widetilde{\boldsymbol{\scriptstyle\mathcal{W}}}_{n-1} + \mu\boldsymbol{q}_n \right)    
\end{align}
Then, similar to (\ref{AC_13}), we obtain the following recursion for ${\widetilde{\boldsymbol{\scriptstyle\mathcal{W}}}}_{n}$:}
\begin{align}\label{AC_13_strong}
     &\mathcal{V}^{\sf T}{\widetilde{\boldsymbol{\scriptstyle\mathcal{W}}}}_{n}  = \left[\begin{array}{c}
         (\pi^{\sf T} \otimes I_M){\widetilde{\boldsymbol{\scriptstyle\mathcal{W}}}}_{n} \\
         (V_R^{\sf T}\otimes I_M){\widetilde{\boldsymbol{\scriptstyle\mathcal{W}}}}_{n}
    \end{array}\right] \overset{\Delta}{=} \left[\begin{array}{c}
         \bar{\boldsymbol{w}}_n \\
          \check{\boldsymbol{w}}_n
    \end{array}\right] \notag\\
    &=  \left[\begin{array}{cc}
        I_M & \boldsymbol{0} \\
        \boldsymbol{0} & J_{\alpha}^{\sf T}\otimes I_M
    \end{array}\right] \left[\begin{array}{c}
         \bar{\boldsymbol{w}}_{n-1} \\
          \check{\boldsymbol{w}}_{n-1}
    \end{array}\right] + {\mu \left[\begin{array}{c}
         \pi^{\sf T} \otimes I_M  \\
          V_R^{\sf T}\otimes I_M
    \end{array}\right]\mathcal{A}_2^{\sf T}\boldsymbol{q}_n}
\end{align}
It follows {from \eqref{qnhat}} that we can split (\ref{AC_13_strong}) into the following two recursions in terms of the transformed quantities $\{\bar{\boldsymbol{w}}_n, \check{\boldsymbol{w}}_n\}$:
\begin{align}
\label{w_bar_1}
    \bar{\boldsymbol{w}}_n &= w^{\star} - \boldsymbol{w}_{c,n}\notag\\ &= \bar{\boldsymbol{w}}_{n-1} + \frac{\mu}{B}\sum_{k=1}^K \pi_k \sum\limits_{b=1}^B\nabla_w Q_k(\boldsymbol{w}_{k,n-1};\widehat{\boldsymbol{x}}_{k,n}^{b},\boldsymbol{y}_{k,n}^b) \\
\label{w_check}
    \check{\boldsymbol{w}}_n &= \mathcal{V}_{R}^{\sf T}\widetilde{{\boldsymbol{\scriptstyle\mathcal{W}}}}_{n} =  \mathcal{J}_\alpha^{\sf T}\check{\boldsymbol{w}}_{n-1} + \mu\mathcal{V}_R^{\sf T}\mathcal{A}_{2}^{\sf T}\boldsymbol{q}_n
\end{align}
Substituting (\ref{d_gn_B}) into (\ref{w_bar_1}), we further obtain:
\begin{align}
\label{w_bar_2}
    \bar{\boldsymbol{w}}_n =& w^{\star} - \boldsymbol{w}_{c,n}\notag\\
    =& \bar{\boldsymbol{w}}_{n-1} + \sum_k\pi_k\mathds{E}\nabla_w Q_k(\boldsymbol{w}_{k,n-1};\widehat{\boldsymbol{x}}_{k,n},\boldsymbol{y}_{k,n}) \notag\\
    &+ \mu\sum_k\pi_k\boldsymbol{s}_{k,n}^{B}
\end{align}
Moreover, the vector $\check{\boldsymbol{w}}_n$ satisfies
\begin{align}\label{check_dd_w}
   \check{\boldsymbol{w}}_n =  \mathcal{V}_{R}^{\sf T}\widetilde{{\boldsymbol{\scriptstyle\mathcal{W}}}}_{n} = \mathcal{V}_{R}^{\sf T}(\mathbbm{1}\otimes w^{\star} - {\boldsymbol{\scriptstyle\mathcal{W}}}_{n}) \overset{(a)}{=} -\mathcal{V}_{R}^{\sf T}{\boldsymbol{\scriptstyle\mathcal{W}}}_{n} = -\ddot{\boldsymbol{w}}_n
\end{align}
where $(a)$ follows from the last equality of (\ref{AC_12a}).

The convergence results for {(\ref{re_w_t}), (\ref{w_bar_1}), and (\ref{w_check}) } are stated next, for which proofs can be found in Appendix \ref{ap_th_mse}.
\begin{theorem}\label{th_mse}
(\textbf{Network mean-square-error stability in the strongly-convex environment}.) Consider a network of $K$ agents running the decentralized adversarial training algorithms from {Algorithms \ref{al_ro_1} and \ref{al_ro_2}}. Under Assumptions \ref{as1}--\ref{as3}, and for sufficiently small step-sizes $\mu$, if the approximate perturbations are used, then after enough iterations $n \ge \bar{n}$, it holds that
\begin{align}
 \mathds{E}\left\Vert \check{\boldsymbol{w}}_{n}\right\Vert^2 \le O(\mu^2),\quad\; \mathds{E}\left\Vert \bar{\boldsymbol{w}}_{n}\right\Vert^2 \le  {O(\mu) + O(\epsilon)}
\end{align}
with
\begin{align}
    \bar{n} =  O\left(\frac{\log \mu}{\log (1 - O(\mu))}\right)
\end{align}
More specifically, the network converges asymptotically to a small neighborhood of the global minimizer $w^{\star}$ at an exponential rate:
\begin{align}\label{th_mse_e_1}
  \mathop{\lim \sup}\limits_{n\to\infty} \mathds{E}\left\Vert {\widetilde{\boldsymbol{\scriptstyle\mathcal{W}}}}_{n}\right\Vert^2 
    \le {O(\lambda^n) + O(\mu) +O(\epsilon)}
\end{align}
with the rate
\begin{align}\label{th_mse_e_2}
    \lambda = {1 - 2\nu\mu  + O(\mu^2) + O(\mu\epsilon)}
\end{align}
{where the $O(\epsilon)$ term  in (\ref{th_mse_e_1}) and the $O(\mu\epsilon)$ term in (\ref{th_mse_e_2}) arise due to the approximation error when solving the maximization problem over perturbations.}

$\hfill\square$
\end{theorem}
{
Note that the $O(\epsilon)$ term  in (\ref{th_mse_e_1}) and the $O(\mu\epsilon)$ term in (\ref{th_mse_e_2}) reflect the approximation error arising from the maximization problem over the perturbation variable $\delta$. Thus, the better the  maximization problem (\ref{maximizer_t}) is solved, the smaller the two terms will be.} Theorem \ref{th_mse} indicates that the proposed algorithms enable the network to approach in the mean-square-error sense an {$O(\mu) + O(\epsilon)$}-neighborhood of the robust minimizer $w^{\star}$ after enough iterations, so that the average worst-case performance over all possible perturbations in the small region bounded by $\epsilon_k$ is effectively minimized.

{If the optimal perturbation of (\ref{maximizer_t}) is unique and available, substituting (\ref{dgnb_2}) into (\ref{w_bar_1}) gives:}
\begin{align}
\label{w_bar_exact}
    \bar{\boldsymbol{w}}_n & = \bar{\boldsymbol{w}}_{n-1} + \sum_k\pi_k\nabla_w J_k(\boldsymbol{w}_{k,n-1})+ \sum_k\pi_k\boldsymbol{s}_{k,n}^{B}
\end{align}
{In this case, the convergence results for (\ref{re_w_t}), (\ref{w_bar_exact}), and (\ref{w_check}) are stated in the following, for which proofs can be found in Appendix \ref{ap_co_mse}.}

\begin{corollary}\label{co_mse}
Consider a network of $K$ agents running the decentralized adversarial training algorithms from Algorithms \ref{al_ro_1} and \ref{al_ro_2}. Under Assumptions \ref{as1}--\ref{as3} and for sufficiently small step-sizes $\mu$, if the true optimal perturbation of (\ref{maximizer_t}) is accessible and unique, then after enough iterations $n \ge \bar{n}$, it holds that 
\begin{align}
\mathds{E}\left\Vert \check{\boldsymbol{w}}_{n}\right\Vert^2 \le O(\mu^2),\quad\; \mathds{E}\left\Vert \bar{\boldsymbol{w}}_{n}\right\Vert^2 \le  O(\mu) 
\end{align}
Specifically, the network converges asymptotically to a small neighborhood of the global minimizer $w^{\star}$ at an exponential rate:
\begin{align}\label{co_1_e_1}
  \mathop{\lim \sup}\limits_{n\to\infty} \mathds{E}\left\Vert {\widetilde{\boldsymbol{\scriptstyle\mathcal{W}}}}_{n}\right\Vert^2 
    \le O(\lambda^n) + O(\mu) 
\end{align}
with the rate
\begin{align}\label{co_e2}
    \lambda = 1 - 2\nu\mu  + O(\mu^2)
\end{align} 
$\hfill\square$
\end{corollary}

The above corollary indicates that, if the true optimal perturbation in \eqref{maximizer_t} is accessible and unique (a condition satisfied by several popular strongly-convex functions), then after a sufficient number of iterations, the proposed algorithms enable the network to converge in the mean-square-error sense to an $O(\mu)$-neighborhood of the robust minimizer $w^{\star}$. %

\subsection{Convergence analysis for the convex case}
In this section, we analyze the convergence of the proposed decentralized adversarial training algorithms in the convex environment. Specifically, we focus on the scenario where the optimal perturbation in (\ref{maximizer_t}) is accessible and unique, which ensures that $f_k(w;\boldsymbol{x}_k, \boldsymbol{y}_k)$, as defined in (\ref{f_g})) and viewed as the ``real" loss function, is differentiable. For the more general non-differentiable case, we will combine its analysis with the non-convex case, to be discussed later in the subsequent section.

{\begin{assumption}\label{as5_convex}
  (\textbf{Convexity}) For each agent k, the loss function $Q_k(w;\cdot)$ is convex over $w$, namely, for any $w_1$ and $w_2$, it holds that
 \begin{align}\label{asconvex5}
 Q_k(w_2;x,y) \ge \; & Q_k(w_1;x,y) + \nabla_{w^{\sf T}} Q_k(w_1;x,y)(w_2 - w_1)
 \end{align}
 $\hfill\square$
\end{assumption}}
{Again, it follows from Danskin's theorem that the maximization over a convex function remains convex. Therefore, the ``real" loss function $f_k(w;\cdot,\cdot)$ in \eqref{f_g}, the adversarial risk $J_k(w)$ defined by \eqref{jk}, and the aggregate risk $J(w)$ in \eqref{formu} are convex as well.}

{
We now present the following important Lemma, and the associated proofs can be found from Appendix~\ref{ap_lip_convex}.}
{
\begin{lemma}\label{lip_convex}
For the differentiable function $f_k(w;\cdot)$, under the affine Lipchitz condition in (\ref{affine_l_2}), for any $w_1,w_2\in\mathbb{R}^M$, it holds that:
\begin{align}\label{u_affine}
    f_k(w_2;\cdot) - f_k(w_1;\cdot) \le& \nabla_{w^{\sf T}} f_k(w_1;\cdot)(w_2-w_1) \notag\\
    &+ \frac{L}{2}\Vert w_2 - w_1\Vert^2 + O(\epsilon)\Vert w_2 - w_1\Vert
\end{align}
Moreover, with the convexity of $f_k$, it holds that
\begin{align}\label{d_g_f_k}
&\frac{1}{2(L+1)}\Vert \nabla_w f_k(w_2;\cdot) - \nabla_w f_k(w_1;\cdot) \Vert^2 \notag\\
&\le f_k(w_1;\cdot) - f_k(w_2;\cdot) - \nabla_{w^{\sf T}} f_k(w_2;\cdot)(w_1 - w_2 ) + O(\epsilon^2)
\end{align}
$\hfill\square$
\end{lemma}}
{
Compared to the traditional results where the loss function is $L-$smooth \cite{sayed_2023}, expression (\ref{u_affine}) indicates that the extra constant term in the affine Lipschitz condition implicitly introduces an additional term to the upper bound of $f_k(w_2;\cdot) - f_k(w_1;\cdot)$. Moreover, assume $w^{\star}$ is a global minimum of $J(w)$, resorting to (\ref{d_g_f_k}) gives
the following lemma, for which proofs can be found in Appendix~\ref{ap_b_gconvex}.}
\begin{lemma}\label{b_gconvex}
 Under the convexity of $f_k$ and (\ref{affine_l_2}), and for any $\boldsymbol{w} \in \mathcal{F}_{n-1}$, it holds that
 \begin{align}
     &\mathds{E}\Big\{\Vert \nabla_w f_k(\boldsymbol{w};\boldsymbol{x}_k,\boldsymbol{y}_k)\Vert^2 \big\vert \mathcal{F}_{n-1}\Big\}\notag\\
     &\le 4(L+1)\big(J_k(\boldsymbol{w}) - J_k(w^{\star}) - \nabla_{w^{\sf T}} J_k(w^{\star})(\boldsymbol{w} - w^{\star})\big) + \gamma_k^2 
 \end{align}
 where the expectation is taken over the random data, and
 \begin{align}
     \gamma_k^2 =  O(\epsilon^2) + 2\mathds{E}\Vert\nabla_w f_k(w^{\star};\boldsymbol{x}_k,\boldsymbol{y}_k)\Vert^2
 \end{align}
 $\hfill\square$
\end{lemma} 

We next establish some useful properties for the gradient noise, as defined in (\ref{dgnb_2}), in the convex case, for which proofs can be founded in Appendix \ref{ap_sg_convex}.
\begin{lemma}\label{sg_convex}
(\textbf{Moments of gradient noise in the convex environment}) For each agent k, the gradient noise defined in (\ref{dgnb_2}) is zero-mean and its variance is bounded:
\begin{align}
\label{s0_convex_2} 
&\mathds{E}\left\{\boldsymbol{s}_{k,n}^{B }|\boldsymbol{\mathcal{F}}_{n-1}\right\}=0 \\
   \label{s2_convex_2} 
&\mathds{E}\left\{\left\Vert \boldsymbol{s}_{k,n}^{B }\right\Vert^2|\boldsymbol{\mathcal{F}}_{n-1}\right\} \notag\\
&\le   \frac{32(L+1)}{B}\Big( J_k(\boldsymbol{w}_{c,n-1}) - J_k(w^{\star}) \notag\\
&\quad\;- \nabla_{w^{\sf T}} J_k(w^\star)(\boldsymbol{w}_{c,n-1} - w^\star)\Big) + \frac{8\gamma_k^2}{B} + O(\epsilon^2) \notag\\
&\quad\;+ O(\Vert\boldsymbol{w}_{k,n-1} - \boldsymbol{w}_{c,n-1}\Vert^2)
\end{align}
$\hfill\square$
\end{lemma}

{Recalling recursions (\ref{AC_14_2_opt}) and (\ref{AC_15}), we next state the convergence of the proposed decentralized adversarial training algorithms in the convex environment, for which proofs can be found in Appendix \ref{ap_th_j_convex}. }
{
\begin{theorem}\label{th_j_convex}
(\textbf{Network excess-risk stability in the convex environment}) Consider a network of $K$ agents running the decentralized adversarial training Algorithms \ref{al_ro_1} and \ref{al_ro_2}. Under Assumptions \ref{as1}--\ref{as4} and \ref{as5_convex}, and for sufficiently small step-sizes $\mu$, the following bounds holds for the network centroid: 
 \begin{align}\label{w_c_convex}
     \frac{1}{N}\sum_{n=0}^{N-1} \mathds{E}J(\boldsymbol{w}_{c,n}) - J(w^{\star}) \le  O(\frac{1}{\mu N})+ O(\mu)  
 \end{align}
Moreover, for any local agent k, a similar bound holds:
  \begin{align}\label{w_k_convex}
     \frac{1}{N}\sum_{n=0}^{N-1} \mathds{E}J(\boldsymbol{w}_{k,n}) - J(w^{\star}) \le  O(\frac{1}{\mu N})+ O(\mu)  
 \end{align}
 $\hfill\square$
\end{theorem}}

{
Theorem~\ref{th_j_convex} verifies the convergence of the proposed algorithms from two  perspectives. First, by combining (\ref{th_wkc}) and (\ref{w_c_convex}), we conclude that, after a sufficient number of iterations, all local models will cluster around the network centroid, which in turn will converge to a small neighborhood of a local minimum of $J(w)$ at the rate of $O(\frac{1}{\mu N})$. Second, the bound in (\ref{w_k_convex}) guarantees that each individual agent also converges to a a small neighborhood of a local minimum of $J(w)$ at the same rate of $O(\frac{1}{\mu N})$.}

\section{Analysis in non-convex environments}
For nonconvex losses, it is generally not possible to obtain closed-form expressions for the maximizers in (\ref{maximizer_t}). {This motivates the use of the approximate maximizers in practice. Moreover, the maximizers in \eqref{maximizer_t} are not necessarily unique. In such cases, as discussed in Section~\ref{sec2}, the ``real" loss function $f_k(w;\cdot)$ can be non-differentiable in general. As a result, the local risk functions $J_k(w)$ and the global risk function $J(w)$ are not differentiable as well. Therefore, we analyze the convergence of the proposed decentralized adversarial training algorithms, as listed in Algorithms \ref{al_ro_1} and \ref{al_ro_2}, using the subdifferential of the risk functions. To do so, we first assume the following conditions concerning non-convex loss functions.}

First, we suppose that the expected norm of the gradient of $Q_k(\cdot)$ over $w$ is bounded, which is common in non-convex optimization \cite{lin2020gradient, kayaalp2022dif, sinha2017certifying}, {especially for non-smooth loss functions \cite{DavisD19, jin2020local, garrigo}}.
\begin{assumption}\label{as5}
(\textbf{Bounded gradients}) For each agent k, the expectation of the norm of the gradient $\nabla_w Q_k(w;\boldsymbol{x}_k+{{\delta}}_k,\boldsymbol{y}_k)$ is bounded, namely, for any $w \in \mathcal{\mathbbm{R}}^M$:
\begin{equation}\label{b1}
    \mathds{E}\left\Vert\nabla_w  Q_k(w;\boldsymbol{x}_k+\delta,\boldsymbol{y}_k)\right\Vert \le G 
\end{equation}
with $\Vert\delta\Vert_{p_k} \le \epsilon_k$.

$\hfill\square$
\end{assumption}

Then, similar to the convex case, we leverage the gradient noise to facilitate the analysis of the proposed algorithms in the nonconvex setting. Recalling the definition of the gradient noise in \eqref{d_gn} and \eqref{d_gn_B}, we note that their expressions remain the same in the nonconvex case. To distinguish between the convex and nonconvex settings, we use different notation:
 \begin{align}\label{dgn}
\bar{\boldsymbol{s}}_{k,n}&\overset{\Delta}{=} \nabla_w Q_k(\boldsymbol{w}_{k,n-1};\widehat{\boldsymbol{x}}_{k,n},\boldsymbol{y}_{k,n})\notag\\
&\quad\;- {\mathds{E}\nabla_w Q_k(\boldsymbol{w}_{k,n-1};\widehat{\boldsymbol{x}}_{k,n},\boldsymbol{y}_{k,n})}
\end{align}
\begin{align}\label{bbs}
  \bar{\boldsymbol{s}}_{k,n}^{B} \overset{\Delta}{=}&  \frac{1}{B}\sum\limits_{b=1}^{B}\nabla_w Q_k(\boldsymbol{w}_{k,n-1};\widehat{\boldsymbol{x}}_{k,n}^{b},\boldsymbol{y}_{k,n}^{b})\notag\\
  &- \mathds{E}\nabla_w Q_k(\boldsymbol{w}_{k,n-1};\widehat{\boldsymbol{x}}_{k,n},\boldsymbol{y}_{k,n})
\end{align}
The proof for \eqref{s0_convex} is still valid in the nonconvex setting. That is, the expectation of the stochastic gradient noise conditioned on the past iterates $\mathcal{F}_{n-1}$ is zero, namely,
\begin{align}\label{s0}
    &\mathds{E}\left\{\bar{\boldsymbol{s}}_{k,n}^{B}|\boldsymbol{\mathcal{F}}_{n-1}\right\}  = 0
\end{align}
However, unlike the convex case, we now assume that the variance of the stochastic gradient noise is bounded by a constant in the nonconvex setting, which can be verified with Assumption \ref{as5}\cite{sinha2017certifying}. 

\begin{assumption}\label{as6}
(\textbf{Moments of gradient noise})
{For each agent $k$, the gradient noise satisfies
\begin{align}
    \label{s2} \mathds{E}\left\{\left\Vert\bar{\boldsymbol{s}}_{k,n}\right\Vert^2|\boldsymbol{\mathcal{F}}_{n-1}\right\} \le  \bar{\sigma}^2
\end{align}
for a non-negative scalar $\bar{\sigma}$.}

$\hfill\square$
\end{assumption}
  
{By following a similar method used to prove (\ref{sb2}), we can verify that the variance of the mini-batch stochastic gradient noise is inversely proportional to the batch size $B$, namely, it holds that:
\begin{align}
    \label{sb2_n}
    \mathds{E}\left\{\left\Vert\bar{\boldsymbol{s}}^{B}_{k,n}\right\Vert^2|\boldsymbol{\mathcal{F}}_{n-1}\right\}
    {\le} \frac{\bar{\sigma}^2}{B}
\end{align}}

Now we consider the weighted average of the weight iterates at all agents defined by
\begin{equation}\label{cluster}
   \boldsymbol{w}_{c,n} \overset{\Delta}{=} \sum\limits_{k=1}^K \pi_k\boldsymbol{w}_{k,n}
\end{equation}
and introduce the following block vectors:
\begin{align}
    &{\boldsymbol{\scriptstyle\mathcal{W}}}_{n} \overset{\Delta}{=} {\mbox{\rm col}}\{\boldsymbol{w}_{k,n}\}_{k=1}^{K}\\
& {\boldsymbol{\scriptstyle\mathcal{W}}}_{c,n} \overset{\Delta}{=} \mathbbm{1}_{K}\otimes\boldsymbol{w}_{c,n}\\
    \label{dq}
    &{\mathcal{G}_{n-1}\overset{\Delta}{=} {\mbox{\rm col}}\left\{\mathds{E}\nabla_w Q_k(\boldsymbol{w}_{k,n-1};\widehat{\boldsymbol{x}}_{k,n},\boldsymbol{y}_{k,n})\right\}} \\
    \label{s_n_bar}
    &\bar{\boldsymbol{s}}_n^B \overset{\Delta}{=} {\mbox{\rm col}} \left\{\bar{\boldsymbol{s}}_{k,n}^B\right\}
\end{align}
    {Then, the proposed algorithms listed in (\ref{d_d})--(\ref{d_combine_e}) and \eqref{d_consen}--\eqref{combine_e_consen} can be rewritten using the following unified recursion:
\begin{equation}\label{r_non_c}
    \boldsymbol{\scriptstyle\mathcal{W}}_{n} = \mathcal{A}^{\sf T}\boldsymbol{\scriptstyle\mathcal{W}}_{n-1} - \mu\mathcal{A}_2^{\sf T}\mathcal{G}_{n-1} - \mu\mathcal{A}_2^{\sf T}\bar{\boldsymbol{s}}_n^B 
\end{equation}
where similar to the convex case, for the diffusion adversarial training algorithm, as listed in (\ref{d_d})--(\ref{d_combine_e}), it holds that
\begin{align}
    \mathcal{A}_1 = I_{KM}, \quad \mathcal{A}_2 = \mathcal{A}
\end{align}
while for the consensus adversarial training algorithm, as listed in \eqref{d_consen}--\eqref{combine_e_consen}, we have
\begin{align}
    \mathcal{A}_1 = \mathcal{A}, \quad \mathcal{A}_2 = I_{KM}
\end{align}
We then have the next statements to measure the difference between $\boldsymbol{\scriptstyle\mathcal{W}}_{n}$ and $\boldsymbol{\scriptstyle\mathcal{W}}_{c,n}$.  The associated proofs can be found in Appendix \ref{ap4}.}
\begin{lemma}\label{lemma3}
(\textbf{Network disagreement}) Under Assumptions \ref{as1}--\ref{as2} and \ref{as5}--\ref{as6}, and with sufficiently small step size $\mu$, the {mean-square} disagreement between $\boldsymbol{\scriptstyle\mathcal{W}}_{n}$ and $\boldsymbol{\scriptstyle\mathcal{W}}_{c,n}$ can be bounded after enough iterations $n_0$, namely, it holds that:
\begin{align}\label{nd}
{\mathds{E}}\left\Vert\boldsymbol{\scriptstyle\mathcal{W}}_{n} - \boldsymbol{\scriptstyle\mathcal{W}}_{c,n}\right\Vert^2 \le  O(\mu^2)
\end{align}
for
\begin{align}
       n\ge n_0 \ge O \left(\frac{\rm{log} \ \mu}{\rm{log}\ \textit{t}}\right) 
\end{align}
where $t =  \left\Vert J_\alpha^{\sf T} \right\Vert <1$.

$\hfill\square$
\end{lemma}

The above result extends Lemma 1 from \cite{vlaski2021distributed,vlaski2021distributed2,kayaalp2022dif}, which studies the unperturbed environment without adversarial observations. Result (\ref{nd}) means that $\boldsymbol{w}_{k,n}$ will cluster around the centroid $\boldsymbol{w}_{c,n}$ after sufficient iterations, and the bounded perturbations do not ruin this property. 

In order to progress further with the network analysis in the non-convex environment, we write down the following recursion for the centroid defined in \eqref{cluster}:
\begin{align}\label{wc}
     \boldsymbol{w}_{c,n} & =(\pi^{\sf T}\otimes I_M)\boldsymbol{\scriptstyle\mathcal{W}}_{n} \notag\\
     &\overset{(a)}{=}(\pi^{\sf T} \otimes I_M)\boldsymbol{\scriptstyle\mathcal{W}}_{n-1} - \mu(\pi^{\sf T}\otimes I_M)(\mathcal{G}_{n-1} +\bar{\boldsymbol{s}}_n^B) \notag\\
     & = \boldsymbol{w}_{c,n-1} - \mu\sum\limits_{k=1}^{K} \pi_k {\mathds{E}\nabla_w Q_k(\boldsymbol{w}_{k,n-1};\widehat{\boldsymbol{x}}_{k,n},\boldsymbol{y}_{k,n})} \notag\\
     &\quad\;- \mu \sum\limits_{k=1}^{K} \pi_k\bar{\boldsymbol{s}}^{B}_{k,n}
\end{align}
where $(a)$ follows from \eqref{r_non_c} and \eqref{a1a2p}. 
We can rework this recursion by using $\boldsymbol{w}_{c,n-1}$ as the argument for {$\mathds{E}\nabla_w Q_k(\cdot;\widehat{\boldsymbol{x}}_{k,n},\boldsymbol{y}_{k,n})$} instead of $\boldsymbol{w}_{k,n-1}$. In this way, we can rewrite (\ref{wc}) in the form
\begin{align}\label{wce}
     \boldsymbol{w}_{c,n} =& \boldsymbol{w}_{c,n-1} -\mu\sum\limits_k \pi_k {\mathds{E}\nabla_w Q_k(\boldsymbol{w}_{c,n-1};\widehat{\boldsymbol{x}}_{k,n},\boldsymbol{y}_{k,n})} \notag\\
     &- \mu\boldsymbol{d}_{n-1} - \mu\widehat{\boldsymbol{s}}_n^B
\end{align}
where we are introducing
\begin{align}
    \label{d_n}
     &\boldsymbol{d}_{n-1} \overset{\Delta}{=}  \sum\limits_{k=1}^{K} \pi_k \Big(\mathds{E}\nabla_w Q_k(\boldsymbol{w}_{k,n-1};\widehat{\boldsymbol{x}}_{k,n},\boldsymbol{y}_{k,n})\notag\\
     &\quad\quad\quad\;-\mathds{E}\nabla_w Q_k(\boldsymbol{w}_{c,n-1};\widehat{\boldsymbol{x}}_{k,n},\boldsymbol{y}_{k,n})\Big)\\
    \label{sn}
     &\widehat{\boldsymbol{s}}_{n}^B \overset{\Delta}{=} \sum\limits_{k=1}^{K} \pi_k \bar{\boldsymbol{s}}^{B}_{k,n}
\end{align}
The vector $\boldsymbol{d}_{n-1}$ arises from the disagreement between the local models $\boldsymbol{w}_{k,n-1}$ and their centroid $\boldsymbol{w}_{c,n-1}$, while $\widehat{\boldsymbol{s}}_{n}^B$ aggregates the gradient noise across the network. {Moreover, in the auxillary term $\mathds{E}\nabla_w Q_k(\boldsymbol{w}_{c,n-1};\widehat{\boldsymbol{x}}_{k,n},\boldsymbol{y}_{k,n})$, the gradient of $Q_k(\cdot)$ is evaluated at $\boldsymbol{w}_{c,n-1}$, while the perturbation used in $\widehat{\boldsymbol{x}}_{k,n}$ is still computed from $\boldsymbol{w}_{k,n-1}$.}  It can be verified that the two error terms $\boldsymbol{d}_{n-1}$ and $\widehat{\boldsymbol{s}}_{n}^B$ are bounded in the mean-square sense, using Assumption \ref{as6} and Lemma \ref{lemma3}. The proofs can be found in Appendix \ref{ap5}.

\begin{lemma}\label{lemma4}
(\textbf{Error bounds}) Under Assumptions \ref{as1}--\ref{as2} and \ref{as5}--\ref{as6}, and with sufficiently small step size $\mu$, the two error terms $\widehat{\boldsymbol{s}}_{n}^B$ and $\boldsymbol{d}_{n-1}$ are bounded in the mean-square sense, namely, it holds that
\begin{align}
\label{shatnb}
    &\mathds{E}\Vert \widehat{\boldsymbol{s}}_n^B\Vert^2 \le {\frac{\bar{\sigma}^2}{B}}\\
    &\mathds{E}\Vert \boldsymbol{d}_{n-1}\Vert^2 \le O(\mu^2)
\end{align}
$\hfill{\square}$
\end{lemma}

One difficulty in the convergence analysis of the proposed decentralized adversarial training algorithms, as listed in (\ref{d_d})--(\ref{d_combine_e}) and \eqref{d_consen}--\eqref{combine_e_consen}, is due to the non-smoothness of $J(w)$. In this case, verifying the stationarity of $J(w)$ requires working with its subdifferential. However, since $J(w)$ is both non-convex and non-smooth, $\partial_w J(w)$ might not even be defined in general. Fortunately, $\partial_w J(w)$ can be defined through \textit{Fr\'{e}chet subdifferntial} and \textit{weak-convexity} \cite{Thekumparampil019, DavisD19}. Basically, the Fr\'{e}chet subdifferntial of a function $J(w)$ is defined as the set:
 \begin{align}\label{def_sub}
    \partial_w J(w) \overset{\Delta}{=} \left\{\zeta \Big\vert \mathop{\mathrm{lim}\ \mathrm{inf} }\limits_{w'\to w}\  \frac{J(w') - J(w) - \zeta^{\sf T}(w' - w)}{\Vert w' - w\Vert} \ge 0\right\}
 \end{align}
Moreover, the smoothness condition implies weak convexity with the same parameter. Basically, the ``real'' loss function $f_k(w;\boldsymbol{x}_k, \boldsymbol{y}_k)$ is $L-$weakly convex over $w$ under the smooth condition of $Q_k$ in \eqref{smooth_q}, namely, for any $w$ and $w'$, it holds that,
\begin{align}\label{f_k_sub}
    &f_k(w';\boldsymbol{x}_k, \boldsymbol{y}_k) - f_k(w;\boldsymbol{x}_k, \boldsymbol{y}_k) \notag\\
    &\ge (\partial_w f_k(w;\boldsymbol{x}_k, \boldsymbol{y}_k))^{\sf T}(w' - w) - \frac{L}{2}\Vert w' - w \Vert^2
\end{align}
where the subdifferential $\partial_w f_k(w;\boldsymbol{x}_k, \boldsymbol{y}_k)$ is defined as the set of vectors that satisfies this inequality. The proof of \eqref{f_k_sub} can be found in section B.3 in \cite{Thekumparampil019}. As a result, since each local risk function $J_k(w)$ is an expectation of $f_k(w;\boldsymbol{x}_k, \boldsymbol{y}_k)$ over all samples, and the global risk function $J(w)$ is a convex combination of $J_k(w)$ for $k=1,\ldots, K$, it follows that $J_k(w)$ and $J(w)$ are $L$-weakly convex as well. Then, the subdifferential of $J(w)$ is defined by the following set:
\begin{align}
   \partial_w J(w)  \overset{\Delta}{=}  \left\{\zeta \vert J(w') - J(w) \ge \zeta^{\sf T}(w'- w) - \frac{L}{2}\Vert w' - w \Vert^2\right\} 
\end{align}
where the notion of Fr\'echet subdifferential, as defined in \eqref{def_sub}, is generalized for non-smooth and $L$-weakly convex functions. If there exists a $w^{\star}$ such that
\begin{align}
    0\in \partial_w J(w^{\star})
\end{align}
then we say $w^{\star}$ is a stationary point of $J(w)$.

However, it is still not tractable to carry out the convergence analysis with $\partial_w J(w)$ directly. Thus, motivated by \cite{jin2020local,rockafellar2015convex,DavisD19,lin2020gradient,Thekumparampil019}, we resort to the Moreau envelope method to study near-stationarity for non-smooth functions. Consider a function $h(z;w)$ defined by
\begin{align}
    h(z; w) \overset{\Delta}{=} J(z)+\frac{1}{2\gamma}\left\Vert w-z\right\Vert^2  
\end{align}
since $J(z)$ is $L$-weakly convex, $h(z; w)$ is strongly convex over $z$ if $\gamma < \frac{1}{L}$. In this case, the Moreau envelope of $J(w)$ is defined by \cite{sayed_2023}:
\begin{equation}
    J_{\gamma}(w)\overset{\Delta}{=} \min\limits_{z}  h(z;w) = \min\limits_{z} J(z) + \frac{1}{2\gamma}\left\Vert w-z\right\Vert^2
\end{equation}
It follows that the function $J(z)$ and its Moreau envelope $J_{\gamma}(w)$ have the same minimum, namely,
\begin{align}
    \min\limits_w J_{\gamma}(w) = \min\limits_z J(z)
\end{align}
Let
\begin{equation}
     \widehat{z} = \mathop{\text{argmin}}\limits_{z} h(z;w)
\end{equation}
then $\widehat{z}$ is uniquely determined by $w$ since $h(z;w)$ is strongly convex over the variable $z$. 

{Even when $J(z)$ is non-differentiable, the Moreau envelope will be differentiable relative to $w$. In particular, it holds that \cite{sayed_2023, lin2020gradient}:
\begin{align}\label{65}
   \frac{1}{\gamma}(w - \widehat{z}) = \nabla_w J_{\gamma}(w), \quad\; \nabla_w J_{\gamma}(w) \in \partial_w J(\widehat{z})
\end{align}
It follows that a stationary point of the Moreau envelope $J_{\gamma}(\cdot)$ is also a stationary point of the primal function $J(\cdot)$. Moreover, if $ J_{\gamma}(w)$ arrives at an approximate stationary point $w^o$, i.e., 
 if $\left\Vert\nabla_w J_{\gamma}(w^o)\right\Vert^2$ is upper bounded by a small value $\kappa^2$, then
 \begin{equation}\label{e_154}
     \min_{\zeta \in \partial_w J(\widehat{z}^o)}\Vert\zeta \Vert^2 \le \frac{1}{\gamma^2}\Vert w^o - \widehat{z}^o\Vert^2 = \Vert \nabla_w J_{\gamma}(w^o) \Vert^2 \le \kappa^2
 \end{equation} 
 where
 \begin{align}
     \widehat{z}^o =  \mathop{\text{argmin}}\limits_{z} h(z;w^o)
 \end{align} 
which implies that $w^o$ is $O(\kappa^2)$ near a $\kappa^2$-approximate stationary point of $J(w)$. Thus, it is reasonable to use the approximate stationarity of $\nabla_w J_\gamma(w)$ as a surrogate for the approximate stationarity of of $\partial_w J(w)$\cite{jin2020local,rockafellar2015convex,DavisD19,lin2020gradient,Thekumparampil019}.}

{We next show the convergence result of the proposed algorithms in the general non-smooth and non-convex environments, for which the proofs can be found in Appendix \ref{ap6}.}

\begin{theorem}\label{th2}(\textbf{Network convergence in non-smooth and non-convex environments}.)
{{Consider a network of $K$ agents running the decentralized adversarial training algorithms, as listed in (\ref{d_d})--(\ref{d_combine_e}) and \eqref{d_consen}--\eqref{combine_e_consen}.} Under Assumptions \ref{as1}--\ref{as2} and \ref{as5}--\ref{as6}, and {assuming} $J(w)$ is lower bounded by $\Delta$, after $N$ iterations, it holds that
\begin{align}\label{th_noncon_g}
    &\frac{1}{N}\sum\limits_{n}\mathds{E}||\nabla_w J_{\frac{1}{2L+1}}(\boldsymbol{w}_{c,n-1})||^2 \notag\\
    &\le  \frac{2(J_{\frac{1}{2L+1}}(w_{c,n_0})-\Delta)}{\mu N}  + O(\mu) + O(\epsilon^2)
\end{align}
where the $O(\epsilon^2)$ term arises from the approximation error when solving the inner maximization problem in \eqref{maximizer_t}.}

$\hfill\square$
\end{theorem}
{Theorem \ref{th2} is based on properties of the Moreau envelope, and it guarantees that when $\mu$ and $\epsilon$ are small, the proposed decentralized adversarial training algorithms will converge asymptotically to a point that is $O(\mu)+O(\epsilon^2)$ close to an approximate stationary point of $J(w)$ on average over iterations, and the convergence rate is $O(1/(\mu N))$. Basically, according to \eqref{th_noncon_g} and \eqref{65}, the following inequality holds for large $N$ with $N \ge O(1/\mu^{1+s})$ for any $s>1$:
\begin{equation}
    \frac{1}{N}\sum_n\mathds{E}\Vert\widehat{\boldsymbol{w}}_{N-1} - \boldsymbol{w}_{c,N-1}\Vert^2 \le \frac{\rho}{(2L+1)^2} 
\end{equation}
where $\widehat{\boldsymbol{w}}_{N-1}$ is an approximate stationary point {for} $J(w)$ satisfying
\begin{align}\label{convergence}
  \frac{1}{N}\sum_n \mathds{E}\left\{\min_{\boldsymbol{\zeta}\in\partial_w J(\widehat{\boldsymbol{w}}_{N-1})}||\boldsymbol{\zeta}||^2 \right\}
  \le  \rho \overset{\Delta}{=} O(\mu) + O(\epsilon^2)
\end{align} 
On the other hand, from the non-asymptotic perspective, for a fixed $N$, selecting the step size $\mu$ such that $\mu\le O(1/\sqrt{N})$ ensures the proposed algorithms achieve a convergence rate of $O(1/\sqrt{N})$. This rate is similar to the results in the literature on non-convex optimization \cite{garrigo} and adversarial learning \cite{sinha2017certifying}. Note that the proof of Theorem \ref{th2} relies on the strong-convexity of the Moreau envelope function $J_\gamma(w)$. As a result, Theorem \ref{th2} can also be used to demonstrate the convergence of the proposed decentralized adversarial training algorithms in non-smooth and convex environments. Basically, when $Q_k$ is convex over $w$, Danskin's theorem guarantees that the global risk function $J(w)$ is also convex. Then, by selecting the same $\gamma$, the function $J_\gamma(w)$ remains strongly convex. Furthermore, in the case of convex functions, any stationary point also admits a global minimum.}

{However, compared to the the traditional non-convex optimization in the clean case \cite{garrigo}, there is an extra error term, i.e., $O(\epsilon^2)$, which affects the final precision in the right-hand side of (\ref{th_noncon_g}).}  The additional error term arises from the approximation error of the inner maximization; its presence is common in studies on adversarial learning, for example, \cite{sinha2017certifying, WangM0YZG19}.  When the maximizer can be found exactly, this term can be eliminated. As it is usually intractable to compute the accurate maximizer in the non-convex case, approximation methods are applied. Then, the smaller the $\epsilon$ is, the better the exact maximizer can be approximated by the Taylor expansion. 

{Moreover, although both this paper and \cite{sinha2017certifying} study the convergence of the stochastic gradient algorithms in the context of non-convex optimization and adversarial training, and both obtain similar convergence rate of $O(1/(\mu N))$, our results differ as they are derived under different assumptions and conditions. Reference \cite{sinha2017certifying} focuses on the convergence of the adversarial extension of the stochastic gradient algorithm in the single-agent setting, and requires the loss function to be strongly-concave with respect to the perturbation variable $\delta$. This condition ensures some nice properties of the risk function $J(w)$. Basically, the strong-concavity guarantees that the optimal perturbation of \eqref{maximizer_t} is unique and satisfies a Lipschitz condition over any two different model parameters.
As a result, functions $f_k(w;\boldsymbol{x}_k,\boldsymbol{y}_k)$ and $J(w)$ remain differentiable, and their gradients satisfy the traditional Lipschitz condition. This allows the traditional analysis method of the stochastic gradient algorithm, developed for nonconvex optimization and standard (clean) training, to be directly extended to the adversarial setting. In contrast, we consider the general nonconvex and nonconcave minimax optimization for adversarial learning in the multi-agent setting. Consequently, functions $f_k(w;\boldsymbol{x}_k,\boldsymbol{y}_k)$ and $J(w)$ may not be differentiable. Thus, we use different conditions, and our analysis follows a different framework from \cite{sinha2017certifying}. }

{Naturally, if we adopt the same conditions as those in \cite{sinha2017certifying}, their results can be extended to the multi-agent setting. Assume the function $Q_k(w;\boldsymbol{x}_k+\delta,\boldsymbol{y}_k)$ is $\tau-$strongly concave over $\delta$, then the optimal perturbation of \eqref{maximizer_t} is unique. As a result, the functions $f_k(w;\boldsymbol{x}_k,\boldsymbol{y}_k)$, $J_k(w)$, and $J(w)$ are differentiable. We next state the convergence of the proposed decentralized adversarial training algorithms under the strongly-concave condition, for which the proofs can be found from Appendix \ref{ap_coro_sinh}.}
\begin{corollary}\label{coro_sinh}
{Consider a network of $K$ agents running the decentralized adversarial training algorithms, as listed in (\ref{d_d})--(\ref{d_combine_e}) and \eqref{d_consen}--\eqref{combine_e_consen}. Under Assumptions \ref{as1}--\ref{as2} and \ref{as5}--\ref{as6}, and assuming $J(w)$ is lower bounded, and the function $Q_k(w;\boldsymbol{x}_k+\delta,\boldsymbol{y}_k)$ is $\tau-$strongly concave over $\delta$, after $N$ iterations, it holds that 
\begin{align}
\frac{1}{N} \sum_n \mathds{E}\Vert \nabla_w J(\boldsymbol{w}_{c,n-1}) \Vert^2 \le O(\frac{1}{\mu N}) + O(\mu) + O(\epsilon^2)
\end{align}
where the $O(\epsilon^2)$ term arises from the approximation error when solving the inner maximization problem in \eqref{maximizer_t}.}

$\hfill\square$
\end{corollary}

{Corollary \ref{coro_sinh} extends the convergence result of \cite{sinha2017certifying} from the single-agent setting to the multi-agent case. Unlike Theorem \ref{th2}, which guarantees that the proposed algorithms converge to a point near an approximate stationary point of $J(w)$ under the general nonconvex nonconcave condition, Corollary \ref{coro_sinh} means that the proposed algorithms can converge directly to an approximate stationary point of $J(w)$ with the nonconvex strongly-concave condition. }

\section{Computer simulations}

\subsection{Logistic regression}
In this section, we illustrate the performance of the adversarial diffusion strategy from Algorithm \ref{al_ro_1} and the adversarial consensus strategy from Algorithm \ref{al_ro_2} using a logistic regression application. Let $\boldsymbol{\gamma}$ be a binary variable that takes values from $\{-1,1\}$, and {let} $\boldsymbol{h} \in \mathbbm{R}^M$ be a feature variable. The robust logistic regression problem by a network of agents employs the risk functions:
\begin{align}\label{18}
J_k (w) = \mathds{E}\max\limits_{\Vert{\delta}\Vert_{p_k}\le \epsilon_k}\left\{\ln{(1+e^{-\boldsymbol{\gamma}(\boldsymbol{h} + {\delta})^{\sf T}{w}}})\right\}
\end{align}
Note that when $p_k = 2$, the analytical solution for the inner maximizer (i.e., the worst-case perturbation) is given by
\begin{align}
    \boldsymbol{\delta}^{\star} = -\epsilon_k\boldsymbol{\gamma}\frac{{w}}{\Vert{w}\Vert}
\end{align}
which is consistent with the perturbation computed from FGM \cite{miyato2016adversarial}. {On the other hand,} if $p_k = \infty$, the closed-form expression for the inner maximizer changes to
\begin{align}
    \boldsymbol{\delta}^{\star} = -\epsilon_k\boldsymbol{\gamma}\mathrm{sign}(w)
\end{align}
which is the perturbation computed from the FGSM method \cite{goodfellow2014explaining}.

\begin{figure*}[htbp]
\centering
\subfigure[]
{\label{fig1}\includegraphics[width=4.5cm]{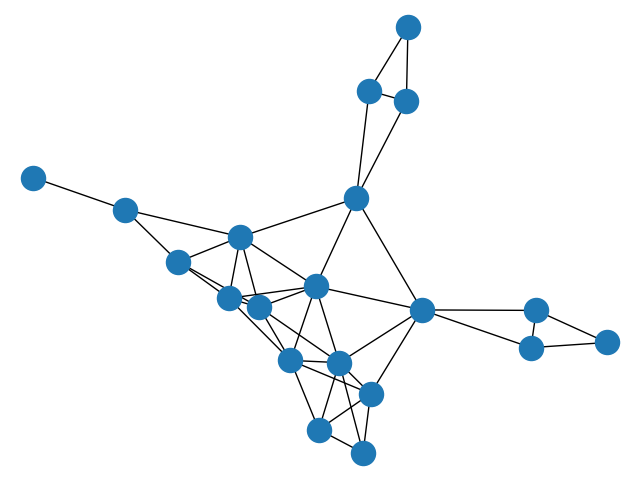}}
\subfigure[]{\label{g_mnist}\includegraphics[width=4.5cm]{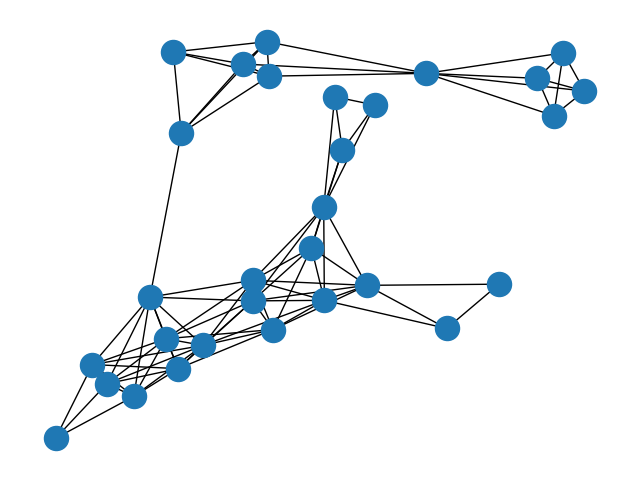}}
\subfigure[]{\label{g_cifar}\includegraphics[width=4.5cm]{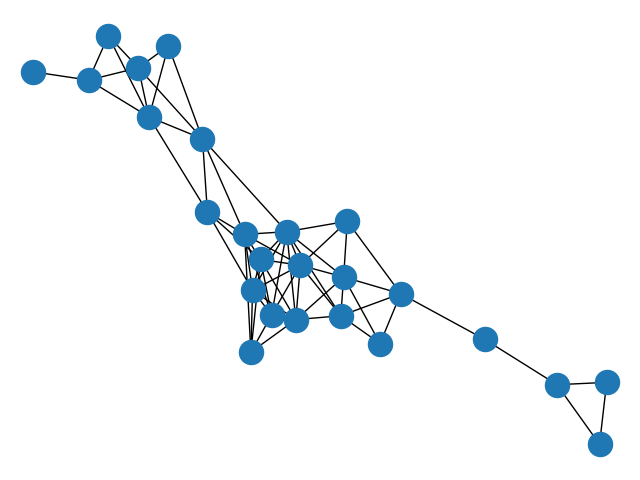}}
\caption{The randomly generated graph structures used in the experiments. (a) Graph for convex scenario (MNIST and CIFAR10). (b) Graph for non-convex scenario (MNIST). (c) Graph for non-convex scenario (CIFAR10). {The process of generating the graph structures consists of the following two steps:
(1) For a graph with $K$ nodes, we randomly generate an adjacency matrix $D$ that represents a connected graph. Specifically, for any two nodes $\ell$ and $k$ with coordinates drawn from a uniform distribution, an edge is formed between them if their mean squared distance is smaller than a predefined threshold (e.g., 0.3), and we set $D_{\ell k} = D_{k \ell} = 1$.  We repeat this process until a connected graph is generated. The connectivity of the graph is verified by the Laplacian matrix: if the second smallest eigenvalue of the Laplacian matrix is non-zero, then the graph is connected.  (2) Given the adjacency matrix D, we generate the weights $a_{\ell k}$ between any two nodes $\ell$ and $k$ using the Metropolis rule from \cite{sayed2014adaptation}.}}
\label{fig4}
\end{figure*}
In our experiments, we use both the MNIST \cite{deng2012mnist} and CIFAR10 \cite{krizhevsky2010cifar} datasets, and randomly generate a graph with 20 nodes {by using the Metropolis rule in \cite{sayed2014adaptation}}, shown in Fig. \ref{fig1}. For each dataset {and decentralized training strategy}, three networks are trained, including two homogeneous adversarial networks {using $\ell_2$ and $\ell_\infty$ bounded perturbations} where all agents are attacked by the same {type} of perturbations, and one clean network {trained with original clean samples}. We limit our simulations to binary classification in this example.  For this reason, we consider samples with digits \textit{$0$} and \textit{$1$} from MNIST, and images for airplanes and automobiles from CIFAR10. For the $\ell_2$ {case}, we set the perturbation bound in (\ref{18}) to $\epsilon_k = 4$ for MNIST and $\epsilon_k = 1.5$ for CIFAR10. {In the same token}, the attack {strengths} are set to $\epsilon_k = 0.3$ for MNIST and $\epsilon_k = 0.035$ for CIFAR10 under the $\ell_\infty$ {setting}. In the test phase, we compute the average classification error across the networks to measure the performance of the multi-agent system against perturbations of different strengths. In other words,  we test the robustness of the trained networks in the worst situation where all agents receive adversarial examples. 

We first illustrate the convergence of the algorithms. From {Fig.} \ref{fig2}, we observe a decrease in the classification error towards a limiting value for each subplot. {Note that for the CIFAT10 dataset, to ensure both the convergence speed and the stability of algorithms, we initially use the a moderately large step size 0.01, which is then reduced by a factor of 10 at the iterations of 10000 and 20000.}
\begin{figure*}[htbp]
\centering
\subfigure[MNIST, $\ell_2$ {perturbations}.]{\label{sub_1}\includegraphics[width=4.4cm]{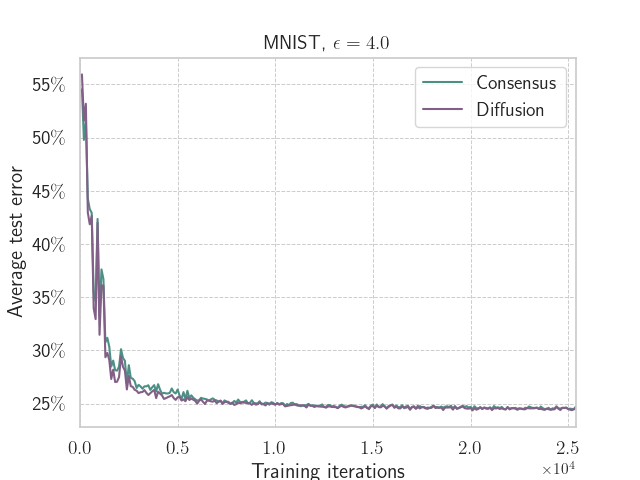}}
\subfigure[MNIST, $\ell_\infty$ perturbations.]{\label{sub_2}\includegraphics[width=4.4cm]{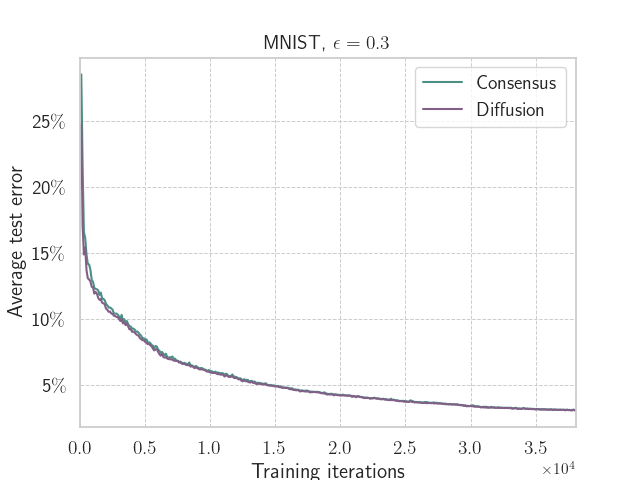}}
\subfigure[CIFAR, $\ell_2$ perturbations.]{\label{sub_3}\includegraphics[width=4.4cm]{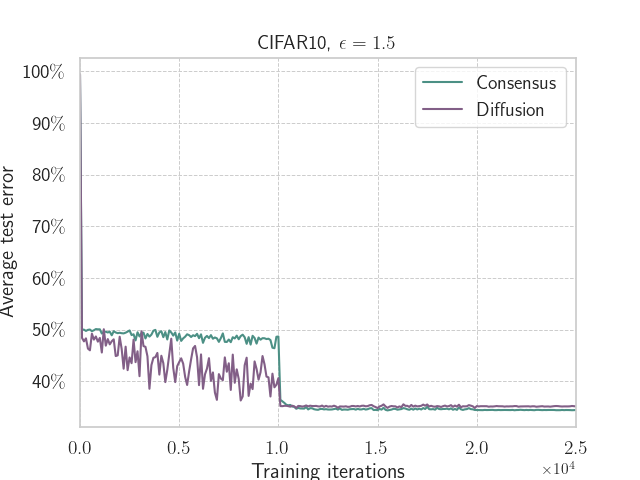}}
\subfigure[CIFAR, $\ell_\infty$ perturbations.]{\label{sub_4}\includegraphics[width = 4.4cm]{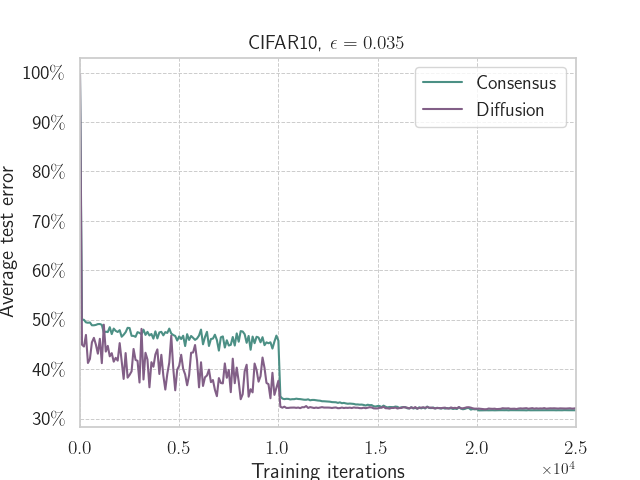}}
\caption{{Convergence plots for the two datasets using the logistic loss. {The legends show the decentralized strategy  used in the training phase. (a) The evolution of the average classification error across all agents over $\ell_2$ adversarial examples bounded by $\epsilon_k = 4$ during training for MNIST. (b)--(d) follow a similar logic but for different norms and datasets.}}} 
\label{fig2}
\end{figure*}

The robust behavior of the algorithms to $\ell_2$ and $\ell_\infty$ attacks is illustrated in Fig. \ref{fig3} for both MNIST and CIFAR10. {We comment on the curves for the adversarial diffusion strategy and MNIST. Similar remarks hold for the consensus strategy and CIFAR10.} In the test phase, we use perturbations generated in one of two ways: using the worst-case (FGM for $\ell_2$ and FGSM for $\ell_\infty$) construction and also using the DeepFool construction \cite{moosavi2016deepfool}. The blue curves are obtained by training the networks using the traditional diffusion learning strategy without accounting for robustness. The networks are fed with worst-case perturbed samples (generated using FGM or FGSM) during testing corresponding to different levels  of attack strength. The blue curves show that the classification error deteriorates rapidly. The red curves repeat the same experiments except that the networks are now trained with decentralized adversarial training algorithms.  It is seen in the green curves that the testing error is more resilient and the degradation is better controlled than the blue curves. The same experiments are repeated using the same adversarially trained networks and the clean ones to show robustness to DeepFool attacks, where the perturbed samples are now generated using DeepFool as opposed to FGM or FGSM in the test phase. The red curves show the robustness of the clean networks to DeepFool, while the purples ones correspond to the adversarially trained networks. Note that for  $\ell_\infty$ attacks, i.e., Figs. \ref{sub_2_3_d} and \ref{sub_4_3_d}, the DeepFool and FGSM possess similar destructive power so that the blue and the red curves, and also the green and purple curves, coincide with each other. Here robustness can be explained from two perspectives. First, the attack strength of DeepFool never outperforms the worst-case perturbations for the same model, so that robustness to the worst-case perturbation guarantees robustness to DeepFool. Second, comparing the purple and red curves, the evolution of the robustness of the networks to DeepFool is better controlled by our algorithms than the clean networks.
\begin{figure*}[htbp]
\centering
\subfigure[MNIST, $\ell_2$ perturbations.]{\label{sub_1_3_c}\includegraphics[width=4.4cm]{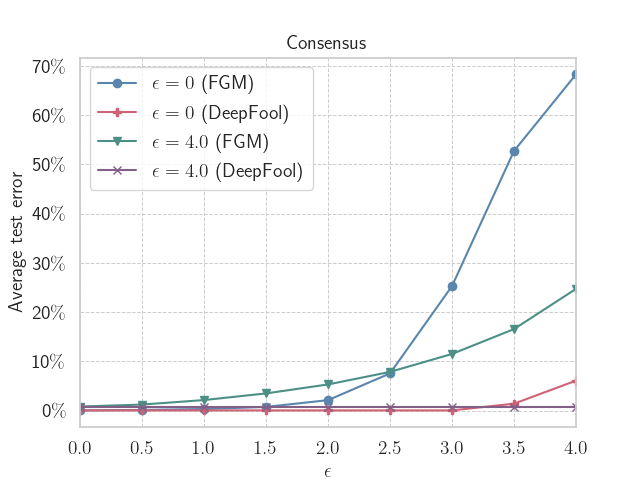}}
\subfigure[MNIST, $\ell_\infty$ perturbations.]{\label{sub_2_3_c}\includegraphics[width=4.4cm]{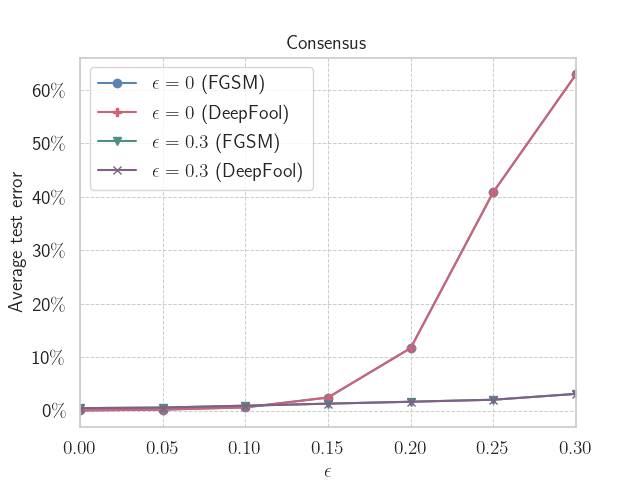}}
\subfigure[CIFAR, $\ell_2$ perturbations.]{\label{sub_3_3_c}\includegraphics[width=4.4cm]{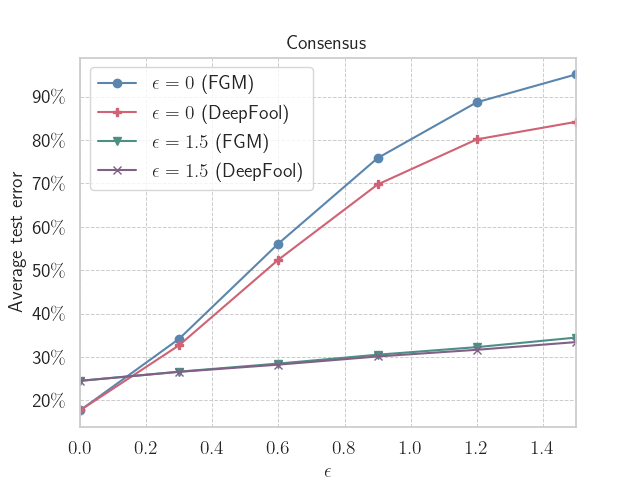}}
\subfigure[CIFAR, $\ell_\infty$ perturbations.]{\label{sub_4_3_c}\includegraphics[width=4.4cm]{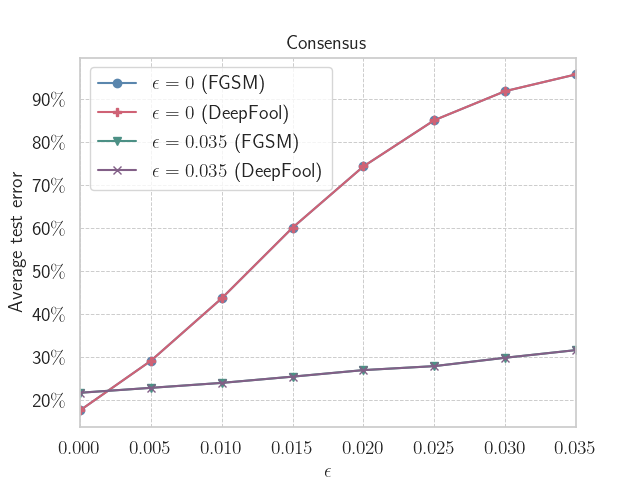}}
\subfigure[MNIST, $\ell_2$ perturbations.]
{\label{sub_1_3_d}\includegraphics[width=4.4cm]{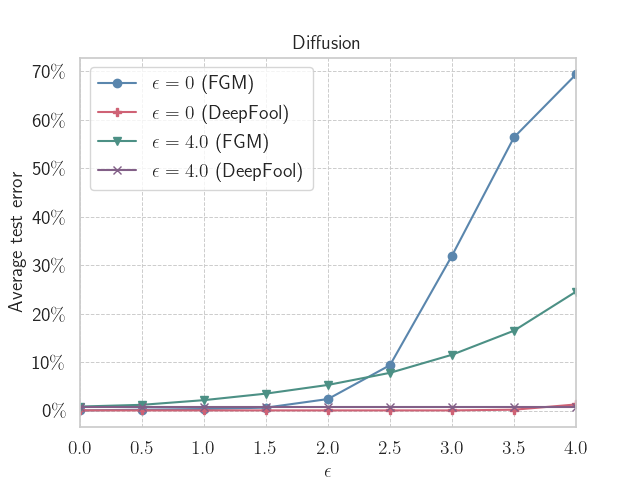}}
\subfigure[MNIST, $\ell_\infty$ perturbations.]{\label{sub_2_3_d}\includegraphics[width=4.4cm]{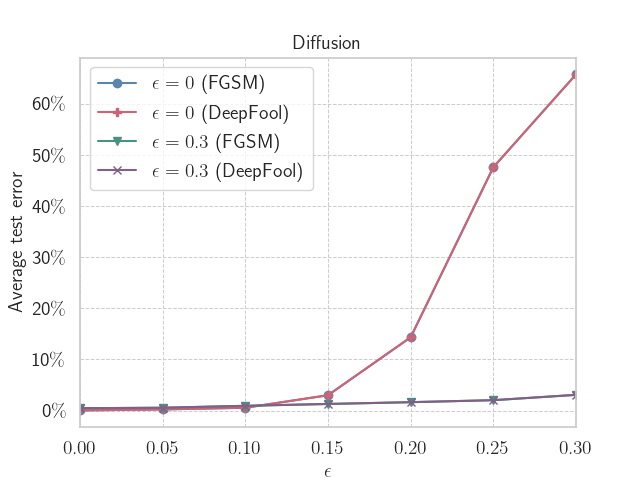}}
\subfigure[CIFAR, $\ell_2$ perturbations.]{\label{sub_3_3_d}\includegraphics[width=4.4cm]{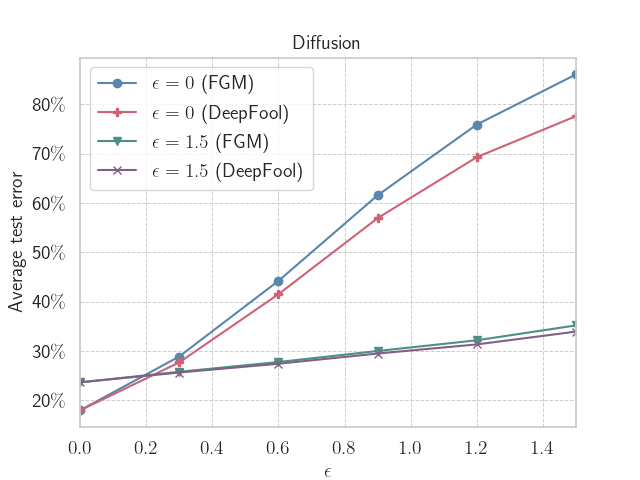}}
\subfigure[CIFAR, $\ell_\infty$ perturbations.]{\label{sub_4_3_d}\includegraphics[width=4.4cm]{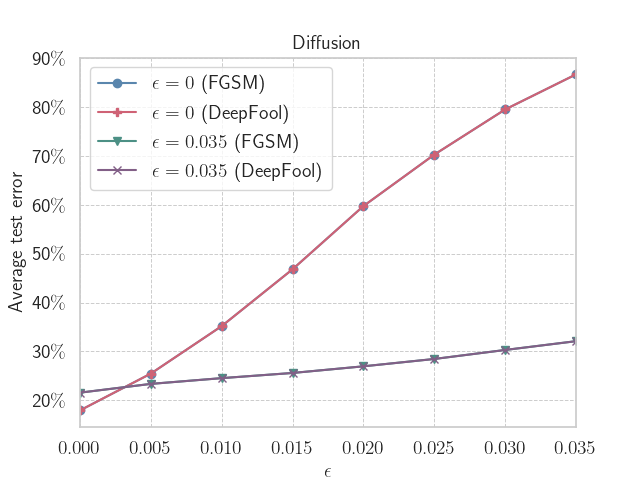}}
\caption{{Robustness plots for the two datasets in convex environments. (a) Average classification error over the graph versus perturbation size for MNIST to $\ell_2$ attacks. The legends demonstrate the perturbation bound used in the training phase and the attack method in the test phase. For instance, $\epsilon = 0$ (FGM) corresponds to robustness of the clean network to FGM attack. Also, the title of the plot shows the decentralized strategy used in the training phase. (b)--(h) follow the similar logic but for different norms, datasets and decentralized strategies.}}
\label{fig3}
\end{figure*}


In Fig. \ref{fig4_visual}, we plot some randomly selected CIFAR10 images, their perturbed versions, and the classification decisions generated by the nonrobust algorithm and its adversarial version {listed in Algorithm \ref{al_ro_1}}. We observe from the figure that no matter which attack method is applied, the perturbations are always imperceptible to the human eye. Moreover, while the nonrobust algorithm fails to classify correctly in most cases, the adversarial algorithm is more robust and leads to fewer classification errors.

\begin{figure}[htbp]
\centerline{\includegraphics[width=8cm]{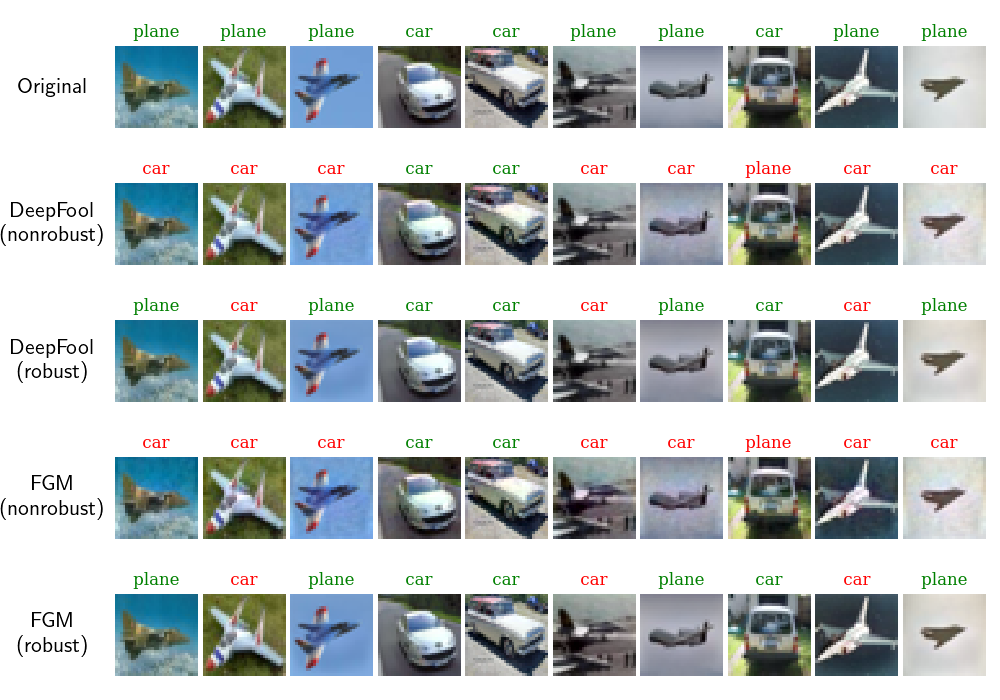}}
\caption{Visualization of the original and adversarial samples. The first row consists of 10 random original samples with the titles representing their true classes. The second row shows the adversarial examples generated by DeepFool and applied to {a graph trained by the standard nonrobust algorithm.} The third row shows the results obtained by the adversarial {diffusion} algorithm. The titles are the predictions by the corresponding models. The same construction is repeated in the last two rows using FGM. If the prediction of an image is wrong, the title is shown in red color. It is seen that the adversarial algorithm fails less frequently.} 
\label{fig4_visual}
\end{figure}

\subsection{Neural networks}
{In this section, we illustrate the effectiveness of the proposed decentralized adversarial training algorithm, i.e., the adversarial diffusion strategy (\ref{d_d})--(\ref{d_combine_e}) and the adversarial consensus strategy \eqref{d_consen}--\eqref{combine_e_consen}} in non-convex environments for the MNIST and CIFAR10 datasets for both homogeneous and heterogeneous networks. {In the decentralized experiments, the full training dataset is divided into $K$ subsets, and each agent can only observe one subset.} Moreover, in the simulation, we utilize the random initialization technique from \cite{WongRK20}, which has been shown to enhance the stability of the adversarial training process in the single-agent case. 

\textbf{MNIST setup}. For MNIST, we randomly generate a graph with 30 nodes for the distributed setting, and its structure is shown in Fig. \ref{g_mnist},  where each node corresponds to a convolutional neural network 
 (CNN) model following a similar structure to the one used in \cite{madry2017towards} but adding the batch normalization \cite{sayed_2023, IoffeS15} layer. The output size of the last layer is 10. We set the perturbation bounds to $\epsilon_k = 2$ for $\ell_2$ attacks and $\epsilon_k = 0.3$ for $\ell_\infty$ ones. {In training robust neural networks, we use the multi-step attacks, i.e., PGD and PGM. }                                                                     
 
 \textbf{CIFAR10 setup}. We use a randomly generated graph with 25 agents, where we now use the CNN structure from \cite{ZhangYJXGJ19} for each agent. The graph structure is shown in Fig. \ref{g_cifar}. The perturbation bounds are set to $\epsilon_k = 0.5$ for $\ell_2$ attacks 
 and $\epsilon_k = 0.03$ for $\ell_\infty$ attacks.  Since the CIFAR10 dataset is more complex than MNIST, we use 
the single-step attacks, i.e., FGSM and FGM, to train neural networks. {Moreover, we apply the piecewise learning rate strategy from \cite{ZhuH0ST23}. Specifically, the initial learning rate is set to $0.1$ and is divided by $10$ when the training process reaches $40\%$, $60\%$, and $80\%$ of the total number of epoches. }
 
\subsubsection{Homogeneous networks} We show the results corresponding to the homogeneous networks, where all agents in {the} graph are attacked by the same type of perturbation (i.e., {$\epsilon_k$ and $p_k$ is uniform for all $k$}). For instance, for MNIST with  $\ell_2$ attacks, all agents receive perturbations bounded by $\epsilon_k = 2$, and the global risk is
\begin{align}
    J(w) = \sum\limits_{k=1}^{30} \pi_k\mathds{E}\max\limits_{\Vert\delta\Vert\le 2}Q_k(w;\boldsymbol{x}_k+\delta,\boldsymbol{y}_k)
\end{align} 

\begin{figure*}[htbp]
\centering
\subfigure[MNIST, $\ell_2$ perturbations.]{\label{sub_1_5}\includegraphics[width=4.4cm]{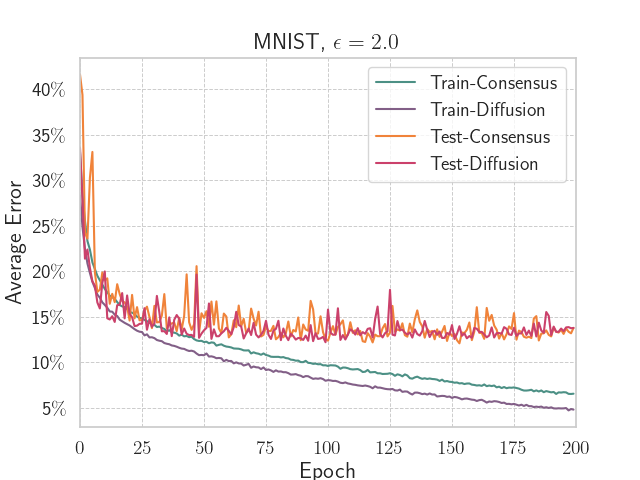}}
\subfigure[MNIST, $\ell_\infty$ perturbations.]{\label{sub_2_5}\includegraphics[width=4.4cm]{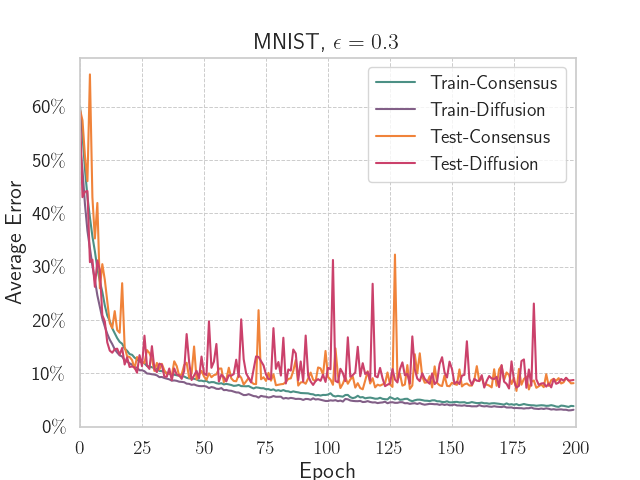}}
\subfigure[CIFAR, $\ell_2$ perturbations.]{\label{sub_3_5}\includegraphics[width=4.4cm]{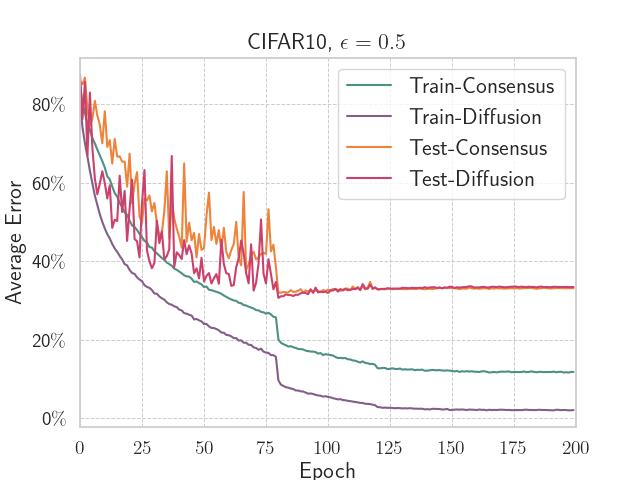}}
\subfigure[CIFAR, $\ell_\infty$ perturbations.]{\label{sub_4_5}\includegraphics[width = 4.4cm]{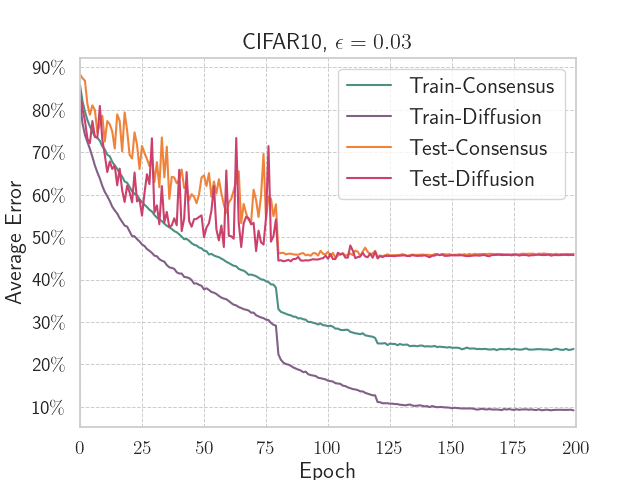}}
\caption{{Convergence plots for the two datasets in non-convex {environments. {The legends show the strategy and whether the results correspond to training or test data used in the training phase. For example, a label such as "Train-consensus" represents the evolution of the training error when using the consensus strategy.}  (a) The evolution of the average classification error across all agents over $\ell_2$ adversarial examples bounded by $\epsilon_k = 2$ during training for MNIST. (b)--(d) follow a similar logic but for different norms and datasets.}}}
\label{fig5}
\end{figure*}

Fig. \ref{fig5} shows the convergence plots for the two datasets, from which we observe that the average classification error across the networks to the trained attacks converges after sufficient iterations.

\begin{figure*}[htbp]
\centering
\subfigure[MNIST, $\ell_2$ perturbations.]{\label{sub_1_6_c}\includegraphics[width=4.4cm]{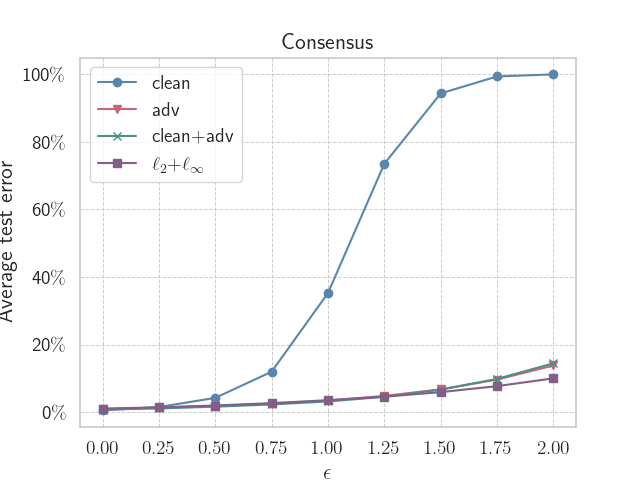}}
\subfigure[MNIST, $\ell_\infty$ perturbations.]{\label{sub_2_6_c}\includegraphics[width=4.4cm]{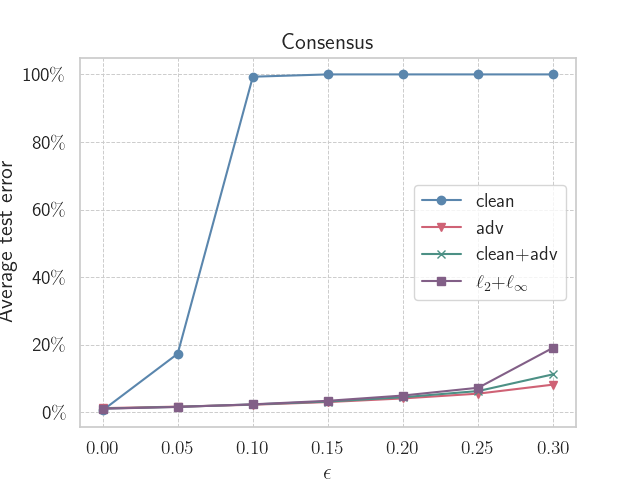}}
\subfigure[CIFAR, $\ell_2$ perturbations.]{\label{sub_3_6_c}\includegraphics[width=4.4cm]{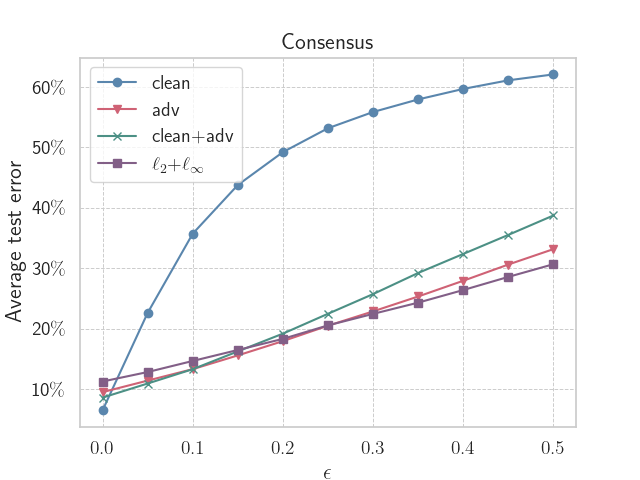}}
\subfigure[CIFAR, $\ell_\infty$ perturbations.]{\label{sub_4_6_c}\includegraphics[width=4.4cm]{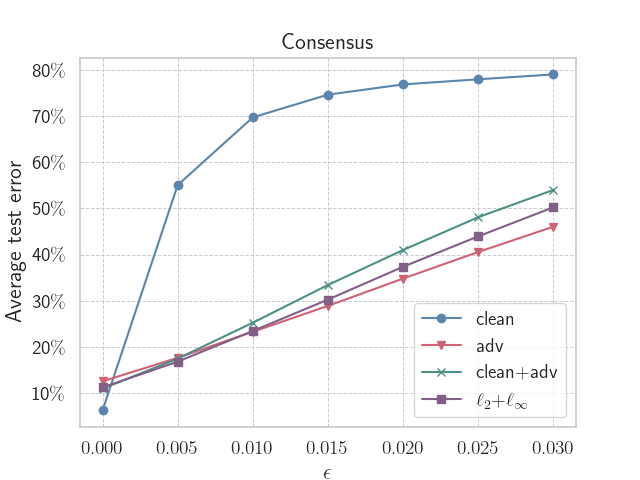}}

\subfigure[MNIST, $\ell_2$ perturbations.]
{\label{sub_1_6_d}\includegraphics[width=4.4cm]{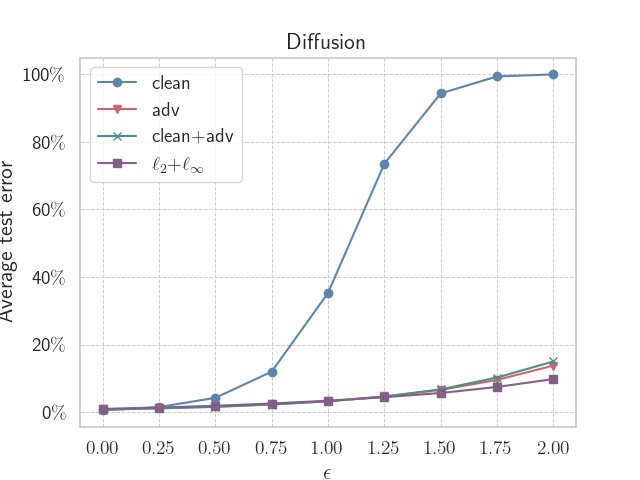}}
\subfigure[MNIST, $\ell_\infty$ perturbations.]{\label{sub_2_6_d}\includegraphics[width=4.4cm]{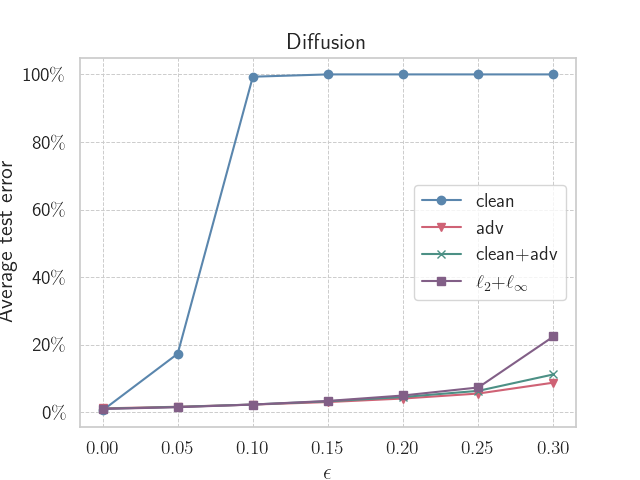}}
\subfigure[CIFAR, $\ell_2$ perturbations.]{\label{sub_3_6_d}\includegraphics[width=4.4cm]{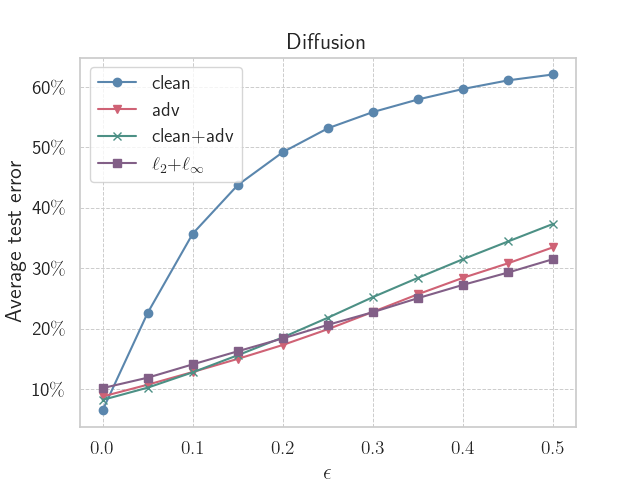}}
\subfigure[CIFAR, $\ell_\infty$ perturbations.]{\label{sub_4_6_d}\includegraphics[width=4.4cm]{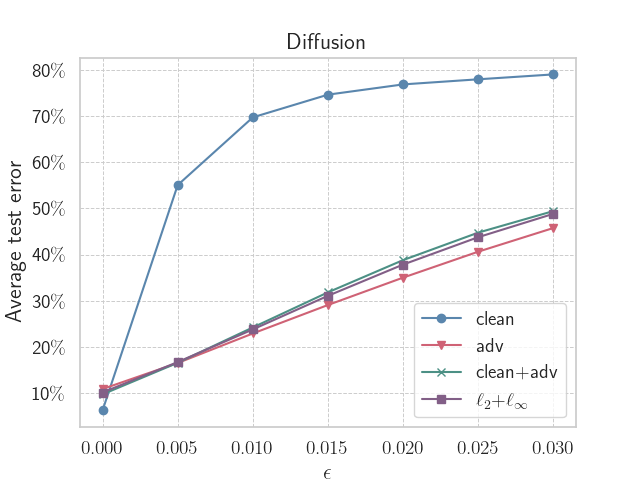}}
\caption{{Robustness plots for the two datasets in non-convex environments. {The blue, red, green and purple curves in each subplot correspond to the robustness behavior of the clean, homogeneous, clean+adv, and $\ell_2+\ell_\infty$ heterogeneous networks respectively.  (a) Average classification error across the networks versus perturbation size to PGM attack for MNIST. (b)--(h) follow a similar logic but for different norms, datasets and training strategies.}}}
\label{fig6}
\end{figure*}

\begin{table}[h!t]
\scriptsize
   \centering
   \begin{spacing}{1.5}
   \caption{{The classification error for the homogeneous networks under various attacks}}
   \label{tab2}
\begin{tabular}{c c c c c c c c c c  c c c c c}
\toprule
data&MNIST&MNIST&MNIST&MNIST&MNIST&MNIST\\
norm&$\ell_2$& $\ell_2$&$\ell_2$&$\ell_\infty$&$\ell_\infty$&$\ell_\infty$\\
attack&PGM&DeepFool&FMN&PGD&DeepFool&FMN\\
error-clean&100\%&100\%&100\%&100\%&100\%&100\%\\
error-nonco&26.00\%&28.76\%&31.78\%&24.58\%&19.96\%&27.04\%\\
error-centra&14.08\%&15.31\%&18.29\%&9.87\%&8.51\%&10.78\%\\
{error-consen}&13.84\%&13.69\%&16.85\%&8.16\%&6.76\%&7.81\%\\
error-diff& 13.81\%&13.51\%&16.96\%&8.76\%&7.02\%&8.07\%\\
\midrule
data&CIFAR&CIFAR&CIFAR&CIFAR&CIFAR&CIFAR\\
norm&$\ell_2$& $\ell_2$&$\ell_2$&$\ell_\infty$&$\ell_\infty$&$\ell_\infty$\\
attack&PGM&DeepFool&FMN&PGD&DeepFool&FMN\\
error-clean&99\%&98\%&83\%&100\%&100\%&100\%\\
error-nonco&66.36\%&64.70\%&64.15\%&84.09\%&81.12\%&83.55\%\\
error-centra&44.96\%&40.90\%&40.14\%&69.52\%&62.05\%&68.34\%\\
{error-consen}&38.53\%&35.51\%&33.49\%&57.78\%&50.99\%&55.64\%\\
error-diff&39.87\%&34.98\%&33.93\%&60.95\%&50.04\%&58.84\%\\
\bottomrule
\end{tabular}
\end{spacing}
\end{table}
We next show how the robustness of the multi-agent systems can be improved by the proposed algorithm compared with the clean networks. Similar to the convex case, three networks are trained for each dataset under each training strategy, including two homogeneous networks adversarially trained by $\ell_2$ and $\ell_\infty$ bounded perturbations respectively, and one clean system fed only with clean samples in the training phase, which is equivalent to the case of $\epsilon = 0$. Fig. \ref{fig6} illustrates the robust behavior of the multi-agent systems to the attacks that are visible in the training process. That is, we observe from the blue curve in each subplot a rapid increase for the average classification error across the network to adversarial examples. This demonstrates the vulnerability of the clean models. Fortunately, this phenomenon can be mitigated by the proposed algorithms as the red curves shows. We further test the robustness of the networks to some multi-step attacks, i.e., PGD or PGM \cite{madry2017towards}, DeepFool\cite{moosavi2016deepfool}, and FMN \cite{PintorRBB21}, with the Foolbox package \cite{rauber2017foolbox}, and the results are shown in Table \ref{tab2}, where the row \textit{error-clean} corresponds to the classification error of the clean networks, the row \textit{error-diff} relates to the robustness of the adversarially trained networks by our diffusion algorithm, and {the row \textit{error-consen} tests the robustness of the adversarially trained networks by our consensus algorithm.} Different from the convex case where the closed-form solution for the inner maximization can be exactly obtained, it is only possible to find an approximation for it in the non-convex case. Thus, the attacks used in the training process are not guaranteed to be the most powerful (i.e., worst-case) perturbations within the norm-bound regions.
\begin{figure*}[htbp]
\centering
\subfigure[MNIST, $\ell_2$ perturbations.]{\label{fs_11}\includegraphics[width=4.4cm]{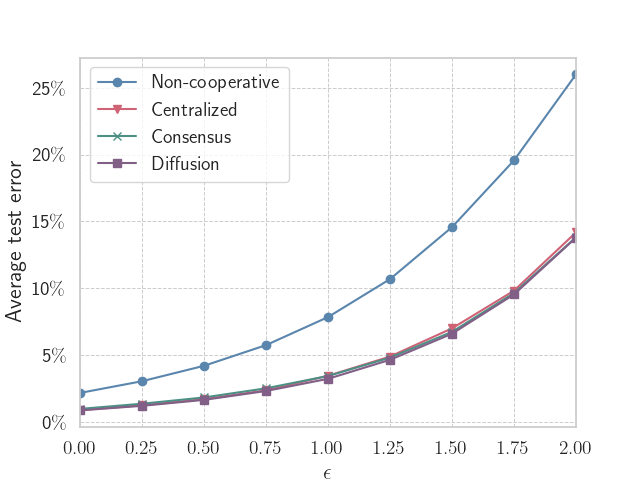}}
\subfigure[MNIST, $\ell_\infty$ perturbations.]{\label{fs_12}\includegraphics[width=4.4cm]{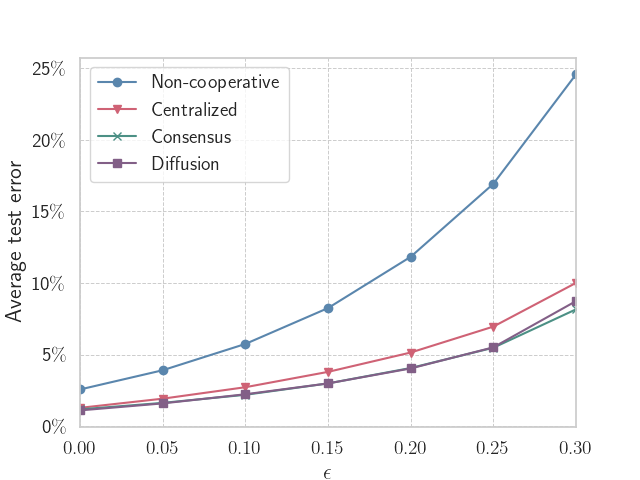}}
\subfigure[CIFAR, $\ell_2$ perturbations. ]{\label{fs_13}\includegraphics[width=4.4cm]{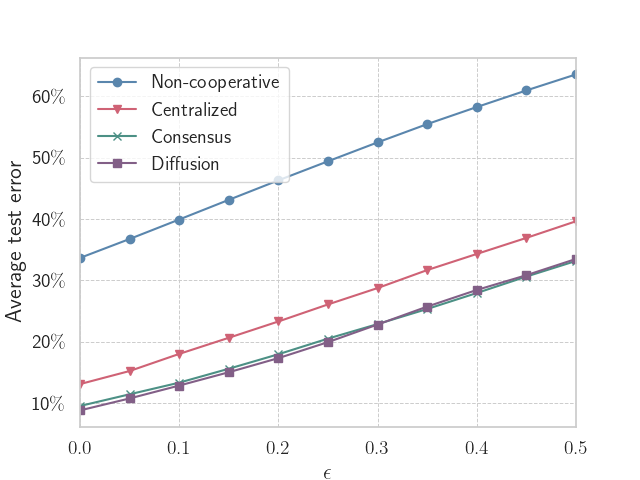}}
\subfigure[CIFAR, $\ell_\infty$ perturbations.]{\label{fs_14}\includegraphics[width = 4.4cm]{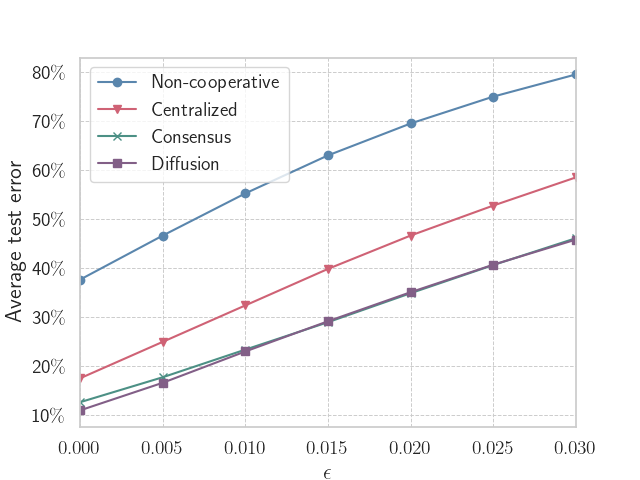}}
\caption{{{Comparison among four distributed strategies: non-cooperative, centralized, consensus and diffusion. (a) Classification error versus perturbation size for MNIST to FGM; (b)--(d) follow a similar logic but for different norms and  datasets.}}}
\label{fig_coopvsnoncoop}
\end{figure*}

 We finally compare the robustness of the networks adversarially trained by four different distributed strategies: consensus, diffusion, non-cooperative, and centralized. Basically, all agents train their own models independently in the non-cooperative setting, while all agents share their data with a fusion center to do the computation in the centralized method. {Thus, the centralized setting is the same as the traditional single-agent learning.} The results are shown in Fig. \ref{fig_coopvsnoncoop} and Table \ref{tab2}, where the row \textit{error-nonco} corresponds to the robustness of the non-cooperative networks, and  \textit{error-centra} shows the results associated with the centralized method. Comparing the results associated with the three methods from Fig. \ref{fig_coopvsnoncoop} and Table \ref{tab2}, we observe that the diffusion networks are more robust than the non-cooperative ones. This is due to the cooperation over the graph topology. Furthermore, we surprisingly observe that the robustness of the decentralized networks, i.e., trained by consensus and diffusion, outperform that of the centralized architectures, especially for the CIFAR10 dataset. This behavior is different from what one would observe under clean training \cite{sayed2014adaptation,kayaalp2022dif}  and from the view that the decentralized method may hurt generalization \cite{ZhuHZNST22}. One explanation for this phenomenon in the context of adversarial training is that decentralized stochastic gradient algorithms have been shown to favor flatter minimizers than the centralized method, and flat minimizers in turn have been shown to generalize better than sharp minimizers \cite{KeskarMNST17,ZhuH0ST23, cao2024trade}. To illustrate this effect, we visualize the flatness of the trained models by the method from \cite{Li0TSG18}. The results are shown in Fig. \ref{loss_visual}, from which we observe that the neighborhood of models trained by the decentralized methods are flatter than the minima for the centralized solution. We also notice that the superiority of the decentralized methods over the centralized solution is amplified in the context of adversarial training. For example, for CIFAR10, {the clean accuracy of the diffusion network and the consensus network is $1\%$ higher than the centralized one in the clean training \cite{cao2024trade}.} However, this gap increases to $10\%$ or even more (as Fig. \ref{fig_coopvsnoncoop} and Table \ref{tab2} show) under adversarial training. This observation suggests that there is merit to networked learning where the graph topology contributes to robustness. %

 \begin{figure}[htbp]
\centering
\subfigure[CIFAR, $\ell_2$ perturbations.]{\label{sub_loss_v_1}\includegraphics[width=5cm]{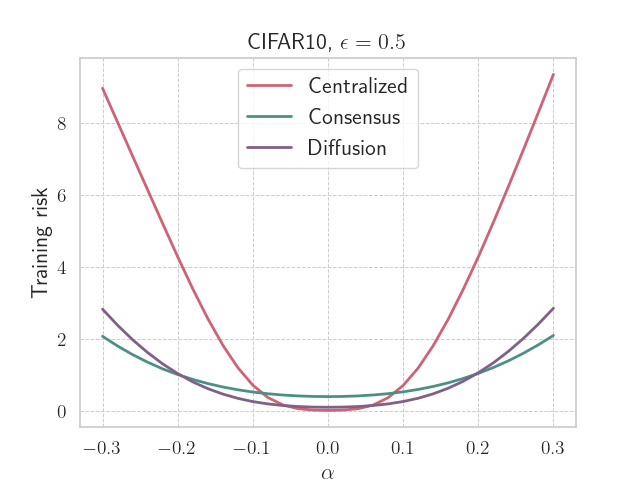}}
\subfigure[CIFAR, $\ell_\infty$ perturbations.]{\label{sub_loss_v_2}\includegraphics[width=5cm]{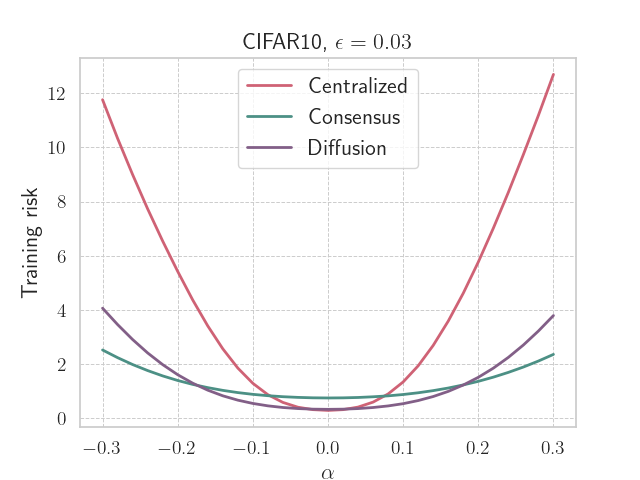}}
\caption{{Flatness visualization of models for CIFAR10.  We use the visualization method from \cite{Li0TSG18}, according to which we compute the average training risk value $J(w + \alpha\boldsymbol{v})$ over different random directions $\boldsymbol{v}$ that match the norm of $w$. Wider valleys correspond to flatter minima.}}
\label{loss_visual}
\end{figure}

\subsubsection{Heterogeneous networks}
It is possible to consider situations where the perturbations at the agents are bounded by different norms $p_k$ or attack strengths $\epsilon_k$. This motivates us to study the robustness in heterogeneous settings. We simulate two kinds of heterogeneity in the experiment with the proposed adversarial diffusion strategy and adversarial consensus strategy separately. In the first case, which is referred to \textit{clean+adv}, half of the agents are trained adversarially while the other half is trained in the normal manner with clean samples. In the second type of heterogeneous networks, which are referred to \textit{$\ell_2$+$\ell_\infty$}, half of the agents receive attacks generated from the $\ell_2$ norm, while the other half are attacked by $\ell_\infty$ perturbations.

\begin{figure*}[htbp]
\centering
\subfigure[MNIST: clean+adv, $\ell_2$]{\label{sub_1_7}\includegraphics[width=4.7cm]{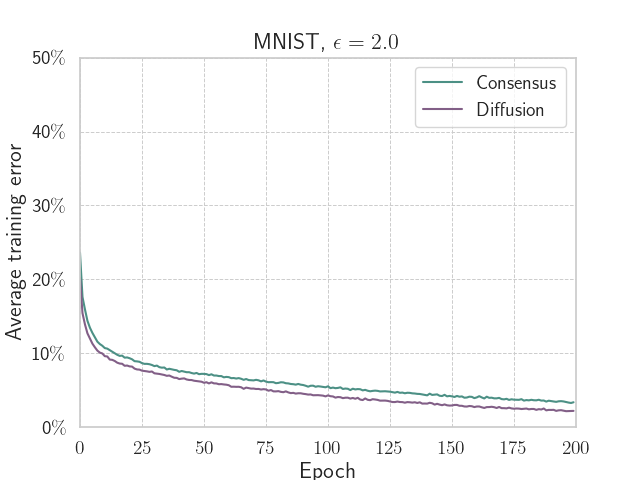}}
\subfigure[MNIST: clean+adv, $\ell_\infty$]{\label{sub_2_7}\includegraphics[width=4.7cm]{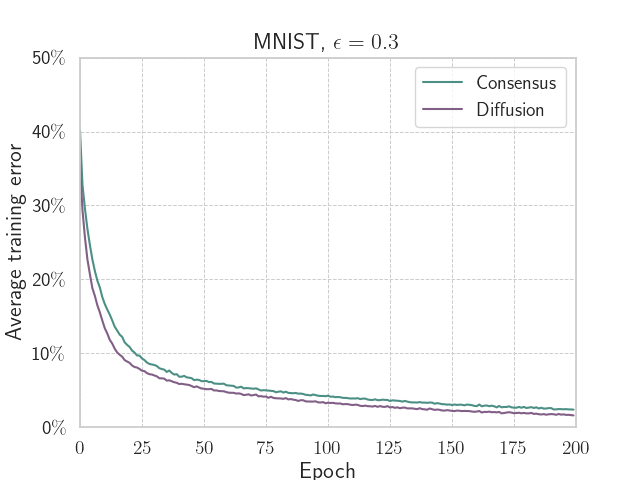}}
\subfigure[MNIST: $\ell_2$+$\ell_\infty$]{\label{sub_3_7}\includegraphics[width=4.7cm]{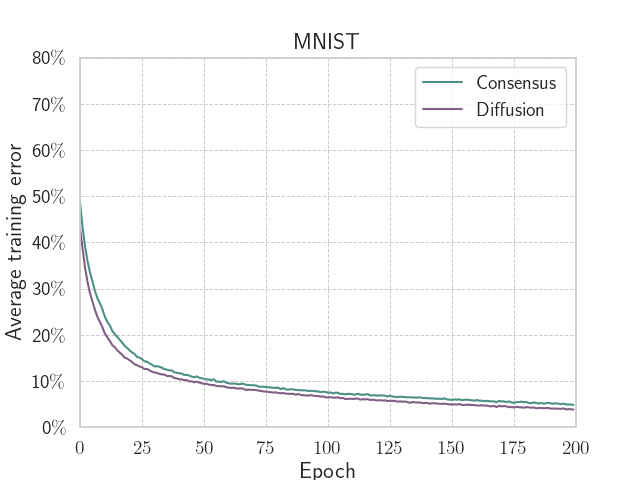}}

\subfigure[CIFAR10: clean+adv, $\ell_2$]{\label{sub_5_7}\includegraphics[width = 4.7cm]{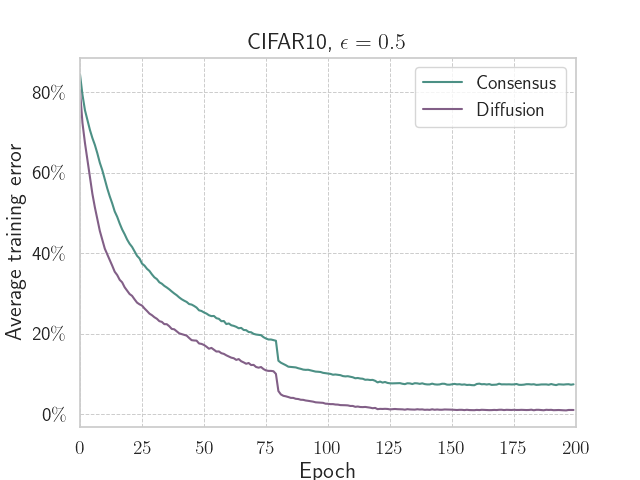}}
\subfigure[CIFAR10: clean+adv, $\ell_\infty$ ]{\label{sub_6_7}\includegraphics[width = 4.7cm]{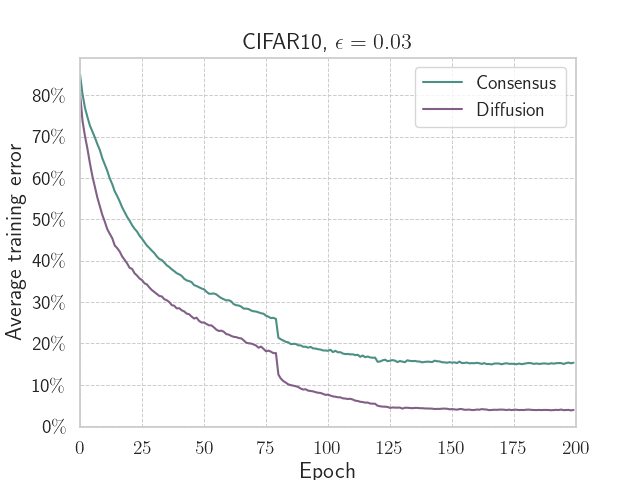}}
\subfigure[CIFAR10: $\ell_2$+$\ell_\infty$]{\label{sub_7_7}\includegraphics[width = 4.7cm]{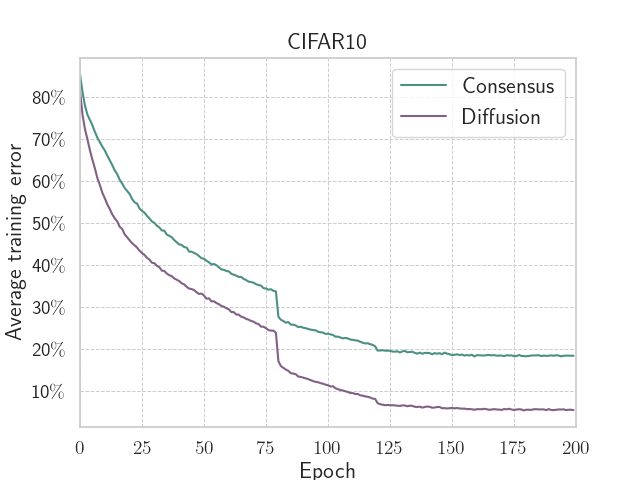}}

\caption{{Convergence plots for the two datasets in non-convex and heterogeneous environments. (a) The evolution of the average training error across the graph topology when training clean+adv networks with clean data and $\ell_2$ attacks bounded by $\epsilon_k = 2.0$ for MNIST. (b) The evolution of the average training error when training clean+adv networks with clean data and $\ell_\infty$ attacks bounded by $\epsilon_k = 0.3$ for MNIST. (c) The evolution of the average classification error when training the $\ell_2$+$\ell_\infty$  networks for MNIST. (d)--(f) follow a similar logic to (a)--(c) except now the dataset is CIFAR10.
}}
\label{fig7}
\end{figure*}

To train both {two kinds of heterogeneous networks}, we repeat the procedure from training the homogeneous networks except that the upper bounds $\epsilon_k$ or the $\ell_{p_k}$ norms may vary over the graph. The convergence plots are shown in Fig. \ref{fig7}, where we show the evolution of the {average training classification error} across all agents in the training process. 

\begin{table*}[h!t]
\scriptsize
   \centering
    \begin{spacing}{1.5}
   \caption{{The classification error of heterogeneous networks to various attacks.}}
   \label{tab3}
\begin{tabular}{c c c c c c c c c c c c c c c}
\toprule
dada&MNIST&MNIST&MNIST&MNIST&MNIST&MNIST&MNIST&MNIST&MNIST&MNIST&MNIST&MNIST\\
attack&PGM&DeepFool&FMN&PGD&DeepFool&FMN&PGM&DeepFool&FMN&PGD&DeepFool&FMN\\
model&clean+adv&clean+adv&clean+adv&clean+adv&clean+adv&clean+adv&$\ell_2$+$\ell_\infty$&$\ell_2$+$\ell_\infty$&$\ell_2$+$\ell_\infty$&$\ell_2$+$\ell_\infty$&$\ell_2$+$\ell_\infty$&$\ell_2$+$\ell_\infty$&\\
$\epsilon$&2&2&2&0.3&0.3&0.3&2&2&2&0.3&0.3&0.3\\
{error-consen}&14.45\%&14.97\%&18.32\%&11.21\%&8.04\%&9.82\%&10.06\%&8.52\%&11.33\%&19.08\%&10.4\%&13.64\%\\
error-diff&15.01\%&15.26\%&18.99\%&11.14\%&8.53\%&10.28\%&9.81\%&8.39\%&11.29\%&22.42\%&10.66\%&14.37\%\\
\midrule
dada&CIFAR&CIFAR&CIFAR&CIFAR&CIFAR&CIFAR&CIFAR&CIFAR&CIFAR&CIFAR&CIFAR&CIFAR\\
attack&PGM&DeepFool&FMN&PGD&DeepFool&FMN&PGM&DeepFool&FMN&PGD&DeepFool&FMN\\
model&clean+adv&clean+adv&clean+adv&clean+adv&clean+adv&clean+adv&$\ell_2$+$\ell_\infty$&$\ell_2$+$\ell_\infty$&$\ell_2$+$\ell_\infty$&$\ell_2$+$\ell_\infty$&$\ell_2$+$\ell_\infty$&$\ell_2$+$\ell_\infty$&\\
$\epsilon$&0.5&0.5&0.5&0.03&0.03&0.03&0.5&0.5&0.5&0.03&0.03&0.03\\
{error-consen}&48.11\%&41.66\%&39.52\%&70.74\%&59.01\%&69.02\%&35.53\%&32.89\%&31.15\%&63.38\%&55.26\%&62.30\%\\
error-diff&46.70\%&39.20\%&38.12\%&69.01\%&53.07\%&67.71\%&37.57\%&33.36\%&31.57\%&65.94\%&52.56\%&64.06\%\\
\bottomrule
\end{tabular}
\end{spacing}
\end{table*}

The robustness of the heterogeneous networks are illustrated by Fig. \ref{fig6} and Table \ref{tab3}. The green curves in Fig. \ref{fig6} show the robustness evolution of the clean+adv networks to the trained perturbations against different attack strength and the purple curves correspond to the $\ell_2+\ell_\infty$ networks. Comparing the results of the clean+adv and traditional clean networks in Fig. \ref{fig6}, from Table \ref{tab2} and Table \ref{tab3}, it is obvious that the former systems perform better than the latter ones. Thus, due to cooperation across the networks, the adversarial agents help improve the robustness of clean agents in the clean+adv networks. Likewise, comparing the results between the homogeneous and clean+adv networks, we observe some improvement for the clean accuracy in the clean+adv networks while their robustness deteriorates since now only a half of the agents are trained adversarially. This observation is associated with the trade-off between clean accuracy and robustness \cite{zhang2019theoretically,PangLYZY22}, which is still an open issue even in the single-agent case.  Similarly, as for the performance of the $\ell_2+\ell_\infty$ networks, we observe that this kind of heterogeneous networks are still more robust than the traditional clean ones. Moreover, their robustness to  $\ell_2$ attacks are enhanced by the $\ell_\infty$ agents. However, their robustness to $\ell_\infty$ agents decreases. This phenomenon relates to the trade-off among multiple perturbations \cite{tramer2019adversarial, maini2020adversarial} in single-agent learning. 

\section{Conclusion}
In this paper, we proposed a {general decentralized defense mechanism} for adversarial attacks over graphs. We analyzed the different convergence properties of the proposed framework {for strongly-convex, convex, and non-convex losses}. We further illustrated the behavior of the trained networks to perturbations generated by different attack methods, and demonstrated the enhanced robust behavior. Although we motivate our framework by focusing on stochastic gradient implementations, we hasten to add that other more complex optimizers can be used as well. Future directions can be on the trade-off problems in heterogeneous networks, which add more flexibility to the system than traditional single-agent learning.


%

\appendices


\section{Analytical solutions for optimal perturbations for convex loss functions}\label{ap_deltas}
In this section, we list the analytical solutions for the optimal perturbations corresponding to some popular (strongly)convex functions.
\subsection*{A.1. Logistic regression}
Consider the logistic loss:
\begin{align}
     Q(w; {x}+\delta, {y} ) = \mathrm{log} (1 + e^{-yw^{\sf T}x})+\rho \Vert w \Vert^2
\end{align}
where $\rho\ge 0$, and $y = \left\{+1, -1\right\}$ is the label of the data. We know that the function 
\begin{align}
    f(d) = \mathrm{log} (1 + e^{-d})
\end{align}
is decreasing since its gradient is always negative. Thus, we have
\begin{align}
  {\delta}^{\star} = \mathop{\mathrm{argmax}}\limits_{\Vert \delta \Vert_p \le \epsilon} Q(w; {x}+\delta, {y} ) =   \mathop{\mathrm{argmin}}\limits_{\Vert \delta \Vert_p \le \epsilon} y w^{\sf T} x
\end{align}
according to which we get
\begin{align}
\delta^{\star} = \left\{\begin{array}{cc}
-\epsilon y\frac{w}{\Vert w\Vert}, & if\;p = 2\\
-\epsilon y \mathrm{sign}(w), & if\;p = \infty
\end{array}\right. 
\end{align}
\subsection*{A.2. Exponential loss}
Consider the exponential loss:
\begin{align}
     Q(w; {x}+\delta, {y} ) = e^{-yw^{\sf T}x}+\rho \Vert w \Vert^2
\end{align}
Since the function 
\begin{align}
    f(d) =  e^{-d}
\end{align}
is also decreasing, the analytical solutions for $\delta^\star$ are the same as the logistic loss.
\subsection*{A.3. Least-mean-squares loss}
Consider the least-mean-squares loss:
\begin{align}
    Q(w; {x}+\delta, {y} ) = \Vert w^{\sf T}({x}+\delta)-{y}\Vert^2+\rho \Vert w \Vert^2
\end{align}
We have
\begin{align}
    {\delta}^{\star} =& \mathop{\mathrm{argmax}}\limits_{\Vert \delta \Vert_p \le \epsilon} Q(w; {x}+\delta, {y} ) \notag\\
    =& \mathop{\mathrm{argmax}}\limits_{\Vert \delta \Vert_p \le \epsilon} \ (w^{\sf T}\delta)^2 + 2(w^{\sf T}\delta)\cdot(w^{\sf T}x - y)
\end{align}
Let $d = w^{\sf T} \delta$ whose domain is closed and bounded since 
\begin{align}
   - \epsilon\Vert w \Vert_q \le - \Vert w \Vert_q \Vert \delta \Vert_p \le d = w^{\sf T} \delta \le \Vert w \Vert_q \Vert \delta \Vert_p \le \epsilon\Vert w \Vert_q
\end{align} 
Note that the function
\begin{align}
    f(d) = d^2 + 2d(w^{\sf T}x - y)
\end{align}
is strongly-convex over $d$ with
\begin{align}
f'(d) = 2d + 2(w^{\sf T}x - y)
\end{align}
Thus, when $f'(d) = 0$, i.e., $d = y - w^{\sf T}x$, $f(d)$ achieves the minimum. Furthermore, to achieve the maximum of $f(d)$, it is easy to get
\begin{align}
    d^{\star} = \epsilon \Vert w \Vert_q \mathrm{sign}(w^{\sf T}x -y)
\end{align}
from which the analytical expression for $\delta^\star$ can be obtained:
\begin{align}
\delta^{\star} = \left\{\begin{array}{cc}
\epsilon\mathrm{sign}(w^{\sf T}x -y)\frac{w}{\Vert w\Vert}, & if\;p = 2\\
\epsilon\mathrm{sign}(w^{\sf T}x -y)\mathrm{sign}(w), & if\;p = \infty
\end{array}\right. 
\end{align}
\subsection*{A.4. Huber loss}
Consider the Huber loss:
\begin{align}
     &Q(w; {x}+\delta, {y} ) \notag\\
     &= \left\{\begin{array}{cc}
\frac{1}{2}(w^{\sf T}(x+\delta) - y)^2,& if\;  \vert w^{\sf T}(x+\delta) - y\vert\le\tau \\
\tau \vert w^{\sf T} (x+\delta) - y\vert - \frac{\tau^2}{2},& otherwise
\end{array}\right. 
\end{align}
Note that 
\begin{align}
    \mathop{\mathrm{argmax}}\limits_{\Vert\delta\Vert_p \le \epsilon} \ \tau \vert w^{\sf T} (x+\delta) - y\vert - \frac{\tau^2}{2} = \mathop{\mathrm{argmax}}\limits_{\Vert\delta\Vert_p \le \epsilon} (w^{\sf T} (x+\delta) - y)^2
\end{align}
Thus, independent of whether $\vert w^{\sf T}(x+\delta) - y\vert$ is larger than $\tau$ or not, it always holds that
\begin{align}
   \mathop{\mathrm{argmax}}\limits_{\Vert\delta\Vert_p \le \epsilon} Q(w; {x}+\delta, {y} ) = \mathop{\mathrm{argmax}}\limits_{\Vert\delta\Vert_p \le \epsilon} (w^{\sf T} (x+\delta) - y)^2
\end{align}
Therefore, we conclude that the analytical solutions for the optimal perturbations corresponding to the Huber loss are the same with the least mean square loss.

\section{Proof of lemma \ref{affine_l}}\label{ap1}
\begin{proof}
Before proving the affine Lipshitz condition, we first note that the $\ell_2$ distance between any two possible perturbations $\delta_{k,1}$ and $\delta_{k,2}$ from the region $\Vert\delta\Vert_{p_k}\le \epsilon_k$ can be bounded {in view of} the equivalence of vector norms. Specifically, for any two vector norms $\Vert\cdot\Vert_a$ and $\Vert\cdot\Vert_b$ defined {over a} finite-dimensional space, there exist real numbers $0<c_1\le c_2$ such that for any $\delta \in \mathbbm{R}^M$: $c_1\Vert\delta\Vert_a \le \Vert\delta\Vert_b \le c_2\Vert\delta\Vert_a$,
according to which we say that for any $\ell_p$ norm, if $\Vert\delta\Vert_p$ is bounded by $\epsilon_k$, then its $\ell_2$ norm $\Vert\delta\Vert$ is bounded by $O(\epsilon_k)$. For example, when $p_k = 1$, we have
\begin{align}
   \Vert \delta_{k,2} - \delta_{k,1}\Vert \le \Vert \delta_{k,2} - \delta_{k,1}\Vert_1 \le 2\epsilon_k
\end{align}
{and when} $p_k = \infty$, we have
\begin{align}
   \Vert \delta_{k,2} - \delta_{k,1}\Vert \le \sqrt{M}\Vert \delta_2 - \delta_1\Vert_\infty \le 2\sqrt{M}\epsilon_k
\end{align}
Furthermore, consider the maximum perturbation bound across all agents and denote it by $\epsilon = \max\limits_{k}\epsilon_k$. Then, the $\ell_2$ distance between any two possible perturbations $\delta_k$ and $\delta_\ell$ from the regions $\Vert\delta\Vert_{p_k}\le \epsilon_k$ and  $\Vert\delta\Vert_{p_\ell}\le \epsilon_\ell$ can also be bounded even if $p_k \neq p_\ell$ and $\epsilon_k \neq \epsilon_\ell$. For example, if $p_k = \infty$ and $p_\ell = 2$, we have
\begin{align}
    \Vert \delta_k - \delta_\ell\Vert \le \Vert\delta_k\Vert + \Vert\delta_\ell\Vert \le \sqrt{M}\epsilon_k + \epsilon_\ell = O(\epsilon)
\end{align}
Thus, for later use, no matter which perturbation norm is used, we can assume for any agents $k$ and $\ell$ that the Euclidean norm of the respective perturbations satisfy
\begin{align}\label{e_p_1}
\Vert \delta_{k,2} - \delta_{k,1}\Vert \le O(\epsilon),\;\Vert \delta_k - \delta_\ell\Vert \le  O(\epsilon)
\end{align}
Next, for any  $w_1, w_2 \in \mathbbm{R}^M$, and $\delta_1, \delta_2 \in \mathbbm{R}^M$ satisfying 
\begin{align}
    \Vert\delta_1\Vert_{p_k}\le \epsilon_k,\quad \Vert\delta_2\Vert_{p_k}\le \epsilon_k
\end{align}
we have
\begin{align}\label{aaf_a}
&\Vert \nabla_w  Q_k(w_2; \boldsymbol{x}_k+\delta_2,\boldsymbol{y}_k) - \nabla_w Q_k(w_1; \boldsymbol{x}_k+\delta_1,\boldsymbol{y}_k)\Vert \notag\\
&\le \Vert\nabla_w Q_k(w_2;\boldsymbol{x}_k+\delta_2, \boldsymbol{y}_k) - \nabla_w Q_k(w_1;\boldsymbol{x}_k+\delta_2, \boldsymbol{y}_k)\Vert \notag\\
&\quad\; + \Vert\nabla_w Q_k(w_1;\boldsymbol{x}_k+\delta_2, \boldsymbol{y}_k) - \nabla_w Q_k(w_1;\boldsymbol{x}_k+\delta_1, \boldsymbol{y}_k)\Vert \notag\\
& \overset{(a)}{\le}  L\Vert w_2 - w_1 \Vert + L \Vert \delta_2 - \delta_1 \Vert = L\Vert w_2 - w_1 \Vert + O(\epsilon)
\end{align}
where $(a)$ follows from the smoothness conditions in Assumption \ref{as2}.  In the case where the maximizer of $Q_k(w;\boldsymbol{x}_{k} +{\delta},\boldsymbol{y}_{k})$ with respect to $\delta$ is unique and accessible, we consider the problems
\begin{align}
  &\boldsymbol{\delta}^\star_{2} =  \mathop{\rm{argmax}}\limits_{\left\Vert{\delta}\right\Vert_{p_k}  \le \epsilon_k}   Q_k(w_2;\boldsymbol{x}_{k} +{\delta},\boldsymbol{y}_{k})\notag\\
  &\boldsymbol{\delta}^\star_{1} =  \mathop{\rm{argmax}}\limits_{\left\Vert{\delta}\right\Vert_{p_k}  \le \epsilon_k}   Q_k(w_1;\boldsymbol{x}_{k} +{\delta},\boldsymbol{y}_{k})
\end{align}
In this case, Danskin's theorem guarantees that $f_k(w;\boldsymbol{x}_k,\boldsymbol{y}_k)$ is differentiable. Similar to (\ref{aaf_a}), it is easy to obtain:
\begin{align}\label{affine_l_1}
  &\Vert\nabla_w f_k(w_2;\boldsymbol{x}_k, \boldsymbol{y}_k) - \nabla_w f_k(w_1;\boldsymbol{x}_k, \boldsymbol{y}_k)\Vert \notag\\
  &\le \Vert \nabla_w  Q_k(w_2; \boldsymbol{x}_k+\boldsymbol{\delta}_2^\star,\boldsymbol{y}_k) - \nabla_w Q_k(w_1; \boldsymbol{x}_k+\boldsymbol{\delta}_1^\star,\boldsymbol{y}_k)\Vert\notag\\
  & \le  L\Vert w_2 - w_1 \Vert + O(\epsilon)
\end{align}
Then, exchanging the expectation and differentiation operations (as permitted by the dominated convergence theorem in our cases of interest [4]) gives 
\begin{align}\label{al_1}
    &\Vert \nabla_w J_k(w_2) - \nabla_w J_k(w_1) \Vert \notag\\
    &\overset{(a)}{\le}  \mathds{E}\Vert \nabla_w  f_k(w_2; \boldsymbol{x}_k,\boldsymbol{y}_k) - \nabla_w f_k(w_1; \boldsymbol{x}_k,\boldsymbol{y}_k)\Vert  \notag\\
    &\le   L\Vert w_2 - w_1 \Vert + O(\epsilon)
\end{align}
where $(a)$ follows from (\ref{J_k_f}).
\end{proof}

\section{Proof of Theorem \ref{thddw}}\label{ap_w_dd}
\begin{proof}
{
 We now verify the convergence of $\mathds{E}\Vert\ddot{\boldsymbol{w}}_{n}\Vert^2$. Recalling (\ref{AC_15})  gives
\begin{align}\label{AC_17}
\mathds{E}\Vert\ddot{\boldsymbol{w}}_{n}\Vert^2& = \mathds{E} \left\Vert\mathcal{J}_\alpha^{\sf T}\ddot{\boldsymbol{w}}_{n-1} + \mu\mathcal{V}_R^{\sf T}\mathcal{A}_2^{\sf T}\boldsymbol{q}_n\right\Vert^2
\end{align} }  
Consider $t$ as follows, and assume a small enough $\alpha$ such that (see \cite[ch.9]{sayed2014adaptation}): 
    \begin{align}\label{AC_18}
        t = \Vert J_{\alpha}^{\sf T}\Vert = \sqrt{\tau (J_\alpha J_\alpha^{\sf T})} = \lambda_2 + \alpha < 1
    \end{align}
{Then, it follows from the Jensen's inequality that
\begin{align}\label{AC_19}
  \mathds{E}\Vert\ddot{\boldsymbol{w}}_{n}\Vert^2 \le t\mathds{E}\Vert\ddot{\boldsymbol{w}}_{n-1}\Vert^2  + \frac{\mu^2 t^2}{1-t}\Vert\mathcal{V}_R^{\sf T}\mathcal{A}_2^{\sf T}\boldsymbol{q}_n\Vert^2
\end{align}}

To prove the convergence of $\mathds{E}\Vert\ddot{\boldsymbol{w}}_{n}\Vert^2$, we have to examine the bound of the second term on the right-hand side of (\ref{AC_19}). Recalling (\ref{a1a2p}) and (\ref{AC_12a}), we have
\begin{align}
    V_{R}^{\sf T}A_2^{\sf T}\mathbbm{1} = V_{R}^{\sf T}\mathbbm{1} = 0
\end{align}
Combining with (\ref{qnhat}), we obtain
\begin{align}\label{AC_20}
   &\left\Vert\mathcal{V}_R^{\sf T}{\mathcal{A}_2^{\sf T}\boldsymbol{q}_n}\right\Vert^2  \notag\\
   &{=}  \left\Vert \mathcal{V}_R^{\sf T}{\mathcal{A}_2^{\sf T}\boldsymbol{q}_n} -\mathcal{V}_R^{\sf T}{\mathcal{A}_2^{\sf T}}(\mathbbm{1}\pi^{\sf T}\otimes I_M){\boldsymbol{q}_n} \right\Vert^2\notag\\
   & {\le}  \Vert\mathcal{V}_R^{\sf T}{\mathcal{A}_2^{\sf T}}\Vert^2 \bigg(\sum_{k=1}^{K} \bigg\Vert \frac{1}{B}\sum_b \nabla_w Q_k(\boldsymbol{w}_{k,n -1}; \widehat{\boldsymbol{x}}_{k,n}^b,\boldsymbol{y}_{k,n}^b) \notag\\
   &\quad\;- \sum_{\ell=1}^{K}\pi_l\cdot \frac{1}{B}\sum_b\nabla_w Q_\ell(\boldsymbol{w}_{\ell,n -1}; \widehat{\boldsymbol{x}}_{\ell,n}^b,\boldsymbol{y}_{\ell,n}^b)\bigg\Vert^2\bigg)\notag\\
    & \overset{(a)}{\le} \Vert \mathcal{V}_R^{\sf T}{\mathcal{A}_2^{\sf T}}\Vert^2 \frac{1}{B}\sum_b \sum_k \sum_\ell \pi_\ell \Vert\nabla_w Q_k(\boldsymbol{w}_{k,n-1}; \widehat{\boldsymbol{x}}_{k,n}^b,\boldsymbol{y}_{k,n}^b) \notag\\
   &\quad\;- \nabla_w Q_\ell(\boldsymbol{w}_{\ell,n -1}; \widehat{\boldsymbol{x}}_{\ell,n}^b,\boldsymbol{y}_{\ell,n}^b)\Vert^2
\end{align}
{where $(a)$ follows from Jensen's inequality, and the gradient disagreement between any two agents can be bounded as follows:
\begin{align}\label{AC_21}
 &\mathds{E}\left\Vert\nabla_w Q_k(\boldsymbol{w}_{k,n -1}; \widehat{\boldsymbol{x}}_{k,n}^b,\boldsymbol{y}_{k,n}^b) - \nabla_w Q_\ell(\boldsymbol{w}_{\ell,n -1}; \widehat{\boldsymbol{x}}_{\ell,n}^b,\boldsymbol{y}_{\ell,n}^b)\right\Vert^2 \notag\\
 &= \mathds{E} \Vert\nabla_w Q_k(\boldsymbol{w}_{k,n -1}; \widehat{\boldsymbol{x}}_{k,n}^b,\boldsymbol{y}_{k,n}^b) - \nabla_w Q_k(\boldsymbol{w}_{\ell,n-1}; \widehat{\boldsymbol{x}}_{k,n}^b,\boldsymbol{y}_{k,n}^b) \notag\\
 &\quad+ \nabla_w Q_k(\boldsymbol{w}_{\ell,n-1}; \widehat{\boldsymbol{x}}_{k,n}^b,\boldsymbol{y}_{k,n}^b) - \nabla_w Q_k(\boldsymbol{w}_{\ell,n-1}; \widehat{\boldsymbol{x}}_{\ell,n}^b,\boldsymbol{y}_{k,n}^b)\notag\\
 & \quad+ \nabla_w Q_k(\boldsymbol{w}_{\ell,n-1}; \widehat{\boldsymbol{x}}_{\ell,n}^b,\boldsymbol{y}_{k,n}^b) - 
 \nabla_w Q_k(\boldsymbol{w}_{\ell,n-1}; \widehat{\boldsymbol{x}}_{\ell,n}^b,\boldsymbol{y}_{\ell,n}^b)\notag\\
 &\quad + \nabla_w Q_k(\boldsymbol{w}_{\ell,n-1}; \widehat{\boldsymbol{x}}_{\ell,n}^b,\boldsymbol{y}_{\ell,n}^b) - \nabla_w Q_\ell(\boldsymbol{w}_{\ell,n-1}; \widehat{\boldsymbol{x}}_{\ell,n}^b,\boldsymbol{y}_{\ell,n}^b)\Vert^2\notag\\
 &\overset{(a)}{\le} 4L^2\mathds{E}\Vert\boldsymbol{w}_{k,n-1}-\boldsymbol{w}_{\ell,n-1}\Vert^2 + 8L^2\mathds{E}\Vert\widehat{\boldsymbol{\delta}}_{k,n}^b - \widehat{\boldsymbol{\delta}}_{\ell,n}^b\Vert^2 \notag\\
&\quad\;+8L^2\mathds{E}\Vert\boldsymbol{x}_{k,n}^b - \boldsymbol{x}_{\ell,n}^b\Vert^2 + 4L^2 \mathds{E}\Vert\boldsymbol{y}_{k,n}^b-\boldsymbol{y}_{\ell,n}^b\Vert^2 + 4C^2\notag\\
&\le 4L^2\mathds{E}\Vert\boldsymbol{w}_{k,n-1}-\boldsymbol{w}_{\ell,n-1}\Vert^2 + a^2\notag\\
&\le 8L^2\mathds{E}\Vert\boldsymbol{w}_{k,n-1}-\boldsymbol{w}_{c,n-1}\Vert^2 + 8L^2\mathds{E}\Vert\boldsymbol{w}_{\ell,n-1}-\boldsymbol{w}_{c,n-1}\Vert^2 \notag\\
&\quad\;+ a^2
\end{align}
where the non-iid data over space is considered, and $(a)$ follows from Jensen's inequality and Assumption \ref{as4}, and the constant $a^2$ admits
\begin{align}
    a^2 \overset{\Delta}{=} &8L^2\mathds{E}\Vert\boldsymbol{x}_{k,n}^b - \boldsymbol{x}_{\ell,n}^b\Vert + 4L^2 \mathds{E}\Vert\boldsymbol{y}_{k,n}^b-\boldsymbol{y}_{\ell,n}^b\Vert^2 + 4C^2 \notag\\
&+ O(\epsilon^2)
\end{align}
Moreover, $\widehat{\boldsymbol{\delta}}_{k,n}^b$ and $\widehat{\boldsymbol{\delta}}_{\ell,n}^b$ are the approximate maximizer defined by:
\begin{align}\label{AC_23}
  &\widehat{\boldsymbol{\delta}}_{k,n}^b \approx  \mathop{\rm{argmax}}\limits_{\left\Vert{\delta}\right\Vert_{p_k} \le \epsilon_k}   Q_k(\boldsymbol{w}_{k,n-1};\boldsymbol{x}_{k,n}^b +\delta,\boldsymbol{y}_{k,n}^b)\notag\\
  &\widehat{\boldsymbol{\delta}}_{\ell,n}^b \approx  \mathop{\rm{argmax}}\limits_{\left\Vert{\delta}\right\Vert_{p_k} \le \epsilon_k}   Q_\ell(\boldsymbol{w}_{\ell,n-1};\boldsymbol{x}_{\ell,n}^b +\delta,\boldsymbol{y}_{\ell,n}^b)
\end{align}
Substituting (\ref{AC_21}) into (\ref{AC_20}), and taking expectations on both sides, we get
\begin{align}\label{AC_25}
    &\mathds{E}\left\Vert\mathcal{V}_R^{\sf T}\mathcal{A}_2^{\sf T}\boldsymbol{q}_n\right\Vert^2  \notag\\
    &\le \Vert\mathcal{V}_R^{\sf T}\mathcal{A}_2^{\sf T}\Vert^2\cdot\frac{1}{B}\sum_b\sum_k\sum_\ell \pi_\ell \big(8L^2 \mathds{E}\Vert\boldsymbol{w}_{k,n-1} - \boldsymbol{w}_{c,n-1} \Vert^2 \notag\\
    &\quad\;+ 8L^2\mathds{E}\Vert\boldsymbol{w}_{\ell,n-1} - \boldsymbol{w}_{c,n-1} \Vert^2 + a^2\big)\notag\\
    &\le \Vert\mathcal{V}_R^{\sf T}\mathcal{A}_2^{\sf T}\Vert^2 (8L^2+8KL^2)\mathds{E}\Vert \boldsymbol{\scriptstyle\mathcal{W}}_{n-1} - \mathbbm{1}\otimes \boldsymbol{w}_{c,n-1}\Vert^2 \notag\\
    &\quad\;+ K\Vert\mathcal{V}_R^{\sf T}\mathcal{A}_2^{\sf T}\Vert^2a^2\notag\\
    &\overset{(a)}{\le} \Vert\mathcal{V}_R^{\sf T}\mathcal{A}_2^{\sf T}\Vert^2 \Vert V_L\Vert^2(8L^2+8KL^2)\mathds{E}\Vert\ddot{\boldsymbol{w}}_{n-1}\Vert^2 \notag\\
    &\quad\;+ K\Vert\mathcal{V}_R^{\sf T}\mathcal{A}_2^{\sf T}\Vert^2a^2
\end{align}
where $(a)$ follows from \eqref{net_dis_convex}.}

Note that when the optimal perturbation is used in $\boldsymbol{q}_n$ (as defined in \eqref{qnstar}), by using the similar techniques from \eqref{AC_20}--\eqref{AC_21}, \eqref{AC_25} still hold.

Now substituting (\ref{AC_25}) into (\ref{AC_19}) gives
\begin{align}\label{AC_29}
&\mathds{E}\Vert\ddot{\boldsymbol{w}}_{n}\Vert^2 \notag\\
&\le t\mathds{E}\Vert\boldsymbol{\ddot{w}}_{n-1}\Vert^2 + \frac{\mu^2t^2}{1-t}\Vert\mathcal{V}_R^{\sf T}\mathcal{A}_2^{\sf T}\Vert^2Ka^2 \notag\\
&\quad\;+  \frac{\mu^2t^2}{1-t}\Vert\mathcal{V}_R^{\sf T}\mathcal{A}_2^{\sf T}\Vert^2\Vert\mathcal{V}_L\Vert^2(8L^2+8KL^2)\mathds{E}\Vert\ddot{\boldsymbol{w}}_{n-1}\Vert^2  \notag\\
    & = \Big(t + O(\mu^2)\Big)\mathds{E}\Vert\ddot{\boldsymbol{w}}_{n-1}\Vert^2 + O(\mu^2)
\end{align}
where $t+O(\mu^2) < 1 $ when $\mu$ is sufficiently small. Then, iterating recursion (\ref{AC_29}) after enough iterations $n \ge \ddot{n}$ with
\begin{align}\label{checkn}
    {\ddot{n}} = O\left(\frac{\log \mu}{\log (t+O(\mu^2))}\right)
\end{align}
the term $\mathds{E}\left\Vert \ddot{\boldsymbol{w}}_{n}\right\Vert^2$ is bounded:
\begin{align}\label{AC_30}
 \mathds{E}\left\Vert {\ddot{\boldsymbol{w}}_{n}}\right\Vert^2 \le \mu^2{\ddot{c}_{\epsilon}^2} = O(\mu^2)
\end{align}
where 
\begin{align}\label{AC_31}
    \ddot{c}_{\epsilon} ^2=& \frac{2t^2\Vert\mathcal{V}_R^{\sf T}\mathcal{A}_2^{\sf T}\Vert^2Ka^2}{(1-t)(1-t-O(\mu^2))}
\end{align}

Furthermore, according to \eqref{net_dis_convex} and (\ref{AC_30}), the network disagreement is verified to be uniformly bounded:
\begin{align}\label{c_w_c}
    \mathds{E}\Vert {\boldsymbol{\scriptstyle\mathcal{W}}}_{n} - {\boldsymbol{\scriptstyle\mathcal{W}}}_{c,n} \Vert ^2 \overset{(a)}{\le}& \Vert \mathcal{V}_L\Vert^2\mathds{E}\left\Vert \ddot{\boldsymbol{w}}_{n}\right\Vert^2 \notag\\
    \le& \mu^2\Vert\mathcal{V}_L\Vert^2\ddot{c}_{\epsilon}^2\notag\\
    =& O(\mu^2)
\end{align}
where $(a)$ follows from (\ref{net_dis_convex}).
\end{proof}

\section{Proof of Lemma \ref{p_gn}}\label{ap2}
\begin{proof}
We first verify that the gradient noise has zero mean. Basically, {when the approximate maximizer is used, we have}
\begin{align}\label{sb_1}
    \mathds{E}\left\{\boldsymbol{s}_{k,n}|\boldsymbol{\mathcal{F}}_{n-1}\right\}  =&  \mathds{E} \big\{\nabla_w Q_k(\boldsymbol{w}_{k,n-1};{\widehat{\boldsymbol{x}}_{k,n}},\boldsymbol{y}_{k,n}) \notag\\
    &- {\mathds{E}\nabla_w Q_k(\boldsymbol{w}_{k,n-1};\widehat{\boldsymbol{x}}_{k,n},\boldsymbol{y}_{k,n})|\boldsymbol{\mathcal{F}}_{n-1}}\big\}\notag\\
    =&  0
\end{align}
Then, for the mini-batch gradient noise, we have
\begin{align}\label{sb0}
    \mathds{E}\Big\{\boldsymbol{s}^{B}_{k,n}|\boldsymbol{\mathcal{F}}_{n-1}\Big\}   =  \frac{1}{B}\sum\limits_{b=1}^{B}\mathds{E} \left\{\boldsymbol{s}_{k,n}|\boldsymbol{\mathcal{F}}_{n-1}\right\} = 0
\end{align}
{Similarly, if the optimal perturbation is used, it is obvious that (\ref{sb_1}) and (\ref{sb0}) are still true since the second term of the right-hand side in (\ref{dgnb_2}) is the expectation of the first term.}

Next we assess the variance of the gradient noise. {We start from (\ref{dgn}) in which the approximate maximizer is used:
\begin{align}\label{AB_2}
     &\mathds{E}\left\{\Vert \boldsymbol{s}_{k,n}\Vert^2|\boldsymbol{\mathcal{F}}_{n-1}\right\} \notag\\
     &=   \mathds{E} \big\{ \Vert\nabla_w Q_k(\boldsymbol{w}_{k,n-1};\widehat{\boldsymbol{x}}_{k,n},\boldsymbol{y}_{k,n}) \notag\\
     &\quad\;- \mathds{E}\nabla_w Q_k(\boldsymbol{w}_{k,n-1};\widehat{\boldsymbol{x}}_{k,n},\boldsymbol{y}_{k,n})\Vert^2|\boldsymbol{\mathcal{F}}_{n-1}\big\}\notag\\
     & \overset{(a)}{\le} 2 \mathds{E} \Vert\nabla_w Q_k(\boldsymbol{w}_{k,n-1};\widehat{\boldsymbol{x}}_{k,n},\boldsymbol{y}_{k,n})\Vert^2 \notag\\
     &\quad\;+ 2 \Vert\mathds{E} \nabla_w Q_k(\boldsymbol{w}_{k,n-1};\widehat{\boldsymbol{x}}_{k,n},\boldsymbol{y}_{k,n})\Vert^2\notag\\
    & \overset{(b)}{\le} 4\mathds{E}\Vert\nabla_w Q_k(\boldsymbol{w}_{k,n-1};\widehat{\boldsymbol{x}}_{k,n},\boldsymbol{y}_{k,n}) \notag\\
    &\quad\;- \nabla_w Q_k(w^{\star};\boldsymbol{x}_{k,n} +\boldsymbol{\delta}^{\star}(w^{\star}),\boldsymbol{y}_{k,n})\notag\\
    &\quad\;+ \nabla_w Q_k(w^{\star};\boldsymbol{x}_{k,n} +\boldsymbol{\delta}^{\star}(w^{\star}),\boldsymbol{y}_{k,n})\Vert^2 \notag\\
    & \overset{(c)}{\le} 8\mathds{E}\Vert\nabla_w Q_k(\boldsymbol{w}_{k,n-1};\widehat{\boldsymbol{x}}_{k,n},\boldsymbol{y}_{k,n}) - \nabla_w Q_k(w^{\star};\widehat{\boldsymbol{x}}_{k,n},\boldsymbol{y}_{k,n})+ \notag\\
    &\quad\; \nabla_w Q_k(w^{\star};\widehat{\boldsymbol{x}}_{k,n},\boldsymbol{y}_{k,n})  - \nabla_w Q_k(w^{\star};\boldsymbol{x}_{k,n} +\boldsymbol{\delta}^{\star}(w^{\star}),\boldsymbol{y}_{k,n})\Vert^2\notag\\
    &\quad\;+ 8\mathds{E}\Vert\nabla_w Q_k(w^{\star};\boldsymbol{x}_{k,n} +\boldsymbol{\delta}^{\star}(w^{\star}),\boldsymbol{y}_{k,n})\Vert^2\notag\\
    &\overset{(d)}{\le} 16L^2\Vert \boldsymbol{w}_{k,n-1} - w^\star\Vert^2 + O(\epsilon^2) \notag\\
    &\quad\;+ 8\mathds{E}\Vert \nabla_w Q_k(w^\star;\boldsymbol{x}_{k,n}+\boldsymbol{\delta}_k^\star(w^\star),\boldsymbol{y}_{k,n})\Vert^2
\end{align}
where $(a)$, $(b)$, and $(c)$  follow from Jensen's inequality, and $(d)$ follows from Assumption \ref{as2} and (\ref{e_p_1}). In addition, $\boldsymbol{\delta}_k^\star(w^\star)$ is the optimal perturbation evaluated at the global minimizer $w^\star$:
\begin{align}\label{delta_w_s}
\boldsymbol{\delta}_k^\star(w^\star) = \mathop{\rm{argmax}}\limits_{\left\Vert{\delta}\right\Vert_{p_k}  \le \epsilon_k}   Q_k(w^\star;\boldsymbol{x}_{k,n} +{\delta},\boldsymbol{y}_{k,n})
\end{align}}

Then for the mini-batch gradient noise, we have
\begin{align}\label{sb2}
        &\mathds{E}\left\{\left\Vert\boldsymbol{s}^{B}_{k,n}\right\Vert^2|\boldsymbol{\mathcal{F}}_{n-1}\right\}\notag\\
        & \overset{(a)}{=}\frac{1}{B^2}\sum\limits_{b=1}^{B}\mathds{E} \left\{\left\Vert\boldsymbol{s}_{k,n}\right\Vert^2|\boldsymbol{\mathcal{F}}_{n-1}\right\} \notag\\
    & \overset{(b)}{\le} \frac{16L^2}{B}\Vert \boldsymbol{w}_{k,n-1} - w^\star\Vert^2 + O(\epsilon^2)  \notag\\
    &\quad\;+ \frac{8}{B}\mathds{E}\Vert \nabla_w Q_k(w^\star;\boldsymbol{x}_{k,n}+\boldsymbol{\delta}_k^\star(w^\star),\boldsymbol{y}_{k,n})\Vert^2
\end{align}
where $(a)$ is due to the fact that all data at each agent are sampled independently, and $(b)$ follows from (\ref{AB_2}). 

{By following a derivation similar to (\ref{AB_2})--(\ref{sb2}), the variance of the gradient noise defined in (\ref{dgnb_2}), which is based on the optimal perturbation, can also be shown to be upper bounded. Specifically, the upper bound is identical to that in (\ref{sb2}).}
\end{proof}

\section{proof of Theorem \ref{th_mse}}\label{ap_th_mse}
\begin{proof}
We first prove the convergence of $\mathds{E}\Vert\check{\boldsymbol{w}}_{n}\Vert^2$. According to (\ref{check_dd_w}) and (\ref{AC_30}), for all $n\ge\ddot{n}$, it holds that
\begin{align}\label{AC_30_1}
 \mathds{E}\left\Vert \check{\boldsymbol{w}}_{n}\right\Vert^2 =  \mathds{E}\left\Vert \ddot{\boldsymbol{w}}_{n}\right\Vert^2 \le \mu^2\ddot{c}_{\epsilon}^2 = O(\mu^2)
\end{align}

We next examine the convergence of $\mathds{E}\Vert\bar{\boldsymbol{w}}_{n}\Vert^2$. Squaring and taking expectations of (\ref{w_bar_2}) conditioned on $\boldsymbol{\mathcal{F}}_{n-1}$ gives
\begin{align}\label{AC_32}
     &\mathds{E}\left\{\Vert\bar{\boldsymbol{w}}_{n}\Vert^2\vert\boldsymbol{\mathcal{F}}_{n-1}\right\}  \notag\\
     &\overset{(a)}{=}  \left\Vert\bar{\boldsymbol{w}}_{n-1} + \mu\sum_k \pi_k{\mathds{E}\nabla_w Q_k(\boldsymbol{w}_{k,n-1};\widehat{\boldsymbol{x}}_{k,n},\boldsymbol{y}_{k,n})} \right\Vert^2 \notag\\
     &\quad\;+ \mu^2\mathds{E}\left\{\left\Vert\sum_k \pi_k \boldsymbol{s}_{k,n}^B\right\Vert^2\vert\boldsymbol{\mathcal{F}}_{n-1}\right\}\notag\\
    &=   \Vert\bar{\boldsymbol{w}}_{n-1}\Vert^2 + 2\mu\sum_k\pi_k{\mathds{E}\nabla_{w^{\sf T}} Q_k(\boldsymbol{w}_{k,n-1};\widehat{\boldsymbol{x}}_{k,n},\boldsymbol{y}_{k,n})}\bar{\boldsymbol{w}}_{n-1} \notag\\
    &\quad\;+ \mu^2 \left\Vert \sum_k \pi_k {\mathds{E}\nabla_w Q_k(\boldsymbol{w}_{k,n-1};\widehat{\boldsymbol{x}}_{k,n},\boldsymbol{y}_{k,n})}\right\Vert^2 \notag\\
    &\quad\ + \mu^2\mathds{E}\left\{\left\Vert\sum_k \pi_k \boldsymbol{s}_{k,n}^B\right\Vert^2\vert\boldsymbol{\mathcal{F}}_{n-1}\right\}
\end{align}
where $(a)$ follows from (\ref{s0_convex}). 
We explore the right-hand terms of (\ref{AC_32}) one by one.  First, for the term associated with the gradient noise, we have
\begin{align}\label{AC_33}
     \mathds{E}\left\{\left\Vert\sum_k \pi_k \boldsymbol{s}_{k,n}^B\right\Vert^2\Big\vert\boldsymbol{\mathcal{F}}_{n-1}\right\} &\overset{(a)}{\le}  \sum_k\pi_k \mathds{E}\left\{\Vert\boldsymbol{s}_{k,n}^B\Vert^2\vert\boldsymbol{\mathcal{F}}_{n-1}\right\}\notag\\
     &\overset{(b)}{\le}  \beta_{\max}^2 \sum_k\pi_k\Vert\widetilde{\boldsymbol{w}}_{k,n-1}\Vert^2 + \sigma^2
\end{align}
where $(a)$ follows from Jensen's inequality, $(b)$ follows from (\ref{s2_convex}), and
\begin{align}\label{AC_34}
    \beta_{\max}^2 = \max\limits_{k}\{\beta_{k}^2\}, \quad \sigma^2 = \sum_k \pi_k\sigma_{k}^2
\end{align}

Second, by recalling (\ref{delta_w_s}), we have
\begin{align}\label{AC_36_2}
     &\left\Vert \sum_k \pi_k \mathds{E}\nabla_w Q_k(\boldsymbol{w}_{k,n-1};\widehat{\boldsymbol{x}}_{k,n},\boldsymbol{y}_{k,n})\right\Vert^2  \notag\\
     &\overset{(a)}{\le} \sum_k\pi_k \mathds{E}\left\Vert\nabla_w Q_k(\boldsymbol{w}_{k,n-1};\widehat{\boldsymbol{x}}_{k,n},\boldsymbol{y}_{k,n})\right\Vert^2\notag\\
     &\overset{(b)}{\le}\sum_k \pi_k\bigg(2\mathds{E}\Vert\nabla_w Q_k(\boldsymbol{w}_{k,n-1};\widehat{\boldsymbol{x}}_{k,n},\boldsymbol{y}_{k,n}) \notag\\
     &\quad\;- \nabla_w Q_k(w^{\star};\boldsymbol{x}_{k,n}+\boldsymbol{\delta}_k^{\star}(w^{\star}), \boldsymbol{y}_{k,n})\Vert^2\notag\\
     &\quad\; + 2\mathds{E}\Vert\nabla_w Q_k(w^{\star};\boldsymbol{x}_{k,n}+\boldsymbol{\delta}_k^{\star}(w^{\star}), \boldsymbol{y}_{k,n})\Vert^2\bigg)\notag\\
    &\overset{(c)}{\le}4L^2\sum_k\pi_k \Vert\widetilde{\boldsymbol{w}}_{k,n-1}\Vert^2 + O(\epsilon^2) \notag\\
    &\quad\;+2\sum_k\pi_k\mathds{E}\Vert\nabla_w Q_k(w^{\star};\boldsymbol{x}_{k,n}+\boldsymbol{\delta}_k^{\star}(w^{\star}),\boldsymbol{y}_{k,n})\Vert^2
\end{align}
where $(a)$ and $(b)$ follow from Jensen's inequality, and $(c)$ follows from (\ref{aaf_a}). With \eqref{AC_36_2}, we then have 
\begin{align}\label{AC_36o}
     &\mu^2 \left\Vert \sum_k \pi_k \mathds{E}\nabla_w Q_k(\boldsymbol{w}_{k,n-1};\widehat{\boldsymbol{x}}_{k,n},\boldsymbol{y}_{k,n})\right\Vert^2  \notag\\
     &\le 4\mu^2L^2\sum_k\pi_k \Vert\widetilde{\boldsymbol{w}}_{k,n-1}\Vert^2 + O(\mu^2)
\end{align}

Next, since $w^\star$ is the global minimizer, we have
\begin{align}\label{AC_35}
    0 \in \sum_k \pi_k \partial_{w} J_k(w^\star)
\end{align}
which implies that there exists 
\begin{align}\label{AC_35_1}
 g_k(w^{\star}) \in \partial_{w} J_k(w^\star)   
\end{align}
 such that 
\begin{align}\label{AC_35_a}
   \sum_k \pi_k g_k(w^{\star}) = 0
\end{align}
Moreover, recalling (\ref{danskin_ch}) gives
\begin{align}
    \partial_w J_k(w^{\star}) &= \mathds{E}\partial_w f_k(w;\boldsymbol{x}_k,\boldsymbol{y}_k) \notag\\
    &= \mathds{E}\bigg\{ \mathrm{conv}\Big\{ \nabla_w Q_k(w;\boldsymbol{x}_k+\boldsymbol{\delta}_k^\star,\boldsymbol{y}_k)\notag\\
    &\quad\quad\;\Big\vert \boldsymbol{\delta}_k^\star \in \mathop{\mathrm{argmax}}\limits_{\left\Vert\delta\right\Vert_{p_k}\le \epsilon_k} Q_k(w;\boldsymbol{x}_k+\delta,\boldsymbol{y}_k)\Big\}\bigg\}
\end{align}
which means the the set of the subgradients of $f_k(w;\boldsymbol{x}_k, \boldsymbol{y}_k)$ is the convex hull of the gradients of $Q_k(\cdot)$ evaluated at all possible maximizers. In turn, the set of the subgradients of $J_k(w)$ is the expectation of these subgradients over all samples. Thus, for each sample $\{\boldsymbol{x}_k,\boldsymbol{y}_k\}$, assume there are $I_{\boldsymbol{x}_k}$ maximizers, then $\exists a_{i\boldsymbol{x}_k}  \ge 0$ and
\begin{align}
\sum_{i=1}^{I_{\boldsymbol{x}_k}}a_{i\boldsymbol{x}_k}  = 1   
\end{align}
such that
\begin{align}
    g_k(w^{\star}) =& \mathds{E}\bigg\{ \sum_{i=1}^{I_{\boldsymbol{x}_k}} a_{i\boldsymbol{x}_k} \nabla_w Q_k(w^{\star};\boldsymbol{x}_k+\boldsymbol{\delta}_{k,i}^\star,\boldsymbol{y}_k)\notag\\
    &\quad\big\vert \boldsymbol{\delta}_{k,i}^\star \in \mathop{\mathrm{argmax}}\limits_{\left\Vert\delta\right\Vert_{p_k}\le \epsilon_k} Q_k(w^{\star};\boldsymbol{x}_k+\delta,\boldsymbol{y}_k)\bigg\}
\end{align}
where we use the subscript $\boldsymbol{x}_k$ in $I_{\boldsymbol{x}_k}$ and $a_{i\boldsymbol{x}_k}$ to emphasize that they may change with samples. Then for the inner product term of  (\ref{AC_32}), we substitute (\ref{AC_35_a}) into it, and obtain
\begin{align}\label{AC_37}
     & \sum_k\pi_k\mathds{E}\nabla_{w^{\sf T}} Q_k(\boldsymbol{w}_{k,n-1};\widehat{\boldsymbol{x}}_{k,n},\boldsymbol{y}_{k,n})\bar{\boldsymbol{w}}_{n-1} \notag\\
     &= \sum_k\pi_k\left(\mathds{E}\nabla_w Q_k(\boldsymbol{w}_{k,n-1};\widehat{\boldsymbol{x}}_{k,n},\boldsymbol{y}_{k,n}) -  g_k(w^{\star})\right)^{\sf T}\bar{\boldsymbol{w}}_{n-1}\notag\\
    &= \sum_k\pi_k\Big(\mathds{E}\nabla_w Q_k(\boldsymbol{w}_{k,n-1};\widehat{\boldsymbol{x}}_{k,n},\boldsymbol{y}_{k,n}) \notag\\
    &\quad\;- \mathds{E}\nabla_w Q_k(\boldsymbol{w}_{k,n-1};\boldsymbol{x}_{k,n}^{\star},\boldsymbol{y}_{k,n}) \notag\\
    &\quad\;+ \mathds{E}\nabla_w Q_k(\boldsymbol{w}_{k,n-1};\boldsymbol{x}_{k,n}^{\star},\boldsymbol{y}_{k,n}) - g_k(w^{\star})\Big)^{\sf T}\bar{\boldsymbol{w}}_{n-1}
\end{align}
where 
\begin{align}
\label{AC_37_1}
    &\boldsymbol{x}_{k,n}^{\star} = \boldsymbol{x}_{k,n} + \boldsymbol{\delta}_{k,n}^{\star}\\
    \label{AC_37_2}
    &\boldsymbol{\delta}_{k,n}^{\star} \in \mathop{\mathrm{argmax}}\limits_{\left\Vert\delta\right\Vert_{p_k}\le \epsilon_k} Q_k(\boldsymbol{w}_{k,n-1};\boldsymbol{x}_{k,n}+\delta,\boldsymbol{y}_{k,n})\\
    \label{AC_37_3}
    &\mathds{E}\nabla_w Q_k(\boldsymbol{w}_{k,n-1};\boldsymbol{x}_{k,n}^{\star},\boldsymbol{y}_{k,n}) \in \partial_w J_k(\boldsymbol{w}_{k,n-1})
\end{align}
Thus, the first error term in (\ref{AC_37}) is related to the approximation error of the maximization, while the second error term measures the difference of subgradients of my $J_k(w)$ at $\boldsymbol{w}_{k,n-1}$ and $w^{\star}$. Basically, tor the first error term, we have
\begin{align}\label{approx_max_e}
    & \sum_k\pi_k \big(\mathds{E}\nabla_w Q_k(\boldsymbol{w}_{k,n-1};\widehat{\boldsymbol{x}}_{k,n},\boldsymbol{y}_{k,n}) \notag\\
    &\quad\;- \mathds{E}\nabla_w Q_k(\boldsymbol{w}_{k,n-1};\boldsymbol{x}_{k,n}^{\star},\boldsymbol{y}_{k,n})\big)^{\sf T}\bar{\boldsymbol{w}}_{n-1} \notag\\
    &\le \sum_k\pi_k \mathds{E}\big\Vert\nabla_w Q_k(\boldsymbol{w}_{k,n-1};\widehat{\boldsymbol{x}}_{k,n},\boldsymbol{y}_{k,n}) \notag\\
    &\quad\;- \nabla_w Q_k(\boldsymbol{w}_{k,n-1};\boldsymbol{x}_{k,n}^{\star},\boldsymbol{y}_{k,n})\big\Vert\Vert\bar{\boldsymbol{w}}_{n-1}\Vert\notag\\
    &\overset{(a)}{\le }\sum_k\pi_k(L\mathds{E}\Vert\widehat{\boldsymbol{\delta}}_{k,n}-\boldsymbol{\delta}_{k,n}^{\star} \Vert\Vert\bar{\boldsymbol{w}}_{n-1}\Vert) \notag\\
    &\overset{(b)}{\le} L\times O(\epsilon)\Vert\bar{\boldsymbol{w}}_{n-1}\Vert\notag\\
    &\overset{(c)}{\le} O(\epsilon) + O(\epsilon)\Vert\bar{\boldsymbol{w}}_{n-1}\Vert^2
\end{align}
where $(a)$ follows from (\ref{as2_4}), and $(b)$ follows from \eqref{e_p_1}. Note that in $(b)$, we use $O(\epsilon)$ to denote the approximation error of the maximization problem over $\delta$ since $\epsilon$ is defined to be small in the context of adversarial training. When the exact solutions of the maximization problem are obtainable, (\ref{approx_max_e}) becomes $0$. With regard to $(c)$, we use the following inequality:
\begin{align}\label{AC_40}
    \sqrt{x} \le \frac{1}{2}(x+1)
\end{align}
for any x.

As for the second error term in (\ref{AC_37}), we have
\begin{align}\label{AC_38_o}
&\sum_k\pi_k \left(\mathds{E}\nabla_w Q_k(\boldsymbol{w}_{k,n-1};\boldsymbol{x}_{k,n}^{\star},\boldsymbol{y}_{k,n}) - g_k(w^{\star})\right)^{\sf T}\bar{\boldsymbol{w}}_{n-1}   \notag\\
&=  \sum_k\pi_k \big(\mathds{E}\nabla_w Q_k(\boldsymbol{w}_{k,n-1};\boldsymbol{x}_{k,n}^{\star},\boldsymbol{y}_{k,n}) - g_k(w^{\star})\big)^{\sf T}\notag\\
&\quad\;\times(\bar{\boldsymbol{w}}_{n-1} - \boldsymbol{w}_{k,n-1} + \boldsymbol{w}_{k,n-1})
\end{align}
for which since $J_k(\boldsymbol{w})$ is $\nu$-strongly convex, 
and according to (\ref{AC_35_1}) and (\ref{AC_37_3}), we get
\begin{align}\label{AC_38}
    &\left(\mathds{E}\nabla_w Q_k(\boldsymbol{w}_{k,n-1};\boldsymbol{x}_{k,n}^{\star},\boldsymbol{y}_{k,n}) - g_k(w^{\star})\right)^{\sf T}\widetilde{\boldsymbol{w}}_{k,n-1} \notag\\
    &\le -\nu\Vert\widetilde{\boldsymbol{w}}_{k,n-1}\Vert^2
\end{align}
and 
\begin{align}\label{AC_39}
    & \sum_k \pi_k\left(\mathds{E}\nabla_w Q_k(\boldsymbol{w}_{k,n-1};\boldsymbol{x}_{k,n}^{\star},\boldsymbol{y}_{k,n}) - g_k(w^{\star})\right)^{\sf T}(\bar{\boldsymbol{w}}_{n-1} 
     \notag\\
     &\quad\;- \widetilde{\boldsymbol{w}}_{k,n-1}) \notag\\
    & \overset{(a)}{\le} \sum_k\pi_k\Vert\mathds{E}\nabla_w Q_k(\boldsymbol{w}_{k,n-1};\boldsymbol{x}_{k,n}^{\star},\boldsymbol{y}_{k,n}) - g_k(w^{\star})\Vert\notag\\
    &\quad\;\times\Vert\bar{\boldsymbol{w}}_{n-1} 
     - \widetilde{\boldsymbol{w}}_{k,n-1}\Vert\notag\\
\end{align}    
 where $(a)$ follows from the Cauchy-Schwarz inequality. Then, taking expectations on (\ref{AC_39}) gives
 \begin{align}\label{AC_39_1}
 & \mathds{E}\bigg\{\sum_k \pi_k\left(\mathds{E}\nabla_w Q_k(\boldsymbol{w}_{k,n-1};\boldsymbol{x}_{k,n}^{\star},\boldsymbol{y}_{k,n}) - g_k(w^{\star})\right)^{\sf T}\notag\\
 &\quad\;\times(\bar{\boldsymbol{w}}_{n-1} 
     - \widetilde{\boldsymbol{w}}_{k,n-1})\bigg\}\notag\\
     &\overset{(a)}{\le} \sum_k\pi_k \sqrt{\mathds{E}\Vert\mathds{E}\nabla_w Q_k(\boldsymbol{w}_{k,n-1};\boldsymbol{x}_{k,n}^{\star},\boldsymbol{y}_{k,n}) - g_k(w^{\star})\Vert^2}\notag\\
&\quad\;\times\sqrt{\mathds{E}\Vert\bar{\boldsymbol{w}}_{n-1}
     - \widetilde{\boldsymbol{w}}_{k,n-1}\Vert^2}\notag\\
     &\overset{(b)}{\le}
     \sum_k\pi_k 
    \sqrt{\begin{aligned}&\mathds{E}\Vert\nabla_w Q_k(\boldsymbol{w}_{k,n-1};\boldsymbol{x}_{k,n}^{\star},\boldsymbol{y}_{k,n}) \notag\\
     &- \sum_{i=1}^{I_{\boldsymbol{x}_k}} a_{i\boldsymbol{x}_k} \nabla_w Q_k(w^{\star};\boldsymbol{x}_k+\boldsymbol{\delta}_{k,i}^\star,\boldsymbol{y}_k)\Vert^2\end{aligned}}\notag\\
&\quad\;\times\sqrt{\mathds{E}\Vert\bar{\boldsymbol{w}}_{n-1}
     - \widetilde{\boldsymbol{w}}_{k,n-1}\Vert^2}\notag\\
     &\overset{(c)}{\le}\sum_k\pi_k h\mu\sqrt{2L^2\mathds{E}\Vert\widetilde{\boldsymbol{w}}_{k,n-1}\Vert^2+O(\epsilon^2)}\notag\\
     & \overset{(d)}{\le}  h \mu L^2 \sum_k\pi_k\mathds{E}\Vert\widetilde{\boldsymbol{w}}_{k,n-1}\Vert^2 +  h\mu\left(O(\epsilon^2)+\frac{1}{2}\right)
\end{align}
where $(a)$ follows from the Cauchy-Schwarz inequality, $(b)$ follows from Jensen's inequality, $(c)$ follows from (\ref{affine_l_o}) and (\ref{c_w_c}), and $h \overset{\Delta}{=} \Vert \mathcal{V}_{L} \Vert \ddot{c}_{\epsilon}$. That is, from (\ref{c_w_c}), we know after enough iterations:
\begin{align}
\sqrt{\mathds{E}\Vert\bar{\boldsymbol{w}}_{n-1}
     - \widetilde{\boldsymbol{w}}_{k,n-1}\Vert^2} \le&  \sqrt{\mathds{E}\Vert  {\boldsymbol{\scriptstyle\mathcal{W}}}_{n} - {\boldsymbol{\scriptstyle\mathcal{W}}}_{c,n} \Vert^2}  &\notag\\
     \le&\Vert \mathcal{V}_{L} \Vert \mu\ddot{c}_{\epsilon} = h\mu
\end{align}

{Substituting (\ref{approx_max_e}), (\ref{AC_38}), and (\ref{AC_39_1}) into (\ref{AC_37}), and taking expectations of it, we have
\begin{align}\label{AC_41}
    &\mathds{E}\left\{\sum_k\pi_k\mathds{E}\nabla_{w^{\sf T}} Q_k(\boldsymbol{w}_{k,n-1};\widehat{\boldsymbol{x}}_{k,n},\boldsymbol{y}_{k,n})\bar{\boldsymbol{w}}_{n-1} \right\} \notag\\
    &\le - (\nu - h \mu L^2) \sum_k\pi_k \mathds{E}\Vert \widetilde{\boldsymbol{w}}_{k,n-1}\Vert^2  + O(\epsilon)\mathds{E}\Vert\bar{\boldsymbol{w}}_{n-1}\Vert^2 \notag\\
    &\quad\;+ h\mu\left(O(\epsilon^2)+\frac{1}{2}\right) + O(\epsilon)
\end{align}}

Now substituting (\ref{AC_33}), (\ref{AC_36o}) and (\ref{AC_41}) into (\ref{AC_32}), and taking expectations on both sides of it gives
\begin{align}\label{AC_42}
\mathds{E}\Vert\bar{\boldsymbol{w}}_{n}\Vert^2 
   & \le  \mathds{E}\Vert\bar{\boldsymbol{w}}_{n-1}\Vert^2  -\big(2\mu \nu - 2 \mu^2h  L^2- {4\mu^2 L^2} \notag\\
   &\quad\;- \mu^2\beta_{\max}^2 \big)\sum_k\pi_k\mathds{E}\Vert\widetilde{\boldsymbol{w}}_{k,n-1}\Vert^2 \notag\\
   &\quad\;+{O(\mu\epsilon)\mathds{E}\Vert\bar{\boldsymbol{w}}_{n-1}\Vert^2} + O(\mu^2) + {O(\mu\epsilon)}\notag\\
   & \overset{(a)}{\le}   \mathds{E}\Vert\bar{\boldsymbol{w}}_{n-1}\Vert^2 -\big(2\mu \nu - 2 \mu^2h  L^2- {4\mu^2 L^2} \notag\\
   &\quad\;- \mu^2\beta_{\max}^2\big)\mathds{E}\Vert\sum_k\pi_k \widetilde{\boldsymbol{w}}_{k,n-1}\Vert^2 \notag\\
   &\quad\;+ {O(\mu\epsilon)\mathds{E}\Vert\bar{\boldsymbol{w}}_{n-1}\Vert^2} + O(\mu^2) + {O(\mu\epsilon)}\notag\\
   & =  \mathds{E}\Vert\bar{\boldsymbol{w}}_{n-1}\Vert^2 -
    (2\mu \nu - 2 \mu^2 h  L^2- 4\mu^2 L^2\notag\\
    &\quad\;- \mu^2\beta_{\max}^2 )\mathds{E}\Vert \bar{\boldsymbol{w}}_{n-1}\Vert^2 \notag\\
    &\quad\;+ {O(\mu\epsilon)\mathds{E}\Vert\bar{\boldsymbol{w}}_{n-1}}\Vert^2 + O(\mu^2) + {O(\mu\epsilon)}\notag\\
   &=  \lambda \mathds{E}\Vert \bar{\boldsymbol{w}}_{n-1}\Vert^2 + \bar{c}_{\epsilon}^2\mu^2 {+ O(\mu\epsilon)}
\end{align}
where $(a)$ follows from Jensen's inequality, and 
\begin{align}
    0< \lambda = 1 - 2\mu \nu {+ O(\mu\epsilon)} + O(\mu^2) <1
\end{align}
when $\mu$ {and $\epsilon$ are} sufficiently small. Also,
\begin{align}
    \bar{c}_{\epsilon}^2 =& 2h\left(O(\epsilon^2)+\frac{1}{2}\right) +O(\epsilon^2) \notag\\
    &+  \sigma^2 {+ 2\sum_k\pi_k\mathds{E}\Vert\nabla_w Q_k(w^{\star};\boldsymbol{x}_{k,n}+\delta_k^{\star}(w^{\star}),\boldsymbol{y}_{k,n})\Vert^2}
\end{align}
Iterating (\ref{AC_42}), and after enough iterations $n \ge \bar{n}$ with 
\begin{align}\label{barn}
    \bar{n} =  O\left(\frac{\log \mu}{\log (1 - O(\mu))}\right)
\end{align}
we have 
\begin{align}\label{AC_45}
 \mathds{E}\left\Vert \bar{\boldsymbol{w}}_{n}\right\Vert^2 \le 2\mu^2 \frac{\bar{c}_{\epsilon}^2}{1-\lambda} {+ O(\epsilon)} = O(\mu){+ O(\epsilon)}
\end{align}
where the $O(\epsilon)$ term corresponds to the approximation error in solving the maximization problem in (\ref{f_g}). If the problems is exactly solved, this error term becomes $0$.

Combining (\ref{checkn}), (\ref{AC_31}), (\ref{AC_30_1}), (\ref{barn}), and (\ref{AC_45}), we conclude that after enough iterations $n$ with 
\begin{align}\label{step_n}
    n \ge& \ddot{n} + \bar{n} \notag\\
    =& O\left(\frac{\log \mu}{\log (t+O(\mu^2))}\right)+O\left(\frac{\log \mu}{\log (1 - O(\mu))}\right)\notag\\
    \overset{(a)}{=}& O\left(\frac{\log \mu}{\log (1 - O(\mu))}\right)
\end{align}
where $(a)$ follows from the following equality:
\begin{align}
    \lim\limits_{\mu \to 0}\frac{\frac{\log \mu}{\log (t+O(\mu^2))}}{\frac{\log \mu}{\log (1 - O(\mu))}} = 0
\end{align}
we have
\begin{align}
    \mathds{E}\left\Vert {\widetilde{\boldsymbol{\scriptstyle\mathcal{W}}}}_{n}\right\Vert^2  = &  \mathds{E}\Vert\mathcal{V}^{\sf -T}\mathcal{V}^{\sf T}{\widetilde{\boldsymbol{\scriptstyle\mathcal{W}}}}_{n}\Vert^2\notag\\
    \overset{(a)}{\le}&  \Vert\mathcal{V}^{\sf -T}\Vert^2 \left\{\mathds{E}\left\Vert \bar{\boldsymbol{w}}_{n}\right\Vert^2+ \mathds{E}\left\Vert \check{\boldsymbol{w}}_{n}\right\Vert^2 \right\}\notag\\
    \le& c_\epsilon^2\mu {+ O(\epsilon)} \notag\\
    =& O(\mu) {+ O(\epsilon)}
\end{align}
where $(a)$ follows from (\ref{AC_13}), and
\begin{align}\label{c_e}
   c_\epsilon^2 = \Vert\mathcal{V}^{\sf -T}\Vert^2\left(\frac{2\bar{c}_{\epsilon}^2}{1-\lambda} + \mu\ddot{c}_{\epsilon}^2\right) 
\end{align}
Note from (\ref{step_n}) that the convergence speed of $\mathds{E}\Vert\boldsymbol{\bar{w}}_n\Vert^2$ dominates the convergence of $\mathds{E}\Vert {\widetilde{\boldsymbol{\scriptstyle\mathcal{W}}}}_{n}\Vert^2$, thus $\mathds{E}\Vert {\widetilde{\boldsymbol{\scriptstyle\mathcal{W}}}}_{n}\Vert^2$ converges linearly with rate $\lambda$.
\end{proof}

\section{proof of Corollary \ref{co_mse}}\label{ap_co_mse}
\begin{proof}
The proof for Corollary \ref{co_mse} is very similar to the proof for Theorem \ref{th_mse} in Appendix \ref{ap_th_mse}. Since (\ref{check_dd_w}) holds no matter the optimal perturbation or an approximate one is used in the algorithms, for $n\ge\ddot{n}$, we still have
\begin{align}\label{AC_30_f_1}
 \mathds{E}\left\Vert \check{\boldsymbol{w}}_{n}\right\Vert^2 = \mathds{E}\left\Vert \ddot{\boldsymbol{w}}_{n}\right\Vert^2 \le O(\mu^2)
\end{align}

Then, for (\ref{w_bar_exact}), by squaring and taking expectations of (\ref{w_bar_2}) conditioned on $\boldsymbol{\mathcal{F}}_{n-1}$, 
we have
\begin{align}\label{AC_f_32}
    &\mathds{E}\left\{\Vert\bar{\boldsymbol{w}}_{n}\Vert^2\vert\boldsymbol{\mathcal{F}}_{n-1}\right\} \notag\\
    &=   \Vert\bar{\boldsymbol{w}}_{n-1}\Vert^2 + 2\mu\sum_k\pi_k\nabla_{w^{\sf T}} J_k(\boldsymbol{w}_{k,n-1})\bar{\boldsymbol{w}}_{n-1} \notag\\
    &\quad\;+ \mu^2 \left\Vert \sum_k \pi_k \nabla_{w} J_k(\boldsymbol{w}_{k,n-1})\right\Vert^2 \notag\\
    &\quad\;+ \mu^2\mathds{E}\left\{\left\Vert\sum_k \pi_k \boldsymbol{s}_{k,n}^B\right\Vert^2\vert\boldsymbol{\mathcal{F}}_{n-1}\right\}
\end{align}
The remaining derivations to prove the convergence of (\ref{AC_f_32}) are very similar to (\ref{AC_33})--(\ref{c_e}), except at the following details:

First, since now the optimal perturbation is used in the training algorithms, the approximation error term in (\ref{AC_37}) is removed.

Second, since now we have
\begin{align}\label{AC_f_35}
    \sum_k \pi_k \nabla_{w} J_k(w^\star)= 0
\end{align}
the proof in (\ref{AC_36_2}) can be simplified:
\begin{align}\label{AC_36}
     &\left\Vert \sum_k \pi_k \nabla_{w} J_k(\boldsymbol{w}_{k,n-1})\right\Vert^2  \notag\\
     &= \left\Vert\sum_k \pi_k \nabla_{w} J_k(\boldsymbol{w}_{k,n-1}) - \sum_k \pi_k \nabla_{w} J_k(w^\star) \right\Vert^2 \notag\\
     &\overset{(a)}{\le}2L^2\sum_k\pi_k\Vert\widetilde{\boldsymbol{w}}_{k,n-1}\Vert^2 + O(\epsilon^2)
\end{align}
where $(a)$ follows from (\ref{affine_l_1}). 

Then following a similar derivation to Appendix (\ref{ap_th_mse}), for $n \ge \bar{n}$, we obtain
\begin{align}\label{AC_f_45}
     \mathds{E}\left\Vert \bar{\boldsymbol{w}}_{n}\right\Vert^2 \le 2\mu^2\frac{\bar{c}_{\epsilon}^2}{1-\lambda} + O(\epsilon) = O(\mu)
\end{align}
and, asymptotically,
\begin{align}
    \mathop{\lim \sup}\limits_{n\to\infty} \mathds{E}\left\Vert {\widetilde{\boldsymbol{\scriptstyle\mathcal{W}}}}_{n}\right\Vert^2  \le O(\lambda^n) + O(\mu) 
\end{align}
\end{proof}

\section{Proof of Lemma \ref{lip_convex}}\label{ap_lip_convex}
\begin{proof}
{The idea for the proof of Lemma \ref{lip_convex} is similar to the proof with the traditional Lipschitz condition \cite{sayed_2023,garrigo}, but the constant in the affine Lipschitz condition (\ref{affine_l_1}) introduces some change.  We first prove (\ref{u_affine}). Consider the function
\begin{align}
    h(t) = f_k((1-t)w_1 + tw_2;\cdot)
\end{align} 
for any $0\le t\le1$, with which we have
\begin{align}
    h(0) = f_k(w_1;\cdot), \quad h(1) = f_k(w_2;\cdot)
\end{align}
Then, differentiating $h(t)$ with respect to $t$ gives
\begin{align}
    h'(t) = \nabla_{w^{\sf T}}f_k((1-t)w_1 + tw_2;\cdot)(w_2-w_1)
\end{align}
with which we have
\begin{align}
    &h'(t) - h'(0) \notag\\
    &= \left(\nabla_{w^{\sf T}}f_k((1-t)w_1 + tw_2;\cdot) - \nabla_{w^{\sf T}}f_k(w_1;\cdot)\right)(w_2-w_1)\notag\\
    &\le 
    \Vert\nabla_{w^{\sf T}}f_k((1-t)w_1 + tw_2;\cdot) - \nabla_{w^{\sf T}}f_k(w_1;\cdot)\Vert\Vert w_2 - w_1\Vert\notag\\
    &\overset{(a)}{\le} tL\Vert w_2 - w_1\Vert^2 + O(\epsilon)\Vert w_2 - w_1\Vert
\end{align}
where $(a)$ follows from(\ref{affine_l_1}).
It then follows from the fundamental theorem of Calculus that
\begin{align}\label{u_affine_ap}
    &f_k(w_2;\cdot) - f_k(w_1;\cdot) \notag\\
    &=  h(1) - h(0) \notag\\
    &= \int_0^1 h'(t)dt \notag\\
    &\le\int_0^1 (h'(0) + tL\Vert w_2 - w_1\Vert^2 + O(\epsilon)\Vert w_2 - w_1\Vert)dt \notag\\
    &\le \nabla_{w^{\sf T}} f_k(w_1;\cdot) + \frac{L}{2}\Vert w_2 - w_1\Vert^2 + O(\epsilon)\Vert w_2 - w_1\Vert
\end{align}}

We now prove (\ref{d_g_f_k}). For any $w\in \mathbb{R}^M$, according to the convexity of $f_k$ and (\ref{u_affine_ap}), we have
\begin{align}\label{G_1}
 &f_k(w_2;\cdot) - f_k(w_1;\cdot) \notag\\
 &=    f_k(w_2;\cdot) - f_k(w;\cdot)  + f_k(w;\cdot)-f_k(w_1; \cdot)\notag\\
 &\overset{(a)}{\le}\nabla_{w^{\sf T}} f_k(w_2;\cdot)(w_2 - w) + f_k(w;\cdot) - f_k(w_1;\cdot)\notag\\
 &\overset{(b)}{\le}\nabla_{w^{\sf T}} f_k(w_2;\cdot)(w_2 - w) + \nabla_{w^{\sf T}} f_k(w_1;\cdot)(w - w_1) \notag\\
 &\quad\;+\frac{L+1}{2}\Vert w-w_1\Vert^2 + O(\epsilon^2)
\end{align}
where $(a)$ follows from the convexity of $f_k$, and $(b)$ follows from (\ref{u_affine_ap}) and (\ref{AC_40}). Then, we minimize the right-hand side of (\ref{G_1}) with respect to $w$ by letting its gradient be 0, and obtain the minimizer
\begin{align}\label{G_2}
    w = \frac{1}{L+1}\left(\nabla_w f_k(w_2;\cdot) - \nabla_w f_k(w_1;\cdot)\right) + w_1
\end{align}
Substituting (\ref{G_2}) into the right-hand side of (\ref{G_1}) gives its supremum, and hence,
\begin{align}
&f_k(w_2;\cdot) - f_k(w_1;\cdot) \notag\\
&\le \nabla_{w^{\sf T}} f_k(w_2;\cdot)(w_2-w_1) \notag\\
&\quad\;- \frac{1}{2(L+1)}\Vert \nabla_w f_k(w_2;\cdot) - \nabla_w f_k(w_1;\cdot) \Vert^2 + O(\epsilon^2)
\end{align}
with which (\ref{d_g_f_k}) can be obtained.
\end{proof}

\section{Proof of Lemma \ref{b_gconvex}}\label{ap_b_gconvex}
\begin{proof}
{
\begin{align}
  &\mathds{E}\left\{\Vert\nabla_w f_k(\boldsymbol{w};\boldsymbol{x}_k,\boldsymbol{y}_k)\Vert^2 \vert \mathcal{F}_{n-1}\right\} \notag\\
  &=   \mathds{E}\Vert\nabla_w f_k(\boldsymbol{w};\boldsymbol{x}_k,\boldsymbol{y}_k) - \nabla_w f_k(w^{\star};\boldsymbol{x}_k,\boldsymbol{y}_k)  \notag\\
  &\quad\;+\nabla_w f_k(w^{\star};\boldsymbol{x}_k,\boldsymbol{y}_k)\Vert^2\notag\\
  &\overset{(a)}{\le} 2\mathds{E}\Vert\nabla_w f_k(\boldsymbol{w};\boldsymbol{x}_k,\boldsymbol{y}_k) - \nabla_w f_k(w^{\star};\boldsymbol{x}_k,\boldsymbol{y}_k)\Vert^2 \notag\\
  &\quad\;+ 2\mathds{E}\Vert\nabla_w f_k(w^{\star};\boldsymbol{x}_k,\boldsymbol{y}_k)\Vert^2\notag\\
  &\overset{(b)}{\le} 4(L+1)\left( J_k(\boldsymbol{w}) - J_k(w^{\star}) - \nabla_{w^{\sf T}} J_k(w^\star)(\boldsymbol{w} - w^\star)\right) \notag\\
  &\quad\;+ O(\epsilon^2) + 2\mathds{E}\Vert\nabla_w f_k(w^{\star};\boldsymbol{x}_k,\boldsymbol{y}_k)\Vert^2
\end{align}
where (a) follows from Jensen's inequality, and (b) follows from (\ref{d_g_f_k}).}
\end{proof}

\section{Proof of Lemma \ref{sg_convex}}\label{ap_sg_convex}
\begin{proof}
The proof of the zero mean is the same to the strongly-convex case in section \ref{ap2}. Thus, we only need to prove the bound for the variance of the gradient noise. Note that when $B=1$, we have
    \begin{align}
     & \mathds{E}\left\{\left\Vert \boldsymbol{s}_{k,n}\right\Vert^2|\boldsymbol{\mathcal{F}}_{n-1}\right\} \notag\\
     &= \mathds{E}\Vert\nabla_w f(\boldsymbol{w}_{k,n-1};\boldsymbol{x}_k,\boldsymbol{y}_k) - \nabla_w J_k(\boldsymbol{w}_{k,n-1})\Vert^2\notag\\
&\overset{(a)}{\le} 4\mathds{E}\Vert\nabla_w f_k(\boldsymbol{w}_{k,n-1};\boldsymbol{x}_k,\boldsymbol{y}_k)\Vert^2\notag\\ 
&\overset{(b)}{\le} 8\mathds{E}\Vert\nabla_w f_k(\boldsymbol{w}_{k,n-1};\boldsymbol{x}_k,\boldsymbol{y}_k) - \nabla_w f_k(\boldsymbol{w}_{c,n-1};\boldsymbol{x}_k,\boldsymbol{y}_k)\Vert^2 \notag\\
&\quad\;+ 8\mathds{E}\Vert \nabla_w f_k(\boldsymbol{w}_{c,n-1};\boldsymbol{x}_k,\boldsymbol{y}_k) \Vert^2\notag\\
&\overset{(c)}{\le}32(L+1)\Big( J_k(\boldsymbol{w}_{c,n-1}) - J_k(w^{\star}) \notag\\
&\quad\;- \nabla_{w^{\sf T}} J_k(w^\star)(\boldsymbol{w}_{c,n-1} - w^\star)\Big)\notag\\ &\quad\;+ 8\gamma_k^2 + O(\epsilon^2) + O(\Vert\boldsymbol{w}_{k,n-1} - \boldsymbol{w}_{c,n-1}\Vert^2)
    \end{align}
 where $(a)$ and $(b)$ follow from Jensen's inequality, and $(c)$ follows from (\ref{affine_l_1}) and (\ref{d_g_f_k}). Then, similar to (\ref{sb2}), we have
 \begin{align}
  &\mathds{E}\left\{\left\Vert \boldsymbol{s}_{k,n}^{B}\right\Vert^2|\boldsymbol{\mathcal{F}}_{n-1}\right\}  \notag\\
 & =  \frac{1}{B} \mathds{E}\left\{\left\Vert \boldsymbol{s}_{k,n}\right\Vert^2|\boldsymbol{\mathcal{F}}_{n-1}\right\} \notag\\
  &= \frac{32(L+1)}{B}\Big( J_k(\boldsymbol{w}_{c,n-1}) - J_k(w^{\star}) \notag\\
  &\quad\;- \nabla_{w^{\sf T}} J_k(w^\star)(\boldsymbol{w}_{c,n-1} - w^\star)\Big) + \frac{8\gamma_k^2}{B} + O(\epsilon^2) \notag\\
  &\quad\;+ O(\Vert\boldsymbol{w}_{k,n-1} - \boldsymbol{w}_{c,n-1}\Vert^2)
 \end{align}
\end{proof}

\section{Proof of Theorem~\ref{th_j_convex}}\label{ap_th_j_convex}
\begin{proof}
Recalling (\ref{AC_14_2_opt}) gives 
    \begin{align}\label{J_s}
        &\mathds{E}\Vert \boldsymbol{w}_{c,n} - w^{\star}\Vert^2 - \mathds{E}\Vert \boldsymbol{w}_{c,n-1} - w^{\star}\Vert^2 \notag\\
        &= \mathds{E}\Vert\boldsymbol{w}_{c,n-1} -\sum_k\pi_k \nabla_w J_k(\boldsymbol{w}_{k,n-1}) - \sum_k \pi_k \boldsymbol{s}_{k,n}^{B}- w^{\star}\Vert^2 \notag\\
        &\quad\; - \mathds{E}\Vert \boldsymbol{w}_{c,n-1} - w^{\star}\Vert^2\notag\\
        &\overset{(a)}{=} 2\mu\mathds{E}\sum_k\pi_k \nabla_{w^{\sf T}} J_k(\boldsymbol{w}_{k,n-1})(w^{\star}- \boldsymbol{w}_{c,n-1}) \notag\\
        &\quad\;+ \mu^2\mathds{E}\Vert\sum_k\pi_k \nabla_w J_k(\boldsymbol{w}_{k,n-1})\Vert^2+ \mu^2\mathds{E}\Vert\sum_k \pi_k \boldsymbol{s}_{k,n}^{B}\Vert^2     
    \end{align}
where (a) follows from (\ref{s0_convex_2}).

To proceed, we now examine the bounds of the terms in the right-hand side of (\ref{J_s}). Basically, we have 
\begin{align}\label{J_1}
&\mathds{E}\Vert\sum_k\pi_k \nabla_w J_k(\boldsymbol{w}_{k,n-1})\Vert^2\notag\\
&\overset{(a)}{\le} \sum_k\pi_k \mathds{E}\Vert \nabla_w f_k(\boldsymbol{w}_{k,n-1};\boldsymbol{x}_k, \boldsymbol{y}_k)\Vert^2  \notag\\
&\overset{(b)}{\le} \sum_k\pi_k 2\mathds{E}\big\Vert\nabla_w f_k(\boldsymbol{w}_{k,n-1};\boldsymbol{x}_k,\boldsymbol{y}_k) \notag\\
&\quad\;- \nabla_w f_k(\boldsymbol{w}_{c,n-1};\boldsymbol{x}_k,\boldsymbol{y}_k)\big\Vert^2 + 2\mathds{E}\Vert\nabla_w f_k(\boldsymbol{w}_{c,n-1};\boldsymbol{x}_k,\boldsymbol{y}_k)\Vert^2\notag\\
&\overset{(c)}{\le} \sum_k\pi_k \Big(O(\mu^2) + O(\epsilon^2) + 8(L+1)\big( \mathds{E}J_k(\boldsymbol{w}_{c,n-1}) - J_k(w^{\star}) \notag\\
&\quad\;- \nabla_{w^{\sf T}} J_k(w^\star)(\boldsymbol{w}_{c,n-1} - w^\star)\big) +  O(\epsilon^2) \notag\\
&\quad\;+ 4\mathds{E}\Vert\nabla_w f_k(w^{\star};\boldsymbol{x}_k,\boldsymbol{y}_k)\Vert^2\Big)\notag\\
&= 8(L+1)(\mathds{E}J(\boldsymbol{w}_{c,n-1}) - J(w^\star)) + O(\mu^2)+ O(\epsilon^2) \notag\\
&\quad\;+ 2\sum_k\pi_k\gamma_k^2
\end{align}
where $(a)$ and $(b)$ follow from Jensen's inequality, and $(c)$ follows from (\ref{affine_l_1}),  (\ref{th_wkc}), and (\ref{d_g_f_k}). Also, the term related to $(\boldsymbol{w}_{c,n-1} - w^{\star})$ is removed since $w^{\star}$ admits
\begin{align}
    \sum_k \pi_k \nabla_w J_k(w^{\star}) = 0
\end{align}
Similarly, for the gradient noise term in (\ref{J_s}), we have
\begin{align}\label{J_2}
 \mathds{E}\Vert\sum_k \pi_k \boldsymbol{s}_{k,n}^{B}\Vert^2 \le& \sum_k \pi_k\mathds{E} \Vert\boldsymbol{s}_{k,n}^{B}\Vert^2\notag\\
 \le & \frac{32(L+1)}{B}\left( \mathds{E}J(\boldsymbol{w}_{c,n-1}) - J(w^{\star}) \right) \notag\\
 &+ \frac{8}{B}\sum_k\pi_k\gamma_k^2 + O(\mu^2) + O(\epsilon^2)    
\end{align}

We then analyze the inner product term from (\ref{J_s}). Basically,
\begin{align}\label{J_3}
    &\sum_k\pi_k \mathds{E}\nabla_{w^{\sf T}} J_k(\boldsymbol{w}_{k,n-1})(w^{\star}- \boldsymbol{w}_{c,n-1}) \notag\\
    &= \sum_k\pi_k \mathds{E}\nabla_{w^{\sf T}} J_k(\boldsymbol{w}_{k,n-1})(w^{\star}- \boldsymbol{w}_{c,n-1} - \boldsymbol{w}_{k,n-1} \notag\\
    &\quad\;+ \boldsymbol{w}_{k,n-1})
\end{align}
for which following a similar technique to (\ref{AC_39_1}) gives
\begin{align}\label{J_4}
  &\sum_k\pi_k \mathds{E}\nabla_{w^{\sf T}} J_k(\boldsymbol{w}_{k,n-1})(\boldsymbol{w}_{k,n-1} - \boldsymbol{w}_{c,n-1}) \notag\\
  &\le \sum_k\pi_k \sqrt{\mathds{E}\Vert\nabla_{w} J_k(\boldsymbol{w}_{k,n-1}) \Vert^2\times \mathds{E}\Vert\boldsymbol{w}_{c,n-1} - \boldsymbol{w}_{k,n-1}\Vert^2}\notag\\
  &\overset{(a)}{\le} O(\mu)\sum_k \pi_k \left(\frac{\mathds{E}\Vert\nabla_{w} J_k(\boldsymbol{w}_{k,n-1}) \Vert^2}{2}+\frac{1}{2}\right)\notag\\
  &\overset{(b)}{\le} O(\mu)(\mathds{E}J(\boldsymbol{w}_{c,n-1}) - J(w^{\star})) + O(\mu)
\end{align}
where $(a)$ follows from (\ref{th_wkc}), and $(b)$ follows from a similar derivation to (\ref{J_1}). Then, since $J_k(w)$ is convex, it holds that
\begin{align}
\nabla_{w^{\sf T}}J_k(\boldsymbol{w}_{k,n-1})(w^{\star} - \boldsymbol{w}_{k,n-1}) \le J_k(w^{\star}) - J_k(\boldsymbol{w}_{k,n-1})   
\end{align}
with which we obtain
\begin{align}\label{J_5}
    &\sum_k\pi_k \mathds{E}\nabla_{w^{\sf T}}J_k(\boldsymbol{w}_{k,n-1})(w^{\star} - \boldsymbol{w}_{k,n-1})  \notag\\
    &\le J(w^{\star}) - \sum_k\pi_k J_k(\boldsymbol{w}_{k,n-1}) \notag\\
    &= J(w^{\star}) - \mathds{E}J(\boldsymbol{w}_{c,n-1}) +  \mathds{E}J(\boldsymbol{w}_{c,n-1}) \notag\\
    &\quad\;- \sum_k\pi_k \mathds{E}J_k(\boldsymbol{w}_{k,n-1})\notag\\
    &\overset{(a)}{\le} J(w^{\star}) - \mathds{E}J(\boldsymbol{w}_{c,n-1}) \notag\\
    &\quad\;+ \sum_k\pi_k \mathds{E}\nabla_{w^{\sf T}}J_k(\boldsymbol{w}_{c,n-1})(\boldsymbol{w}_{k,n-1} - \boldsymbol{w}_{c,n-1})\notag\\
   & \overset{(b)}{\le}  J(w^{\star}) - \mathds{E}J(\boldsymbol{w}_{c,n-1}) + O(\mu)(\mathds{E}J(\boldsymbol{w}_{c,n-1}) - \mathds{E}J(w^{\star})) \notag\\
   &\quad\;+O(\mu)
\end{align}
where $(a)$ follows from the convexity of $J_k$, and the derivation of $(b)$ is similar to (\ref{J_4}).

Substituting (\ref{J_4}) and (\ref{J_5}) into \eqref{J_3}, we obtain
\begin{align}\label{J_6}
    &\sum_k\pi_k \mathds{E}\nabla_{w^{\sf T}} J_k(\boldsymbol{w}_{k,n-1})(w^{\star}- \boldsymbol{w}_{c,n-1}) \notag\\
    &\le J(w^{\star}) - J(\boldsymbol{w}_{c,n-1}) + O(\mu)(\mathds{E}J(\boldsymbol{w}_{c,n-1}) - \mathds{E}J(w^{\star})) \notag\\
    &\quad\;+O(\mu)
\end{align}

Next, substituting (\ref{J_1}), (\ref{J_2}) and (\ref{J_6}) into (\ref{J_s}) gives
\begin{align}
 &\mathds{E}\Vert \boldsymbol{w}_{c,n} - w^{\star}\Vert^2 - \mathds{E}\Vert \boldsymbol{w}_{c,n-1} - w^{\star}\Vert^2\notag\\
 &\le 2\mu(J(w^{\star}) - J(\boldsymbol{w}_{c,n-1}))  + O(\mu^2)(J(w^{\star}) - J(\boldsymbol{w}_{c,n-1})) \notag\\
 &\quad\;+ O(\mu^2)  
\end{align}
when $\mu$ is sufficiently small such that
\begin{align}
    2\mu - O(\mu^2)>0
\end{align}
the telescopic cancellation gives
\begin{align}\label{J_7}
&\frac{1}{N}\sum_{n=1}^{N} \mathds{E}J(\boldsymbol{w}_{c,n-1}) - J(w^{\star}) \notag\\
&\le \frac{1}{N(2\mu-O(\mu^2))}\Vert w_{c,0} - w^{\star}\Vert^2 + O(\mu)  \notag\\
&= O(\frac{1}{\mu N})+ O(\mu)
\end{align}
The above inequality verifies the convergence of the algorithms with $\boldsymbol{w}_{c,n-1}$. We can also verify the convergence of the algorithms with $\boldsymbol{w}_{k,n-1}$. Basically, for any local agent $k$, it holds that 
\begin{align}\label{J_8}
    &\mathds{E}J(\boldsymbol{w}_{k,n-1}) - J(w^{\star}) \notag\\
    &= \mathds{E}J(\boldsymbol{w}_{k,n-1}) - \mathds{E}J(\boldsymbol{w}_{c,n-1}) +  \mathds{E}J(\boldsymbol{w}_{c,n-1}) - J(w^{\star})
\end{align}
for which we have
\begin{align}\label{J_9}
 &\mathds{E}J(\boldsymbol{w}_{k,n-1}) - \mathds{E}J(\boldsymbol{w}_{c,n-1}) \notag\\
 &\overset{(a)}{\le} \mathds{E}\nabla_{w^{\sf T}}J(\boldsymbol{w}_{c,n-1})(\boldsymbol{w}_{k,n-1} - \boldsymbol{w}_{c,n-1}) \notag\\
 &\quad\;+ \frac{L}{2}\mathds{E}\Vert \boldsymbol{w}_{k,n-1} - \boldsymbol{w}_{c,n-1} \Vert^2 + O(\epsilon )\mathds{E}\Vert \boldsymbol{w}_{k,n-1} - \boldsymbol{w}_{c,n-1} \Vert\notag\\
 &\overset{(b)}{\le}   \mathds{E}\nabla_{w^{\sf T}}J(\boldsymbol{w}_{c,n-1})(\boldsymbol{w}_{k,n} - \boldsymbol{w}_{c,n}) + O(\mu)\notag\\
 &\overset{(c)}{\le}  O(\mu)(\mathds{E}J(\boldsymbol{w}_{c,n-1}) - J(w^{\star})) + O(\mu)
\end{align}
where (a) follows from (\ref{u_affine}), (b) follows from (\ref{th_wkc}), and (c) follows from  similar techniques to (\ref{J_4}). Then substituting (\ref{J_7}) and (\ref{J_9}) into (\ref{J_8}) gives
\begin{align}
    \frac{1}{N}\sum_{n=1}^{N} \mathds{E}J(\boldsymbol{w}_{k,n-1}) - J(w^{\star}) \le  O(\frac{1}{\mu N})+ O(\mu)
\end{align}
\end{proof}

\section{Proof of Lemma \ref{lemma3}}\label{ap4}
\begin{proof}
We now analyze the difference between the iterates at the agents and the centroid. To begin with, recalling the recursion in \eqref{r_non_c}, the block centroid $\boldsymbol{\scriptstyle\mathcal{W}}_{c,n}$ satisfies the following relation:
\begin{align}
        \boldsymbol{\scriptstyle\mathcal{W}}_{c,n} &= \mathbbm{1}\otimes \boldsymbol{w}_{c,n} \notag\\
        &=  (\mathbbm{1}\pi^{\sf T}\otimes I)\boldsymbol{\scriptstyle\mathcal{W}}_{n}\notag\\
        &\overset{(a)}{=} (\mathbbm{1}\pi^{\sf T}\otimes I)(\boldsymbol{\scriptstyle\mathcal{W}}_{n-1} - \mu\mathcal{G}_{n-1} - \mu\bar{\boldsymbol{s}}_n^B)
\end{align}
where $(a)$ follows from (\ref{r_non_c}) and \eqref{a1a2p}. We thus obtain the following recursion which relates to the network disagreement:
{
\begin{align}
\label{rewc}
      \boldsymbol{\scriptstyle\mathcal{W}}_{n} - \boldsymbol{\scriptstyle\mathcal{W}}_{c,n} & = (\mathcal{A}^{\sf T} - \mathbbm{1}\pi^{\sf T}\otimes I )\boldsymbol{\scriptstyle\mathcal{W}}_{n-1} \notag\\
      &\quad\;- \mu(\mathcal{A}_2^{\sf T} - \mathbbm{1}\pi^{\sf T}\otimes I )(\mathcal{G}_{n-1}+ \bar{\boldsymbol{s}}_n^B)\notag\\
    & \overset{(a)}{=}  (\mathcal{A}^{\sf T} - \mathbbm{1}\pi^{\sf T}\otimes I )(I -\mathbbm{1}\pi^{\sf T}\otimes I  )\boldsymbol{\scriptstyle\mathcal{W}}_{n-1}\notag\\
    &\quad\;- \mu(\mathcal{A}_2^{\sf T} - \mathbbm{1}\pi^{\sf T}\otimes I )(\mathcal{G}_{n-1}+ \bar{\boldsymbol{s}}_n^B)\notag\\
    &= (\mathcal{A}^{\sf T} - \mathbbm{1}\pi^{\sf T}\otimes I )(\boldsymbol{\scriptstyle\mathcal{W}}_{n-1}-\boldsymbol{\scriptstyle\mathcal{W}}_{c,n-1})  \notag\\
    &\quad\; - \mu(\mathcal{A}_2^{\sf T} - \mathbbm{1}\pi^{\sf T}\otimes I )(\mathcal{G}_{n-1}+ \bar{\boldsymbol{s}}_n^B)
\end{align}}
where $(a)$ follows from
\begin{equation}
    \mathcal{A}^{\sf T} - \mathbbm{1}\pi^{\sf T}\otimes I = (\mathcal{A}^{\sf T} - \mathbbm{1}\pi^{\sf T}\otimes I )(I -\mathbbm{1}\pi^{\sf T}\otimes I  )
\end{equation}

Using eigen-structure of the combination matrix $A$ from (\ref{decompose})--(\ref{AC_12a}), we have
\begin{equation}\label{ad}
    \mathcal{A}^{\sf T} -  \mathbbm{1}\pi^{\sf T}\otimes I_M =   (\mathcal{V}^{\sf T})^{-1}\widetilde{\mathcal{J}}^{\sf T}\mathcal{V}^{\sf T}
\end{equation}
where
\begin{equation}
    \widetilde{\mathcal{J}}= \widetilde{J}\otimes I_M = \left[
    \begin{array}{cc}
         0&0  \\
         0& J_{\alpha}
    \end{array}\right]\otimes I_M 
\end{equation}
{Also, for the consensus strategy, we have:
\begin{align}\label{ad_consen}
\mathcal{A}_2^{\sf T} -  \mathbbm{1}\pi^{\sf T}\otimes I_M = &  I_{KM} -  \mathbbm{1}\pi^{\sf T}\otimes I_M \notag\\
\overset{(a)}{=}& (\mathcal{V}^{\sf T})^{-1}\left[
    \begin{array}{cc}
         0&0  \\
         0& I_{(K-1)M}
    \end{array}\right]\mathcal{V}^{\sf T} 
\end{align}
where $(a)$ follows from \eqref{AC_12a}.}

Now consider
\begin{equation}
\overline{\boldsymbol{\scriptstyle\mathcal{W}}}_n \overset{\Delta}{=} \mathcal{V}^{\sf T}(\boldsymbol{\scriptstyle\mathcal{W}_{n}} - \boldsymbol{\scriptstyle\mathcal{W}_{c,n}})
\end{equation}
and substituting (\ref{rewc}) and (\ref{ad}) into it, we get
\begin{align}\label{12}
\left\Vert\overline{\boldsymbol{\scriptstyle\mathcal{W}}}_n\right\Vert^2=&\Big\Vert \mathcal{V}^{\sf T}( (\mathcal{V}^{\sf T})^{-1}\widetilde{\mathcal{J}}^{\sf T}\mathcal{V}^{\sf T})(\boldsymbol{\scriptstyle\mathcal{W}}_{n-1} - \boldsymbol{\scriptstyle\mathcal{W}}_{c,n-1})\notag\\
&- \mu\mathcal{V}^{\sf T}(\mathcal{A}_2^{\sf T} -  \mathbbm{1}\pi^{\sf T}\otimes I_M)(\mathcal{G}_{n-1} +\bar{\boldsymbol{s}}_n^B)\Big\Vert^2\notag\\
=&\left\Vert\widetilde{\mathcal{J}}^{\sf T}\overline{\boldsymbol{\scriptstyle\mathcal{W}}}_{n-1}-\mu\mathcal{V}^{\sf T}(\mathcal{A}_2^{\sf T} -  \mathbbm{1}\pi^{\sf T}\otimes I_M)(\mathcal{G}_{n-1}+ \bar{\boldsymbol{s}}_n^B)\right\Vert^2\notag\\
        \overset{(a)}{\le}&\frac{1}{t}\left\Vert\widetilde{\mathcal{J}}^{\sf T}\right\Vert^2\left\Vert\overline{\boldsymbol{\scriptstyle\mathcal{W}}}_{n-1}\right\Vert^2 \notag\\
        &+ \frac{\mu^2\left\Vert\mathcal{V}^{\sf T}(\mathcal{A}_2^{\sf T} -  \mathbbm{1}\pi^{\sf T}\otimes I_M)\right\Vert^2}{1-t}\left\Vert\mathcal{G}_{n-1}+\bar{\boldsymbol{s}}_n^B\right\Vert^2
\end{align}
where $(a)$ follows from Jensen's inequality for any $0<t<1$. Similar to (\ref{AC_18}), we select
\begin{align}\label{vt}
t = \left\Vert\widetilde{\mathcal{J}}^{\sf T}\right\Vert = \left\Vert J_\alpha^{\sf T} \right\Vert = \sqrt{\tau(J_\alpha J_\alpha^{\sf T})} = \lambda_2+\alpha <1
\end{align}
where $\alpha$ is arbitrarily small to guarantee the rightmost inequality. Then, we have
\begin{align}\label{13}
 &\mathds{E}\left\{ \left\Vert\overline{\boldsymbol{\scriptstyle\mathcal{W}}}_n\right\Vert^2|\boldsymbol{\mathcal{F}}_{n-1}\right\}\notag\\
 &\overset{(a)}{\le}t\left\Vert\overline{\boldsymbol{\scriptstyle\mathcal{W}}}_{n-1}\right\Vert^2+ \frac{\mu^2 \left\Vert\mathcal{V}^{\sf T}(\mathcal{A}_2^{\sf T} -  \mathbbm{1}\pi^{\sf T}\otimes I_M)\right\Vert^2}{1-t}\notag\\
 &\quad\;\times\mathds{E}\left\{\sum\limits_{k=1}^K\left\Vert\mathds{E}\nabla_w Q_k(\boldsymbol{w}_{k,n-1};\widehat{\boldsymbol{x}}_{k,n},\boldsymbol{y}_{k,n})+ \bar{\boldsymbol{s}}^{B}_{k,n}\right\Vert^2|\boldsymbol{\mathcal{F}}_{n-1}
        \right\}\notag\\
&\overset{(b)}{=} t\left\Vert\overline{\boldsymbol{\scriptstyle\mathcal{W}}}_{n-1}\right\Vert^2+\frac{\mu^2 \left\Vert\mathcal{V}^{\sf T}(\mathcal{A}_2^{\sf T} -  \mathbbm{1}\pi^{\sf T}\otimes I_M)\right\Vert^2}{1-t}\notag\\
&\quad\;\times\mathds{E}\Bigg\{\sum\limits_{k=1}^K\left\Vert\mathds{E}\nabla_w Q_k(\boldsymbol{w}_{k,n-1};\widehat{\boldsymbol{x}}_{k,n},\boldsymbol{y}_{k,n})\right\Vert^2\notag\\
&\quad\;+ \left\Vert\bar{\boldsymbol{s}}^{B}_{k,n}\right\Vert^2|\boldsymbol{\mathcal{F}}_{n-1}\Bigg\}\notag\\
&\overset{(c)}{\le}  t\left\Vert\overline{\boldsymbol{\scriptstyle\mathcal{W}}}_{n-1}
\right\Vert^2 + 
\frac{\mu^2 \left\Vert\mathcal{V}^{\sf T}(\mathcal{A}_2^{\sf T} -  \mathbbm{1}\pi^{\sf T}\otimes I_M)\right\Vert^2}{1-t}K\left(G^2+\frac{\bar{\sigma}^2}{B}\right)
\end{align}
where $(a)$ follows from (\ref{dq}) and \eqref{s_n_bar}, $(b)$ follows from \eqref{s0}, and  $(c)$ follows from (\ref{b1}) and (\ref{sb2_n}). Taking expectations of both sides, we arrive at
\begin{align}
{\mathds{E}}\left\Vert\overline{\boldsymbol{\scriptstyle\mathcal{W}}}_n\right\Vert^2  \le &t\mathds{E}\left\Vert\overline{\boldsymbol{\scriptstyle\mathcal{W}}}_{n-1}
\right\Vert^2 \notag\\
&+ 
\frac{\mu^2 \left\Vert\mathcal{V}^{\sf T}{(\mathcal{A}_2^{\sf T} -  \mathbbm{1}\pi^{\sf T}\otimes I_M)}\right\Vert^2}{1-t}K\left(G^2+{\frac{\bar{\sigma}^2}{B}}\right)
\end{align}
Then after enough iterations $n_0$ such that
\begin{align}
   t^{n_0 + 1}\mathds{E}\Vert\overline{\boldsymbol{\scriptstyle\mathcal{W}}}_{-1}\Vert^2 \le& \frac{\mu^2 \left\Vert\mathcal{V}^{\sf T}{(\mathcal{A}_2^{\sf T} -  \mathbbm{1}\pi^{\sf T}\otimes I_M)}\right\Vert^2}{1-t}K\left(G^2+{\frac{\bar{\sigma}^2}{B}}\right)
\end{align}
which is equivalent to
\begin{align}
    n_0 \ge O \left(\frac{\rm{log} \mu}{\rm{log} \textit{t}}\right) 
\end{align}
we get
\begin{align}\label{nd_a}
&{\mathds{E}}\left\Vert\boldsymbol{\scriptstyle\mathcal{W}}_{n} - \boldsymbol{\scriptstyle\mathcal{W}}_{c,n}\right\Vert^2 \notag\\
&={\mathds{E}}\left\Vert(\mathcal{V}^{\sf T})^{-1}\overline{\boldsymbol{\scriptstyle\mathcal{W}}}_n\right\Vert^2\notag\\
&\le{\frac{2\mu^2 \left\Vert\mathcal{V}^{\sf T}(\mathcal{A}_2^{\sf T} -  \mathbbm{1}\pi^{\sf T}\otimes I_M)\right\Vert^2\left\Vert(\mathcal{V}^{\sf T})^{-1}\right\Vert^2}{(1-t)}K\left(G^2+\frac{\bar{\sigma}^2}{B}\right)}
\end{align}

{Note that for the consensus strategy, recalling \eqref{ad_consen} gives
\begin{align}
 \left\Vert\mathcal{V}^{\sf T}(\mathcal{A}_2^{\sf T} -  \mathbbm{1}\pi^{\sf T}\otimes I_M)\right\Vert^2 = \left\Vert\mathcal{V}^{\sf T} \right\Vert^2  
\end{align}
with which we have
\begin{align}\label{nd_a_consen}
{\mathds{E}}\left\Vert\boldsymbol{\scriptstyle\mathcal{W}}_{n} - \boldsymbol{\scriptstyle\mathcal{W}}_{c,n}\right\Vert^2 &\le\frac{2\mu^2 \left\Vert\mathcal{V}^{\sf T}\right\Vert^2\left\Vert(\mathcal{V}^{\sf T})^{-1}\right\Vert^2}{(1-t)}K\left(G^2+\frac{\bar{\sigma}^2}{B}\right) \notag\\
&= O(\mu^2)
\end{align}
while for diffusion, recalling \eqref{ad} gives
\begin{align}
 \left\Vert\mathcal{V}^{\sf T}(\mathcal{A}_2^{\sf T} -  \mathbbm{1}\pi^{\sf T}\otimes I_M)\right\Vert^2 \le  \left\Vert J_\alpha^{\sf T} \right\Vert^2 \left\Vert\mathcal{V}^{\sf T} \right\Vert^2   = t^2 \left\Vert\mathcal{V}^{\sf T} \right\Vert^2
\end{align}
with which we have
\begin{align}\label{nd_a_diff}
{\mathds{E}}\left\Vert\boldsymbol{\scriptstyle\mathcal{W}}_{n} - \boldsymbol{\scriptstyle\mathcal{W}}_{c,n}\right\Vert^2 &\le\frac{2\mu^2 t^2\left\Vert\mathcal{V}^{\sf T}\right\Vert^2\left\Vert(\mathcal{V}^{\sf T})^{-1}\right\Vert^2}{(1-t)}K\left(G^2+\frac{\bar{\sigma}^2}{B}\right) \notag\\
&= O(\mu^2)
\end{align}}

Furthermore,
\begin{align}\label{disa_w1}
    \mathds{E}\Vert\boldsymbol{w}_{k,n}-\boldsymbol{w}_{c,n}\Vert&\le(\mathds{E}\Vert\boldsymbol{w}_{k,n}-\boldsymbol{w}_{c,n}\Vert^2)^{\frac{1}{2}}\notag\\
&\le({\mathds{E}}\left\Vert\boldsymbol{\scriptstyle\mathcal{W}}_{n} - \boldsymbol{\scriptstyle\mathcal{W}}_{c,n}\right\Vert^2)^{\frac{1}{2}}\notag\\
&\le O(\mu)
\end{align}
\end{proof}

\section{Proof of Lemma \ref{lemma4}}\label{ap5}
\begin{proof}
From \eqref{s0}, (\ref{sn}), and (\ref{sb0}), we know that
\begin{align}\label{psz}
\mathds{E}\left\{\widehat{\boldsymbol{s}}_n^B|\boldsymbol{\mathcal{F}}_{n-1}
\right\} &= \mathds{E}\left\{\sum\limits_k \pi_k\bar{\boldsymbol{s}}^B_{k,n}|\boldsymbol{\mathcal{F}}_{n-1}\right\}\notag\\
&= \sum_k \pi_k \mathds{E}\left\{\bar{\boldsymbol{s}}^B_{k,n}|\boldsymbol{\mathcal{F}}_{n-1}\right\} \notag\\
&= 0 
\end{align}
Moreover, expressions (\ref{sb2_n}) and (\ref{sn}) gives
\begin{align}\label{pbs}
      {\mathds{ E}}\left\{\left\Vert\widehat{\boldsymbol{s}}_n^B\right\Vert^2\right\} \overset{(a)}{\le} {\mathds{ E}}\left\{{\mathds{ E}}\left\{\sum_k \pi_k \left\Vert\bar{\boldsymbol{ s}}^B_{k,n}\right\Vert^2|\boldsymbol{\mathcal{F}}_{n-1}\right\}\right\} \le {\frac{\bar{\sigma}^2}{B}}
\end{align}
where $(a)$ follows from Jensen's inequality. 

As for $\boldsymbol{d}_{n-1}$, after enough iterations $n_0$, we get
\begin{align}\label{51}
         {\mathds{ E}}\left\Vert\boldsymbol{d}_{n-1}\right\Vert^2 & \overset{(a)}{\le}\sum\limits_k \pi_k \mathds{E}\big\Vert\nabla_w Q_k(\boldsymbol{w}_{k,n-1};\widehat{\boldsymbol{x}}_{k,n},\boldsymbol{y}_{k,n})\notag\\
         &\quad\;-\mathds{E}\nabla_w Q_k(\boldsymbol{w}_{c,n-1};\widehat{\boldsymbol{x}}_{k,n},\boldsymbol{y}_{k,n})\big\Vert^2\notag\\
        &\overset{(b)}{\le} \sum\limits_k \pi_k L^2 \mathds{E}\left\Vert\boldsymbol{w}_{k,n-1}-\boldsymbol{w}_{c,n-1}\right\Vert^2\notag\\
        &\le \pi_{\max}L^2\mathds{E}\left\Vert\boldsymbol{\scriptstyle\mathcal{W}}_{n-1} - \boldsymbol{\scriptstyle\mathcal{W}}_{c,n-1}\right\Vert^2\notag\\
        &\overset{(c)}{\le}O(\mu^2)
\end{align}
where $(a)$ follows from Jensen's inequality,  $(b)$ follows from (\ref{smooth_q}), $(c)$ follows from (\ref{nd_a}), and $\pi_{\max}$ is the largest entry of $\pi$.
\end{proof}
\section{Proof of Theorem \ref{th2}}\label{ap6}
\begin{proof}
 Let
 \begin{equation}
     \gamma \overset{\Delta}{=}  \frac{1}{2L+1} < \frac{1}{L}
 \end{equation}
and consider the proximal operator associated with the risk function $J(w)$ at location $\boldsymbol{w}_{c,n-1}$:
\begin{align}\label{zn}
    \boldsymbol{\widehat{w}}_{n-1} = \mathop{\mathrm{argmin}}\limits_{\boldsymbol{z}} \left\{J({z}) + \left(L+\frac{1}{2}\right)\left\Vert{z} - \boldsymbol{w}_{c,n-1}\right\Vert^2\right\}
\end{align}
The resulting Moreau envelope is given by:
\begin{align}
    J_{\frac{1}{2L+1}}(\boldsymbol{w}_{c,n-1}) =  \min\limits_{{z}} J({z}) + \left(L+\frac{1}{2}\right)\left\Vert{z} - \boldsymbol{w}_{c,n-1} \right\Vert^2
\end{align}
We know from expression (\ref{pr1}) in Appendix \ref{ap7} that
\begin{align}\label{pr2}
    &{\mathds{E}\nabla_{w^{\sf T}}  Q_k(\boldsymbol{w}_{c,n-1};{\widehat{\boldsymbol{x}}}_{k,n},\boldsymbol{y}_{k,n})}(\boldsymbol{\widehat{w}}_{n-1}-\boldsymbol{w}_{c,n-1})\notag\\
    &\le J_k(\boldsymbol{\widehat{w}}_{n-1}) - J_k(\boldsymbol{w}_{c,n-1}) + \boldsymbol{\eta}_k  +\frac{L}{2}\left\Vert\boldsymbol{\widehat{w}}_{n-1} - \boldsymbol{w}_{c,n-1}\right\Vert^2
\end{align}
where
{
\begin{align}
    \boldsymbol{\eta}_k \le O(\Vert\boldsymbol{w}_k - \boldsymbol{w}_c \Vert) + O(\Vert\boldsymbol{w}_k - \boldsymbol{w}_c \Vert^2) + O(\epsilon^2)
\end{align}}
It is argued in \cite{Thekumparampil019} that the function {$f_k(\boldsymbol{w};\boldsymbol{x}_k, \boldsymbol{y}_k)$ defined by (\ref{f_g}) is $L$-weakly convex if (\ref{smooth_q}) is satisfied. It follows that the the function $f_k(\boldsymbol{w};\boldsymbol{x}_k, \boldsymbol{y}_k) + \left(L+\frac{1}{2}\right)||\boldsymbol{w}||^2$ is $(L+1)$-strongly-convex over $\boldsymbol{w}$. As a result, the function $J(\boldsymbol{w}) + \left(L+\frac{1}{2}\right)||\boldsymbol{w}||^2$ is also $(L+1)$-strongly-convex.} This implies that
\begin{align}\label{pr3}
    & J(\boldsymbol{w}_{c,n-1}) - J(\boldsymbol{\widehat{w}}_{n-1})- \frac{1}{2}(L+1)\left\Vert\boldsymbol{w}_{c,n-1}-\boldsymbol{ \widehat{w}}_{n-1}\right\Vert^2\notag\\
    &= J(\boldsymbol{w}_{c,n-1}) + \left(L+\frac{1}{2}\right)\left\Vert\boldsymbol{w}_{c,n-1}-\boldsymbol{w}_{c,n-1}\right\Vert^2 - J(\boldsymbol{\widehat{w}}_{n-1}) \notag\\
    &\quad\; -\left(L+\frac{1}{2}\right)\left\Vert\boldsymbol{w}_{c,n-1} - \boldsymbol{\widehat{w}}_{n-1}\right\Vert^2 + \frac{L}{2} \left\Vert\boldsymbol{w}_{c,n-1} - \boldsymbol{\widehat{w}}_{n-1}\right\Vert^2\notag\\
    &= J(\boldsymbol{w}_{c,n-1}) + \left(L + \frac{1}{2}\right)\left\Vert\boldsymbol{w}_{c,n-1}-\boldsymbol{w}_{c,n-1}\right\Vert^2 \notag\\
    &\quad\;- \min\limits_{{z}}\left\{J({z}) + \left(L+\frac{1}{2}\right)\left\Vert{z} - \boldsymbol{w}_{c,n-1}\right\Vert^2\right\}\notag\\
    &\quad\;+ \frac{L}{2}\left\Vert\boldsymbol{w}_{c,n-1} -\boldsymbol{ \widehat{w}}_{n-1}\right\Vert^2\notag\\
    &\overset{(a)}{\ge} (L+ \frac{1}{2})\left\Vert\boldsymbol{w}_{c,n-1} - \boldsymbol{\widehat{w}}_{n-1}\right\Vert^2\notag\\
    &\overset{(b)}{=} \frac{1}{4L+2}\left\Vert\nabla_w J_{\frac{1}{2L+1}}(\boldsymbol{w}_{c,n-1})\right\Vert^2
\end{align}
where $(a)$ follows from the strong convexity of the function $J(\boldsymbol{w}) + \left(L+\frac{1}{2}\right)||\boldsymbol{w}||^2$, and $(b)$ follows from (\ref{65}) with $\gamma = \frac{1}{2L+1}$. Then we get
\begin{align}\label{pr4}
 &J_{\frac{1}{2L+1}}(\boldsymbol{w}_{c,n}) \notag\\
 &=  \min\limits_{{z}} J({z}) + \left(L+\frac{1}{2}\right)\left\Vert\boldsymbol{w}_{c,n} - {z}\right\Vert^2 \notag\\
& \le  J(\boldsymbol{\widehat{w}}_{n-1}) + \left(L+\frac{1}{2}\right)\left\Vert\boldsymbol{w}_{c,n}-\boldsymbol{\widehat{w}}_{n-1}\right\Vert^2\notag\\
        & \overset{(a)}{=}  J(\boldsymbol{\widehat{w}}_{n-1}) + \left(L+\frac{1}{2}\right)\bigg\Vert\boldsymbol{w}_{c,n-1}-\boldsymbol{\widehat{w}}_{n-1}- \mu\boldsymbol{d}_{n-1}-\mu\widehat{\boldsymbol{s}}_n^B  \notag\\
        &\quad\;- \mu\sum\limits_k \pi_k{\mathds{E}\nabla_{w}  Q_k(\boldsymbol{w}_{c,n-1};{\widehat{\boldsymbol{x}}}_{k,n},\boldsymbol{y}_{k,n})} \bigg\Vert^2\notag\\
        &=  J(\boldsymbol{\widehat{w}}_{n-1}) + \left(L+\frac{1}{2}\right)\left\Vert\boldsymbol{w}_{c,n-1}-\boldsymbol{\widehat{w}}_{n-1}\right\Vert^2 +\mu^2\left(L+\frac{1}{2}\right)\notag\\
        &\quad\ \times\Big\Vert\sum\limits_k \pi_k{\mathds{E}\nabla_{w}  Q_k(\boldsymbol{w}_{c,n-1};{\widehat{\boldsymbol{x}}}_{k,n},\boldsymbol{y}_{k,n})}+ \boldsymbol{d}_{n-1}+\widehat{\boldsymbol{s}}_n^B \Big\Vert^2\notag\\
        &\quad\ + 2\mu \left(L+\frac{1}{2}\right) \Big(\boldsymbol{d}_{n-1}+ \widehat{\boldsymbol{s}}_n^B \notag\\
        &\quad\ +\sum\limits_k \pi_k {\mathds{E}\nabla_{w}  Q_k(\boldsymbol{w}_{c,n-1};{\widehat{\boldsymbol{x}}}_{k,n},\boldsymbol{y}_{k,n})} \Big)^{\sf T}\left(\boldsymbol{\widehat{w}}_{n-1} -\boldsymbol{w}_{c,n-1}\right) \notag\\
         &\overset{(b)}{\le} J_{\frac{1}{2L+1}}(\boldsymbol{w}_{c,n-1}) \notag\\
         &\quad\ +2\mu \left(L+\frac{1}{2}\right) (\boldsymbol{d}_{n-1}+\widehat{\boldsymbol{s}}_n^B)^{\sf T}(\boldsymbol{\widehat{w}}_{n-1}-\boldsymbol{w}_{c,n-1}) \notag\\    
         &\quad\ + 2\mu \left(L+\frac{1}{2}\right)\bigg(\sum\limits_k \pi_k J_k(\boldsymbol{\widehat{w}}_{n-1}) - \sum\limits_k \pi_k J_k(\boldsymbol{w}_{c,n-1}) \notag\\
         &\quad\ +\sum\limits_k \pi_k\boldsymbol{\eta}_k  +\frac{L}{2}\left\Vert\boldsymbol{\widehat{w}}_{n-1} - \boldsymbol{w}_{c,n-1}\right\Vert^2\bigg) \notag\\
         &\quad\   + \mu^2\left(L+\frac{1}{2}\right)\Big\Vert\sum\limits_k \pi_k{\mathds{E}\nabla_{w}  Q_k(\boldsymbol{w}_{c,n-1};{\widehat{\boldsymbol{x}}}_{k,n},\boldsymbol{y}_{k,n})}\notag\\
         &\quad\ +\boldsymbol{d}_{n-1}+\widehat{\boldsymbol{s}}_n^B\Big\Vert^2\notag\\
         &\le J_{\frac{1}{2L+1}}(\boldsymbol{w}_{c,n-1})  + 2\mu \left(L+\frac{1}{2}\right)\Big(J(\boldsymbol{\widehat{w}}_{n-1}) - J(\boldsymbol{w}_{c,n-1}) \notag\\
         &\quad\ +\boldsymbol{\eta} +   \frac{L}{2}\left\Vert\boldsymbol{\widehat{w}}_{n-1} - \boldsymbol{w}_{c,n-1}\right\Vert^2 \Big)\notag\\
         &\quad\ + 2\mu \left(L+\frac{1}{2}\right) \left(\boldsymbol{d}_{n-1}+\widehat{\boldsymbol{s}}_n^B\right)^{\sf T}(\boldsymbol{\widehat{w}}_{n-1} -\boldsymbol{w}_{c,n-1})\notag\\
         &\quad\ + \mu^2(L +\frac{1}{2})\Big\Vert\sum\limits_k \pi_k{\mathds{E}\nabla_{w}  Q_k(\boldsymbol{w}_{c,n-1};{\widehat{\boldsymbol{x}}}_{k,n},\boldsymbol{y}_{k,n})}\notag\\
         &\quad\ +\boldsymbol{d}_{n-1}+\widehat{\boldsymbol{s}}_n^B\Big\Vert^2
\end{align}
where $(a)$ follows from (\ref{wce}), $(b)$ follows from (\ref{pr2}), and
\begin{align}
\boldsymbol{\eta} &= \sum\limits_k \pi_k\boldsymbol{\eta}_k\le O(\sum_k \pi_k\Vert\boldsymbol{w}_k - \boldsymbol{w}_c \Vert) + O(\epsilon^2)
\end{align}
follows from (\ref{eta_kkk}). Then, conditioning both side of (\ref{pr4}) and substituting (\ref{psz}) into it, we get
\begin{align}\label{pr5}
    &\mathds{E}\left\{J_{\frac{1}{2L+1}}(\boldsymbol{w}_{c,n})|\boldsymbol{\mathcal{F}}_{n-1}\right\}\notag\\
    &\le J_{\frac{1}{2L+1}}(\boldsymbol{w}_{c,n-1}) + 2\mu \left(L+\frac{1}{2}\right) \Big( J(\boldsymbol{\widehat{w}}_{n-1}) -  J(\boldsymbol{w}_{c,n-1})\notag\\
   & \quad +\boldsymbol{\eta}  +\frac{1}{2}(L+1)\left\Vert\boldsymbol{\widehat{w}}_{n-1} - \boldsymbol{w}_{c,n-1}\right\Vert^2  \Big)\notag\\
   &\quad+ \mu \left(L+\frac{1}{2}\right)\mathds{E}\left\{\left\Vert\boldsymbol{d}_{n-1}\right\Vert^2|\boldsymbol{\mathcal{F}}_{n-1}\right\} \notag\\
   &\quad +  \mu^2 \left(L+\frac{1}{2}\right)\mathds{E}\bigg\{2\left\Vert\sum\limits_k \pi_k {\mathds{E}\nabla_{w}  Q_k(\boldsymbol{w}_{c,n-1};{\widehat{\boldsymbol{x}}}_{k,n},\boldsymbol{y}_{k,n})}\right\Vert^2  \notag\\
   &\quad+ 2\left\Vert\boldsymbol{d}_{n-1}\right\Vert^2 +\left\Vert\widehat{\boldsymbol{s}}_n^B\right\Vert^2|\boldsymbol{\mathcal{F}}_{n-1}\bigg\}
\end{align}
where we use the inequality
\begin{align}
    \boldsymbol{d}^{\sf T}_{n-1}(\boldsymbol{\widehat{w}}_{n-1} - \boldsymbol{w}_{c,n-1}) \le \frac{1}{2}\left\Vert\boldsymbol{d}_{n-1}\right\Vert^2+ \frac{1}{2}\left\Vert\boldsymbol{\widehat{w}}_{n-1} - \boldsymbol{w}_{c,n-1}\right\Vert^2
\end{align}
Taking the expectation of both sides of (\ref{pr5}), we obtain
\begin{align}\label{pr6}
         &\mathds{E}J_{\frac{1}{2L+1}}(\boldsymbol{w}_{c,n}) \notag\\
         &\le  \mathds{E} J_{\frac{1}{2L+1}}(\boldsymbol{w}_{c,n-1}) +2\mu \left(L+\frac{1}{2}\right) \mathds{E} 
        \Big\{J(\boldsymbol{\widehat{w}}_{n-1}) \notag\\
        &\quad\ - J(\boldsymbol{w}_{c,n-1}) + \boldsymbol{\eta}
   +\frac{1}{2}(L+1)\left\Vert|\boldsymbol{\widehat{w}}_{n-1} - \boldsymbol{w}_{c,n-1}\right\Vert^2\Big\}
    \notag\\
    &\quad\ +\mu^2 \left(L+\frac{1}{2}\right) \mathds{E}\left\Vert\widehat{\boldsymbol{s}}_n^B\right\Vert^2+(\mu + 2 \mu^2)\left(L+\frac{1}{2}\right)\mathds{E}\left\Vert \boldsymbol{d}_{n-1}\right\Vert^2\notag\\
   &\quad\ + 2\mu^2\left(L+\frac{1}{2}\right)\mathds{E}\left\Vert\sum\limits_k \pi_k{\mathds{E}\nabla_{w^{\sf T}}  Q_k(\boldsymbol{w}_{c,n-1};{\widehat{\boldsymbol{x}}}_{k,n},\boldsymbol{y}_{k,n})}\right\Vert^2 
\end{align}
Then, substituting (\ref{pr3}) into (\ref{pr6}) gives
\begin{align}\label{pr7}
     & 2 \mu \left(L+\frac{1}{2}\right)\cdot\frac{1}{4L+2} \mathds{E}\left\Vert\nabla_w J_{\frac{1}{2L + 1}}(\boldsymbol{w}_{c,n-1})\right\Vert^2 \notag\\
     & \le  2 \mu \left(L+\frac{1}{2}\right)\cdot \mathds{E} \bigg\{ J(\boldsymbol{w}_{c,n-1}) - J(\boldsymbol{\widehat{w}}_{n-1}) \notag\\
     &\quad\ -  \frac{1}{2}(L+1)\left\Vert\boldsymbol{\widehat{w}}_{n-1} - \boldsymbol{w}_{c,n-1}\right\Vert^2\bigg\}\notag\\
   &\le \mathds{E}J_{\frac{1}{2L+1}}(\boldsymbol{w}_{c,n-1})-\mathds{E}J_{\frac{1}{2L+1}}(\boldsymbol{w}_{c,n}) + 2\mu\left(L+\frac{1}{2}\right)\mathds{E}\boldsymbol{\eta}  \notag\\
   &\quad\ + 2\mu^2\left(L+\frac{1}{2}\right)\mathds{E}\left\Vert\sum\limits_k \pi_k{\mathds{E}\nabla_{w^{\sf T}}  Q_k(\boldsymbol{w}_{c,n-1};{\widehat{\boldsymbol{x}}}_{k,n},\boldsymbol{y}_{k,n})}\right\Vert^2\notag\\
   &\quad\ +(\mu + 2 \mu^2)\left(L +\frac{1}{2}\right)\mathds{E}\left\Vert \boldsymbol{d}_{n-1}\right\Vert^2+\mu^2 \left(L+\frac{1}{2}\right) \mathds{E}\left\Vert\widehat{\boldsymbol{s}}_n^B\right\Vert^2\notag\\
\end{align}
Substituting (\ref{b1}), \eqref{nd_a}, (\ref{disa_w1}), (\ref{pbs}), and (\ref{51}) into (\ref{pr7}), after $N$ iterations, we obtain
\begin{align}\label{pr10}
&\frac{1}{N}\sum\limits_{n}\mathds{E}||\nabla_w J_{\frac{1}{2L+1}}(\boldsymbol{w}_{c,n-1})||^2 \notag\\
&\le  \frac{2(J_{\frac{1}{2L+1}}(w_{c,n_0})-\Delta)}{\mu N}  + O(\mu) + O(\epsilon^2)
\end{align}
 where $\Delta = \min\limits_{{w}} J(w)$.

\end{proof}

\section{Proof of (\ref{pr2})}\label{ap7}
\begin{proof}
Let 
\begin{align}
    &\boldsymbol{\delta}_c^\star = \mathop{\mathrm{argmax}}\limits_{\left\Vert\delta\right\Vert_{p_k} \le \epsilon_k} Q_k(\boldsymbol{w}_c;\boldsymbol{x}_k+ \delta,\boldsymbol{y}_k)\notag\\
    &\boldsymbol{\delta}_k^\star = \mathop{\mathrm{argmax}}\limits_{\left\Vert\delta\right\Vert_{p_k} \le \epsilon_k} Q_k(\boldsymbol{w}_k;\boldsymbol{x}_k+ \delta,\boldsymbol{y}_k)\notag\\
&\widehat{\boldsymbol{\delta}}_k\approx\mathop{\mathrm{argmax}}\limits_{\left\Vert\delta\right\Vert_{p_k} \le \epsilon_k} Q_k(\boldsymbol{w}_k;\boldsymbol{x}_k+ \delta,\boldsymbol{y}_k)\notag\\
&\widehat{\boldsymbol{x}}_{k} = \boldsymbol{x}_k + \widehat{\boldsymbol{\delta}}_k
\end{align}
and for any $\boldsymbol{w}_c, \boldsymbol{w}_k \in \mathcal{F}_{n-1}$, the approximation error arising from the maximization problem in \eqref{f_g} admits:
\begin{align}\label{eofmax}
&\mathds{E}\{Q_k(\boldsymbol{w}_{c};\boldsymbol{x}_k+\boldsymbol{\delta}_{c}^\star,\boldsymbol{y}_k) -  Q_k(\boldsymbol{w}_c;\boldsymbol{x}_k +\widehat{\boldsymbol{\delta}}_k,\boldsymbol{y}_k) \vert\mathcal{F}_{n-1}\}\notag\\
&= \mathds{E}\Big\{Q_k(\boldsymbol{w}_{c};\boldsymbol{x}_k+\boldsymbol{\delta}_{c}^\star,\boldsymbol{y}_k) - Q_k(\boldsymbol{w}_{k};\boldsymbol{x}_k+\boldsymbol{\delta}_{c}^\star,\boldsymbol{y}_k) \notag\\
&\quad\ + Q_k(\boldsymbol{w}_{k};\boldsymbol{x}_k+\boldsymbol{\delta}_{c}^\star,\boldsymbol{y}_k) - Q_k(\boldsymbol{w}_{k};\boldsymbol{x}_k+\boldsymbol{\delta}_{k}^\star,\boldsymbol{y}_k) \notag\\
&\quad\ + Q_k(\boldsymbol{w}_{k};\boldsymbol{x}_k+\boldsymbol{\delta}_{k}^\star,\boldsymbol{y}_k) - Q_k(\boldsymbol{w}_{k};\widehat{\boldsymbol{x}}_k,\boldsymbol{y}_k) \notag\\
&\quad\ + Q_k(\boldsymbol{w}_{k};\widehat{\boldsymbol{x}}_k,\boldsymbol{y}_k) - Q_k(\boldsymbol{w}_{c};\widehat{\boldsymbol{x}}_k,\boldsymbol{y}_k)\vert\mathcal{F}_{n-1}\Big\}
\end{align}
for which we have
\begin{align}\label{eofmax_1}
&\mathds{E}\{Q_k(\boldsymbol{w}_{c};\boldsymbol{x}_k+\boldsymbol{\delta}_{c}^\star,\boldsymbol{y}_k) - Q_k(\boldsymbol{w}_{k};\boldsymbol{x}_k+\boldsymbol{\delta}_{c}^\star,\boldsymbol{y}_k)\vert\mathcal{F}_{n-1}\} \notag\\
&\overset{(a)}{\le} \mathds{E}\{\nabla_w Q_k(\boldsymbol{w}_{k};\boldsymbol{x}_k+\boldsymbol{\delta}_{c}^\star,\boldsymbol{y}_k)\vert\mathcal{F}_{n-1}\}^{\sf T}(\boldsymbol{w}_c - \boldsymbol{w}_k) \notag\\
&\quad\ + \frac{L}{2}\Vert\boldsymbol{w}_k - \boldsymbol{w}_c\Vert^2\notag\\
&\overset{(b)}{\le} G\Vert\boldsymbol{w}_c - \boldsymbol{w}_k\Vert  + \frac{L}{2}\Vert\boldsymbol{w}_k - \boldsymbol{w}_c\Vert^2
\end{align}
where $(a)$ follows from \eqref{smooth_q}, and $(b)$ follows from Assumption \ref{as5}. Then, as $\boldsymbol{\delta}_k^{\star}$ admits the maximum of $Q_k$ associated with $\boldsymbol{w}_{k}$, we have
\begin{align}\label{eofmax_2}
  \mathds{E}\{Q_k(\boldsymbol{w}_{k};\boldsymbol{x}_k+\boldsymbol{\delta}_{c}^\star,\boldsymbol{y}_k) - Q_k(\boldsymbol{w}_{k};\boldsymbol{x}_k+\boldsymbol{\delta}_{k}^\star,\boldsymbol{y}_k)\vert\mathcal{F}_{n-1}\} \le 0 
\end{align}
Next,
\begin{align}\label{eofmax_3}
    &\mathds{E}\{ Q_k(\boldsymbol{w}_{k};\boldsymbol{x}_k+\boldsymbol{\delta}_{k}^\star,\boldsymbol{y}_k) - Q_k(\boldsymbol{w}_{k};\widehat{\boldsymbol{x}}_k,\boldsymbol{y}_k)\vert \mathcal{F}_{n-1}\} \notag\\
    &\overset{(a)}{\le} \mathds{E}\{\nabla_{x^{\sf T}} Q_k(\boldsymbol{w}_{k};\widehat{\boldsymbol{x}}_k,\boldsymbol{y}_k)(\boldsymbol{\delta}_{k}^\star -\widehat{\boldsymbol{\delta}}_k) + \frac{L}{2}\Vert\boldsymbol{\delta}_{k}^\star -\widehat{\boldsymbol{\delta}}_k\Vert^2\vert\mathcal{F}_{n-1}\}\notag\\
    &=\mathds{E}\bigg\{\Big(\nabla_{x} Q_k(\boldsymbol{w}_{k};\widehat{\boldsymbol{x}}_k,\boldsymbol{y}_k) - \nabla_{x} Q_k(\boldsymbol{w}_{k};{\boldsymbol{x}}_k,\boldsymbol{y}_k) \notag\\
    &\quad\ +\nabla_{x} Q_k(\boldsymbol{w}_{k};{\boldsymbol{x}}_k,\boldsymbol{y}_k)\Big)^{\sf T}(\boldsymbol{\delta}_{k}^\star -\widehat{\boldsymbol{\delta}}_k) + \frac{L}{2}\Vert\boldsymbol{\delta}_{k}^\star -\widehat{\boldsymbol{\delta}}_k\Vert^2\vert\mathcal{F}_{n-1}\bigg\}\notag\\
    &\overset{(b)}{\le}\mathds{E}\{ \Vert\nabla_{x} Q_k(\boldsymbol{w}_{k};\widehat{\boldsymbol{x}}_k,\boldsymbol{y}_k) - \nabla_{x} Q_k(\boldsymbol{w}_{k};{\boldsymbol{x}}_k,\boldsymbol{y}_k)\Vert \Vert\boldsymbol{\delta}_{k}^\star -\widehat{\boldsymbol{\delta}}_k\Vert \notag\\
    &\quad\ + \frac{L}{2}\Vert\boldsymbol{\delta}_{k}^\star -\widehat{\boldsymbol{\delta}}_k\Vert^2
      \vert\mathcal{F}_{n-1}\}\notag\\
     & = O(\epsilon^2)
\end{align}
where $(a)$ and $(c)$ follow from the smoothness condition in \eqref{as2_2}, and $(b)$ is due to
\begin{align}
    \nabla_{x^{\sf T}} Q_k(\boldsymbol{w}_k;\boldsymbol{x}_k,\boldsymbol{y}_k)(\boldsymbol{\delta}_k^\star-\widehat{\boldsymbol{\delta}}_k) \le 0
\end{align}
since $\widehat{\boldsymbol{\delta}}_k$ is the exact maximizer of the linear approximation in \eqref{pin} if the single-step approximation methods, e.g., FGM and FGSM, are used. For the multi-step attack methods, e.g., PGD and PGM, deriving a direct upper bound for the approximation error is not tractable, as these methods do not have exact expressions. Fortunately, multi-step attack methods have been empirically shown to provide a better approximation of the optimal perturbation \cite{madry2017towards}. As a result, their associated approximation error is expected to be smaller than the bound shown in \eqref{eofmax_3}. Thus, the bound in \eqref{eofmax_3} can also be applied to multi-step attacks. 

{Finally, we have
\begin{align}\label{eofmax_4}
&\mathds{E}\{ Q_k(\boldsymbol{w}_{k};\widehat{\boldsymbol{x}}_k,\boldsymbol{y}_k) - Q_k(\boldsymbol{w}_{c};\widehat{\boldsymbol{x}}_k,\boldsymbol{y}_k)\vert\mathcal{F}_{n-1}\}\notag\\
&\overset{(a)}{\le} \mathds{E}\{\nabla_w Q_k(\boldsymbol{w}_{c};\widehat{\boldsymbol{x}}_k,\boldsymbol{y}_k) \vert\mathcal{F}_{n-1} \}^{\sf T}(\boldsymbol{w}_k - \boldsymbol{w}_c) \notag\\
&\quad\ + \frac{L}{2}\Vert\boldsymbol{w}_k - \boldsymbol{w}_c\Vert^2\notag\\
&\overset{(b)}{\le} G\Vert\boldsymbol{w}_c - \boldsymbol{w}_k\Vert  + \frac{L}{2}\Vert\boldsymbol{w}_k - \boldsymbol{w}_c\Vert^2
\end{align}
where $(a)$ follows from \eqref{smooth_q}, and $(b)$ follows from Assumption \ref{as5}.}

Substituting \eqref{eofmax_1},\eqref{eofmax_2},\eqref{eofmax_3},\eqref{eofmax_4} into (\ref{eofmax}) gives
\begin{align}\label{eofmax_f}
& \mathds{E}\{Q_k(\boldsymbol{w}_{c};\boldsymbol{x}_k+\boldsymbol{\delta}_{c}^\star,\boldsymbol{y}_k) -  Q_k(\boldsymbol{w}_c;\boldsymbol{x}_k +\widehat{\boldsymbol{\delta}}_k,\boldsymbol{y}_k) \vert\mathcal{F}_{n-1}\}\notag\\
&\le O(\Vert\boldsymbol{w}_k - \boldsymbol{w}_c\Vert) +   O(\Vert\boldsymbol{w}_k - \boldsymbol{w}_c\Vert^2) + O(\epsilon^2)
\end{align}
where the term $O(\epsilon^2)$ appears due to the approximation error when solving the maximization problem over perturbations. If the approximation error is 0, the term can be removed.

Then, for any given ${\boldsymbol{w}} \in  \boldsymbol{\mathcal{F}}_{n-1}$, we have
\begin{align}\label{pr1}
         J_k({\boldsymbol{w}}) & =\mathds{E}\left\{\max\limits_{\left\Vert\delta_k\right\Vert_{p_k}\le \epsilon_k} Q_k(\boldsymbol{w};\boldsymbol{x}_{k}+\delta_k,\boldsymbol{y}_k) \right\}\notag\\
         &\ge  \mathds{E} Q_k(\boldsymbol{{w}};\widehat{\boldsymbol{x}}_{k},\boldsymbol{y}_k)\notag\\
        &\overset{(a)}{\ge} \mathds{E}Q_k(\boldsymbol{w}_{c,n-1};\widehat{\boldsymbol{x}}_{k},\boldsymbol{y}_k)  - \frac{L}{2}\left\Vert\boldsymbol{{w}}-\boldsymbol{w}_{c,n-1}\right\Vert^2 \notag\\
        &\quad\ + {\mathds{E} \nabla_{w^{\sf T}}  Q_k(\boldsymbol{w}_{c,n-1};{\widehat{\boldsymbol{x}}}_{k},\boldsymbol{y}_k)}(\boldsymbol{{w}}-\boldsymbol{w}_{c,n-1})\notag\\
        &\overset{(b)}{\ge} \mathds{E}\left\{\max\limits_{\left\Vert\delta_k\right\Vert_{p_k} \le \epsilon_k} Q_k(\boldsymbol{w}_{c,n-1};\boldsymbol{x}_{k}+\delta_k,\boldsymbol{y}_k)\right\} - \boldsymbol{\eta}_k \notag\\
        &\quad\ - \frac{L}{2}\left\Vert\boldsymbol{{w}}-\boldsymbol{w}_{c,n-1}\right\Vert^2 \notag\\
        &\quad\ + {\mathds{E} \nabla_{w^{\sf T}}  Q_k(\boldsymbol{w}_{c,n-1};{\widehat{\boldsymbol{x}}}_{k},\boldsymbol{y}_k)}(\boldsymbol{{w}}-\boldsymbol{w}_{c,n-1}) \notag\\
         &= J_k(\boldsymbol{w}_{c,n-1}) - \boldsymbol{\eta}_k - \frac{L}{2}\left\Vert\boldsymbol{{w}}-\boldsymbol{w}_{c,n-1}\right\Vert^2  \notag\\
         &\quad\ + {\mathds{E} \nabla_{w^{\sf T}}  Q_k(\boldsymbol{w}_{c,n-1};{\widehat{\boldsymbol{x}}}_{k},\boldsymbol{y}_k)}(\boldsymbol{{w}}-\boldsymbol{w}_{c,n-1}) 
\end{align}
where $(a)$ follows from the L-weak convexity of $Q_k$ over $w$, {$(b)$ follows from (\ref{eofmax_f}) such that
\begin{align}\label{eta_kkk}
\boldsymbol{\eta}_k \le  O(\Vert\boldsymbol{w}_k - \boldsymbol{w}_c\Vert) +   O(\Vert\boldsymbol{w}_k - \boldsymbol{w}_c\Vert^2) + O(\epsilon^2)
\end{align}}
\end{proof}

\section{Proof of Corollary \ref{coro_sinh}}\label{ap_coro_sinh}
\begin{proof}
{
When the function $Q_k(w;\boldsymbol{x}_k+\delta,\boldsymbol{y}_k)$ is $\tau-$strongly concave over $\delta$, then the functions $f_k(w;\boldsymbol{x}_k,\boldsymbol{y}_k)$, $J_k(w)$ and $J(w)$ are differentiable. In this case, the recursion in \eqref{wc} can be written as:
\begin{align}\label{wc_new}
 \boldsymbol{w}_{c,n} =  \boldsymbol{w}_{c,n-1} - \mu\nabla_w J(\boldsymbol{w}_{c,n-1}) - \mu\widehat{s}_{n}^B - \mu \boldsymbol{e}_{n-1} - \mu\widehat{\boldsymbol{d}}_{n-1}
\end{align}
with
\begin{align}
\label{sn_co}
&\widehat{\boldsymbol{s}}_{n}^B \overset{\Delta}{=} \sum\limits_{k=1}^{K} \pi_k \bar{\boldsymbol{s}}^{B}_{k,n}\\
\label{en_new}
&\boldsymbol{e}_{n-1} \overset{\Delta}{=} \sum_k\pi_k \Big(\mathds{E}\nabla_w Q_k(\boldsymbol{w}_{k,n-1};\widehat{\boldsymbol{x}}_{k,n},\boldsymbol{y}_{k,n}) \notag\\
&\quad\quad\quad\; - \nabla_w J_k(\boldsymbol{w}_{k,n-1})\Big)\\
\label{dn_new}
&\widehat{\boldsymbol{d}}_{n-1} \overset{\Delta}{=}  \sum\limits_{k=1}^{K} \pi_k \Big(\nabla_w J_k(\boldsymbol{w}_{k,n-1})-\nabla_w J_k(\boldsymbol{w}_{c,n-1})\Big) \notag\\
&\quad\quad\; = \nabla_w J(\boldsymbol{w}_{k,n-1})-\nabla_w J(\boldsymbol{w}_{c,n-1})
\end{align}
where $\widehat{s}_{n}^B$ corresponds to the stochastic gradient noise, $\boldsymbol{e}_{n-1}$ arises from the approximation error when solving the inner maximization problem over the perturbation variable, and $\widehat{\boldsymbol{d}}_{n-1}$ is associated with network disagreement. Note that in this case, \eqref{sb2_n}--\eqref{wc} and \eqref{shatnb} still hold.}

{Moreover, it has been verified in \cite{sinha2017certifying} that the gradients of $f_k$ satisfy the following Lipschitz condition, for which detailed proof can be found from Sec B.2 of \cite{sinha2017certifying}:
\begin{align}
    \Vert \nabla_w f_k(w_2;\boldsymbol{x}_k,\boldsymbol{y}_k) - \nabla_w f_k(w_1;\boldsymbol{x}_k,\boldsymbol{y}_k)\Vert \le \widehat{L}\Vert w_2 - w_1\Vert
\end{align}
with
\begin{align}
    \widehat{L} = L + \frac{L^2}{\tau}
\end{align} 
As a result, $\nabla_w J_k(w)$ and $\nabla_w J(w)$ also satisfy the following Lipschitz condition:
\begin{align}
\label{jkl_new}
&\Vert \nabla_w J_k(w_2) - \nabla_w J_k(w_1)\Vert \le \widehat{L}\Vert w_2 - w_1\Vert\\
\label{jl_new}
   & \Vert \nabla_w J(w_2) - \nabla_w J(w_1)\Vert \le \widehat{L}\Vert w_2 - w_1\Vert
\end{align}}

Then, similar to \eqref{51}, by applying \eqref{jl_new} and Lemma \ref{lemma3}, the second-order moment of $\widehat{d}_{n-1}$ can be bounded:
\begin{align}\label{dhupp}
 \mathds{E}\Vert\widehat{d}_{n-1}\Vert^2 =&  \mathds{E}\Vert\nabla_w J(\boldsymbol{w}_{k,n-1})-\nabla_w J(\boldsymbol{w}_{c,n-1}) \Vert^2 \notag\\
 \le& \widehat{L}^2\mathds{E}\Vert \boldsymbol{w}_{k,n-1} - \boldsymbol{w}_{c,n-1}\Vert^2 \notag\\
 =& O(\mu^2)
\end{align}
Also, the second-order moment of the error term $\boldsymbol{e}_{n-1}$ can be upper bounded:
\begin{align}\label{eupper}
    &\mathds{E}\Vert \boldsymbol{e}_{n-1}\Vert^2 \notag\\
    &= \mathds{E}\bigg\Vert \sum_k\pi_k \Big(\mathds{E}\nabla_w Q_k(\boldsymbol{w}_{k,n-1};\widehat{\boldsymbol{x}}_{k,n},\boldsymbol{y}_{k,n}) \notag\\
    &\quad\ - \nabla_w J_k(\boldsymbol{w}_{k,n-1})\Big)\bigg\Vert^2\notag\\
    & \overset{(a)}{\le} \sum_k\pi_k\mathds{E}\Vert\nabla_w Q_k(\boldsymbol{w}_{k,n-1};\widehat{\boldsymbol{x}}_{k,n},\boldsymbol{y}_{k,n}) \notag\\
    &\quad\ - \nabla_w Q_k(\boldsymbol{w}_{k,n-1};\boldsymbol{x}_{k,n}^{\star},\boldsymbol{y}_{k,n})\Vert^2\\
    & \overset{(b)}{\le} O(\epsilon^2)
\end{align}
where $(a)$ follows from Jensen's inequality, and $(b)$ follows from \eqref{as2_4}.

Next, recalling \eqref{jl_new} gives
\begin{align}\label{luex}
    &\mathds{E}\{J(\boldsymbol{w}_{c,n})\vert\mathcal{F}_{n-1} \}\notag\\
    &\le J(\boldsymbol{w}_{c,n-1}) + \nabla_{w^{\sf T}} J(\boldsymbol{w}_{c,n-1})(\mathds{E}\boldsymbol{w}_{c,n} - \boldsymbol{w}_{c,n-1}) \notag\\
    &\quad\ + \frac{\widehat{L}}{2}\Vert\mathds{E}\boldsymbol{w}_{c,n} - \boldsymbol{w}_{c,n-1} \Vert^2\notag\\
    &\overset{(a)}{=}J(\boldsymbol{w}_{c,n-1}) - \mu\Vert\nabla_w J(\boldsymbol{w}_{c,n-1})\Vert^2 - \mu\nabla_{w^{\sf T}}J(\boldsymbol{w}_{c,n-1})\widehat{\boldsymbol{d}}_{n-1}\notag\\
    &\quad\ - \mu\nabla_{w^{\sf T}}J(\boldsymbol{w}_{c,n-1})\boldsymbol{e}_{n-1}  + \frac{\widehat{L}\mu^2}{2}\mathds{E}\Vert\widehat{\boldsymbol{s}}_{n}^B\Vert^2\notag\\
    &\quad\ + \frac{\widehat{L}\mu^2}{2}\Vert\nabla_w J(\boldsymbol{w}_{c,n-1})+\widehat{\boldsymbol{d}}_{n-1}+ \boldsymbol{e}_{n-1}\Vert^2 
\end{align}
where $(a)$ follows from \eqref{wc_new} and \eqref{psz}. Then applying the following inequality and Jensen's inequality:
\begin{align}
    \pm a^{\sf T}b \le \frac{1}{4}\Vert a\Vert^2 + \Vert b\Vert^2
\end{align}
to (\ref{luex}), and taking its expectation gives
\begin{align}\label{luex_2}
 \mathds{E}J(\boldsymbol{w}_{c,n})  \le&   \mathds{E}J(\boldsymbol{w}_{c,n-1}) - \frac{\mu}{2}(1-3\widehat{L}\mu)\mathds{E}\Vert \nabla_w J(\boldsymbol{w}_{c,n-1}) \Vert^2 \notag\\
 & + \mu(1+\frac{3\widehat{L}\mu}{2})\left(\mathds{E}\Vert\widehat{\boldsymbol{d}}_{n-1}\Vert^2 + \mathds{E}\Vert\boldsymbol{e}_{n-1}\Vert^2\right) \notag\\
 & + \frac{3\widehat{L}\mu^2}{2}\mathds{E}\Vert\widehat{\boldsymbol{s}}_{n}^B\Vert^2\notag\\
 \overset{(a)}{\le}& \mathds{E}J(\boldsymbol{w}_{c,n-1}) - \frac{\mu}{2}(1-3\widehat{L}\mu)\mathds{E}\Vert \nabla_w J(\boldsymbol{w}_{c,n-1}) \Vert^2 \notag\\
 &+ O(\mu^2) + O(\mu\epsilon^2)
\end{align}
where $(a)$ following from \eqref{shatnb}, \eqref{dhupp}, and \eqref{eupper}. Finally, when $\mu$ is sufficiently small such that $1-3\widehat{L}\mu<0$, the telescopic cancellation on \eqref{luex_2} gives
\begin{align}
    \frac{1}{N} \sum_n \mathds{E}\Vert \nabla_w J(\boldsymbol{w}_{c,n-1}) \Vert^2 \le O(\frac{1}{\mu N}) + O(\mu) + O(\epsilon^2)
\end{align}
\end{proof}

\section*{Acknowledgments}
We acknowledge the assistance of ChatGPT in improving the English presentation in this paper.
\bibliographystyle{ieeetr} 
\bibliography{reference} 

\vskip -0.3\baselineskip plus -1fil
\begin{IEEEbiographynophoto}{Ying Cao}
is currently pursuing a Ph.D. in the Electrical Engineering Doctoral program of EPFL. She received her bachelor's and master’s degree in Electronic and Information Engineering from Northwestern Polytechnical University, Xi’an, China, in 2017 and 2020, respectively. She received the China National Scholarship in 2018 and 2019. Her research interests include decentralized learning and the robustness of machine learning. 
\end{IEEEbiographynophoto} 
\vskip -2\baselineskip plus -1fil
\begin{IEEEbiographynophoto}{Elsa Rizk}
received the B.E. degree in Computer and Communications Engineering and the M.E. degree in Electrical and Computer Engineering from the American University of Beirut (AUB), Beirut, Lebanon. She received the Ph.D. degree in Computer Science from École Polytechnique Fédérale de Lausanne (EPFL), Lausanne, Switzerland. She is currently a Machine Learning Research Scientist with Daedalean AI, Zürich, Switzerland. Her research interests include distributed machine learning and explainable artificial intelligence.
\end{IEEEbiographynophoto}
\vskip -2\baselineskip plus -1fil
\begin{IEEEbiographynophoto}{Stefan Vlaski}
received his Ph.D.
degree in electrical and computer engineering from the
University of California, Los Angeles, USA, in 2019. He is currently a lecturer with Imperial College, U.K. From 2019 to 2021, he was a postdoctoral researcher with the Adaptive Systems Laboratory, École Polytechnique Fédérale de Lausanne, Switzerland. His research interests
include the intersection of machine learning, network science, and optimization. 
\end{IEEEbiographynophoto}
\vskip -3\baselineskip plus -1fil

\begin{IEEEbiographynophoto}{Ali H. Sayed}
is Dean of Engineering at EPFL, Switzerland, where he also directs the Adaptive Systems Laboratory. He served before as Distinguished Professor and Chair of Electrical Engineering at UCLA. He is a member of the US National Academy of Engineering (NAE) and The World Academy of Sciences (TWAS). He served as President of the IEEE Signal Processing Society in 2018 and 2019. An author of over 650 scholarly publications and 10 books, his research involves several areas including adaptation and learning theories,  statistical inference, and multi-agent systems. His work has been recognized with several awards including more recently the 2022 IEEE Fourier Technical Field Award and the 2020 IEEE Wiener Society Award. He is a Fellow of IEEE, EURASIP, and the American Association for the Advancement of Science (AAAS).
\end{IEEEbiographynophoto}

\end{document}